\documentclass{article}
\usepackage{microtype}
\usepackage{graphicx}
\usepackage{arydshln}
\usepackage{multirow}
\usepackage{multicol} 
\usepackage{makecell}
\usepackage[ruled]{algorithm2e}
\usepackage{etoolbox}
\makeatletter
\usepackage{wrapfig}
\usepackage{caption}
\usepackage{microtype}
\usepackage{graphicx}
\usepackage{booktabs}
\usepackage{ulem}
\usepackage{amsmath, amsthm, amssymb}
\usepackage{bm}
\usepackage{pifont}
\usepackage{hyperref}
\usepackage{amsthm}
\usepackage{microtype}
\usepackage{graphicx}
\usepackage{subcaption}
\usepackage{booktabs} 
\usepackage{hyperref}

\usepackage{graphicx}
\usepackage{subcaption}
\usepackage{booktabs}
\usepackage{makecell}
\usepackage{xcolor}
\usepackage{pifont}
\usepackage{colortbl}
\usepackage{ragged2e}
\usepackage{adjustbox} % key for preventing cutoff
\usepackage{tabularx}

\usepackage[most]{tcolorbox}
\usepackage{xcolor}
\usepackage{enumitem}
\usepackage{tabularx}
\usepackage{booktabs}
\usepackage{cuted} 
\usepackage{titletoc}
\usepackage{fancyvrb}
\usepackage{xparse}
\usepackage{tcolorbox}
\tcbuselibrary{skins,breakable,listings}
\usepackage{listings}
\usepackage{graphicx}
\usepackage{authblk}
\usepackage{xcolor}

\usepackage{booktabs}
\usepackage{array}

\newcolumntype{G}{>{\columncolor{gray!12}}c}

\newcommand{\best}[1]{\textcolor{red!80!black}{\textbf{#1}}}
\newcommand{\second}[1]{\textcolor{blue!80!black}{\textbf{#1}}}
\newcommand{\third}[1]{\textcolor{green!60!black}{\textbf{#1}}}

\DefineVerbatimEnvironment{SmallVerbatim}{Verbatim}{fontsize=\footnotesize}

\tcbuselibrary{listings,breakable,skins}

\newtcolorbox{promptbox}[1][]{%
  enhanced,
  breakable,
  colback=gray!5,
  colframe=gray!50,
  boxrule=0.6pt,
  arc=2pt,
  fontupper=\small,
  left=3pt,right=3pt,top=3pt,bottom=3pt,
  #1
}

\usepackage[preprint]{icml2026}
\makeatletter
\renewcommand{\printAffiliationsAndNotice}[1]{\global\icml@noticeprintedtrue}
\makeatother
\usepackage{amsmath}
\usepackage{amssymb}
\usepackage{mathtools}
\usepackage{amsthm}
\usepackage[table]{xcolor}  % 你已经用过就别重复

\definecolor{colblue}{RGB}{227,245,246}
\usepackage[capitalize,noabbrev]{cleveref}

\theoremstyle{plain}

\theoremstyle{definition}

\theoremstyle{remark}

\usepackage[table]{xcolor} % 或 \usepackage{xcolor} + \usepackage{colortbl}
\definecolor{colredlight}{HTML}{FFEAED} % 约等于 RGB(255,234,237)，和你图里很接近

\usepackage[textsize=tiny]{todonotes}

\begin{document}

\twocolumn[
 \icmltitle{T2S-Bench \& Structure-of-Thought: Benchmarking and Prompting Comprehensive Text-to-Structure Reasoning}




 \icmlsetsymbol{equal}{*}

 \begin{icmlauthorlist}
 \icmlauthor{Qinsi Wang}{equal,duke}
 \icmlauthor{Hancheng Ye}{equal,duke}
 \icmlauthor{Jinhee Kim}{equal,duke}
 \icmlauthor{Jinghan Ke}{ut}
 \icmlauthor{Yifei Wang}{duke}
 \icmlauthor{Martin Kuo}{duke}
 \icmlauthor{Zishan Shao}{duke}
 \icmlauthor{Dongting Li}{duke}
 \icmlauthor{Yueqian Lin}{duke}
 \icmlauthor{Ting Jiang}{duke}
 \icmlauthor{Chiyue Wei}{duke}
 \icmlauthor{Qi Qian}{meta}
 \icmlauthor{Wei Wen}{meta}
 \icmlauthor{Helen Li}{duke}
 \icmlauthor{Yiran Chen}{duke}
 \end{icmlauthorlist}

\vspace{4pt}
{\centering\normalsize
\textsuperscript{1}Duke University \quad
\textsuperscript{2}UT Austin \quad
\textsuperscript{3}Meta \par
\vspace{2pt}
\large\url{https://t2s-bench.github.io/T2S-Bench-Page/}\par
}

 \icmlaffiliation{duke}{Department of Electrical and Computer Engineering, Duke University.}
 \icmlaffiliation{ut}{Department of Computer Science, The University of Texas at Austin.}
 \icmlaffiliation{meta}{Meta}

 \icmlcorrespondingauthor{Qinsi Wang}{qinsi.wang@duke.edu}
 \icmlcorrespondingauthor{Hancheng Ye}{hancheng.ye@duke.edu}
 \icmlcorrespondingauthor{Jinhee Kim}{jinhee.kim@duke.edu}

 % You may provide any keywords that yo u find helpful for describing your
 \icmlkeywords{Machine Learning, ICML}

 \vskip 0.3in
]



\printAffiliationsAndNotice{} % no special notice (required even if empty)

\begin{abstract}
%为了提高模型在通用的复杂文本任务处理上的性能，在本文中，我们首先引入 \textbf{Structure of Thought (SoT)}, a prompting technique that explicitly guides models to construct intermediate text structures。实验表明在8个主流文本任务和3个不同架构的模型上SoT展现出一致的性能提升，暗示文本结构可以作为一种通用和robust的中间表示(IR)，从而提升各下游文本任务上的性能。
% To enhance model performance on general complex text tasks, in this work, we first introduce \textbf{Structure of Thought (SoT)}, a prompting technique explicitly guiding models to construct intermediate text structures. Experimental results consistently demonstrate significant performance improvements with SoT across eight mainstream text tasks and three distinct model architectures, indicating that structured text can serve as a universal and robust intermediate representation (IR) for enhancing downstream textual task performance.
Think about how human handles complex reading tasks: marking key points, inferring their relationships, and structuring information to guide understanding and responses. Likewise, \textit{can a large language model benefit from text structure to enhance text-processing performance?}
To explore it, in this work, we first introduce \textbf{Structure of Thought (SoT)}, a prompting technique that explicitly guides models to construct intermediate text structures, consistently boosting performance across eight tasks and three model families.
Building upon this insight, we present \textbf{T2S-Bench}, the first benchmark designed to evaluate and improve text-to-structure capabilities of models. T2S-Bench includes 1.8K samples across 6 scientific domains and 32 structural types, rigorously constructed to ensure accuracy, fairness, and quality. Evaluation on 45 mainstream models reveals substantial improvement potential: the average accuracy on the multi-hop reasoning task is only 52.1\%, and even the most advanced model achieves 58.1\% node accuracy in end-to-end extraction. 
Furthermore, on Qwen2.5-7B-Instruct, SoT alone yields an average +5.7\% improvement across eight diverse text-processing tasks, and fine-tuning on T2S-Bench further increases this gain to +8.6\%. These results highlight the value of explicit text structuring and the complementary contributions of SoT and T2S-Bench.
Dataset and eval code have been released at \href{https://t2s-bench.github.io/T2S-Bench-Page/}{here}.
% Furthermore, fine-tuning on the T2S-Bench training dataset achieves at most 8.5\% performance improvement on eight text-processing tasks, highlighting the importance of text structuring and the essential contribution of T2S-Bench.

% , with even top-tier models like Gemini 2.5 Pro achieving limited accuracy. Reinforcement learning fine-tuning using T2S-Bench significantly boosts downstream task performance, underscoring the critical role of structural comprehension in advancing LLM capabilities.

% Motivated by this insight, we introduce \textbf{T2S-Bench}, the first comprehensive benchmark designed to evaluate and advance text-to-structure capability. T2S-Bench contains 1.7K high-quality samples, including T2S-Train-1.2K, T2S-Bench-MR (500 multi-hop reasoning test instances), and T2S-Bench-E2E (87 end-to-end structure extraction test instances). The benchmark spans 6 major scientific domains, 13 subfields, and 32 structure types, and is built with a pipeline that enforces structural correctness, general and fair evaluation, and high sample quality. 
% Evaluating 45 widely-used models, we find substantial room for improvement: the average EM on T2S-Bench-MR is only 52.1\%, and even Gemini 2.5 Pro achieves just 58.1\% node accuracy in end-to-end extraction. Finally, RL fine-tuning on T2S-Train-1.2K improves downstream performance by 8.5\% (Qwen2.5-7B-Instruct) and 7.6\% (Llama3.1-8B-Instruct) on 8 text-processing tasks, highlighting the importance of text structuring and the significance of T2S-Bench.

\end{abstract}

% \vspace{-20pt}
\section{Introduction}
% \vspace{-5pt}
With the rapid integration of Large Language Models (LLMs) into real-world applications such as search engines~\cite{liang2025reasoning, xi2025survey}, office productivity tools~\cite{zheng2025pptagent, li2023sheetcopilot, fu2022doc2ppt} and scientific writing~\cite{zhang2025evolving, song2025evaluating}, high-quality text processing is evolving from merely demonstrating model capabilities into critical infrastructure directly impacting societal costs. 
Users increasingly depend on models to \textbf{Find} (identify evidence and relevant documents from massive datasets), \textbf{Fuse} (align and integrate viewpoints or facts from multiple sources), and \textbf{Form} (generate actionable conclusions, reports, decision-making evidence, or structured outputs). This "Find–Fuse–Form" pipeline underpins everyday model-driven workflows.
%and directly shapes the user experience.

% However, despite the growing demand, \textit{robust text processing remains a significant bottleneck in the further development of LLMs.}
% Extensive prior research highlights several persistent issues.
% First, unreliable "Find" ability. Comparative studies consistently show that model performance is highly sensitive to evidence placement and organization within inputs; simply relocating identical evidence from the beginning or end of a long input sequence to the middle can substantially degrade performance, indicating that models struggle to reliably locate and leverage critical information.
% Second, Fragile "fuse" ability. In typical RAG setups, adding more retrieved documents often yields non-monotonic gains and can even hurt performance, especially when stronger retrievers introduce more hard negatives or distractors. This exposes weaknesses in alignment, denoising, disambiguation, and conflict resolution.
% Third, instability "Form", exacerbated by the preceding Find and Fuse issues. Prior research and benchmarks indicate that LLMs frequently produce schema-violating outputs, missing or mismatched fields, and value-level hallucinations, especially when generating structured products such as tables, nested JSON, or CSV formats.
% Hence, the core challenge remains clear: \textbf{how can we uniformly measure and enhance the robustness of models in handling the universal Find–Fuse–Form text-processing pipeline?}

\begin{figure*}
% \vspace{-7pt}
 \centering
 \centerline{\includegraphics[width=\linewidth]{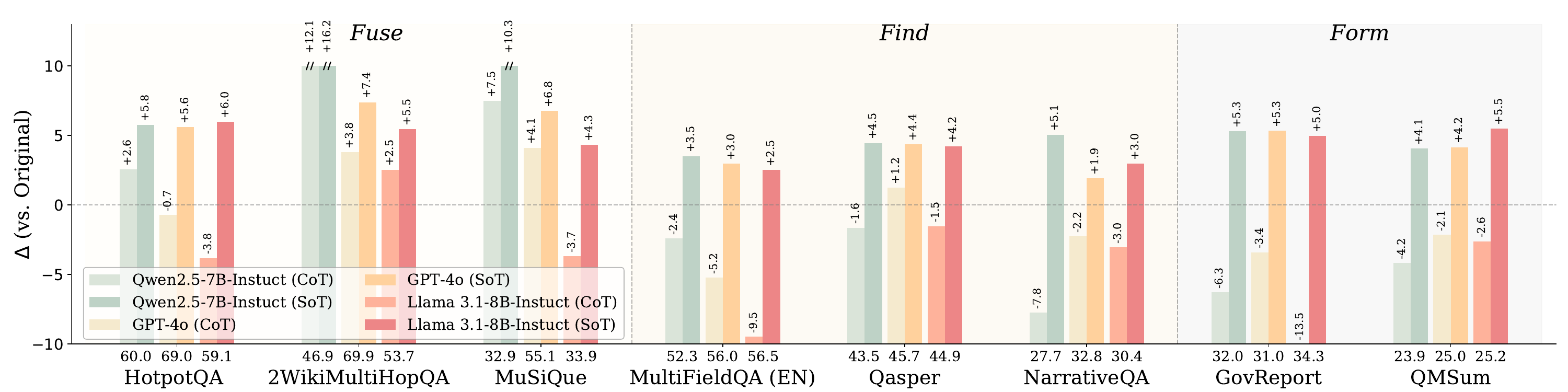}}
 % \vspace{-3pt}
     \caption{\textbf{Performance of SoT and Importance of Text Structuring.} We evaluated three models on eight distinct text-processing tasks using three prompting strategies: direct answering, Chain-of-Thought (CoT), and Structure of Thought (SoT). The horizontal axis shows the model's performance with direct answering, while the vertical axis indicates the performance change relative to direct answering. Our evaluations follow standards from lm-eval and Longbench tasks. SoT consistently boosts performance across different tasks and models.}
 % Notably, CoT often negatively impacts performance in Find and Form tasks, whereas SoT consistently boosts performance across different tasks and models..}
 % SoT在各任务上的性能表现以及文本结构化的重要性。我们在8个不同的文本处理任务上分别使用直接回答、CoT和SoT的prompt策略来评估三个模型的性能。横轴的数字表示模型在该任务上直接回答的性能，纵轴的数字表示相比于直接回答性能的变化。我们的评估遵循lm-eval以及Longbench task的评估标准。可以看到在Find和form任务上，CoT会带来性能的负增益，而SoT则能在不同的task和模型上实现一致的性能增长。
 \label{fig_1}
 % \vspace{-10pt}
\end{figure*}
% 描述目前高性能模型在long-context等复杂文本处理任务上仍然表现不佳，原因是因为缺少统一可泛化的中间表征来可扩展地表示复杂任务中文本之间的联系。例如，xxx。由此，我们提出xxx
%However, despite the growing demand, 当前模型在long-context等复杂文本处理任务上的性能仍然不佳，即使是最先进的模型在在Longbench上的性能也只是60%左右。这种性能poor的主要原因之一是当前的模型常把文本任务当成“直接生成”的单步task，缺少稳定的中间表示，导致不稳定的检索和不可控的生成。例如，highlight-guided generation通过先从长文中自规划出句子级“高亮/要点”作为 content plan，实现更好的summary和report生成；SRAG首先把多文档文本通过 SQL 驱动的模块抽取/检索并转成语义一致的关系表再做schema-aware 的表上推理，实现了mutidocumentqa任务上的显著提升。然而当前的这些方法通常是task-specific and input-structure-dependent的，无法泛化到通用的文本任务上。The core challenge remains:我们如何能找到一种通用且稳定的中间表示（IR）,并通过其有效评估和提升模型通用文本处理任务上的性能？

However, despite the growing demand, \textit{current models still struggle with complex text processing, especially long-context settings,} even state-of-the-art model reach only around 60\% on LongBench~\cite{bai2024longbench}.
One major reason is that nowadays models treat these tasks as end-to-end text generation, lacking stable intermediate representations (IR), resulting in unstable retrieval and uncontrollable generation.
Recent advances have shown promise by introducing intermediate steps: for instance, highlight-guided generation first extracts sentence-level highlights from long texts as a content plan to enhance summarization and reporting quality~\cite{du2025enhancing}; similarly, SRAG~\cite{lin2025srag} employs SQL-driven extraction modules to transform multi-document inputs into coherent relational tables, significantly improving multi-document QA tasks. However, these existing approaches are often task-specific and heavily reliant on input structures, thus failing to generalize effectively across diverse text tasks. 
Therefore, the core challenge remains: \textbf{\textit{How to find a universal and reliable intermediate representation (IR), and use it to systematically evaluate and improve LLMs on general text-processing tasks?}}

% However, despite the growing demand, \textit{robust text processing remains a significant bottleneck in the further development of LLMs.}
% For instance, even \hancheng{the state-of-the-art models can only} 
% % Gemini2.5-Pro 
% achieve 
% % an overall accuracy of only 
% about 63.3\% on the LongBench dataset~\cite{bai2024longbench}, which targets extensive context understanding, revealing a substantial gap in models’ ability to handle long-context comprehension. 
% % Controlled studies further show that this performance gap is not solely due to retrieval failures; models are notably sensitive to the position and organization of evidence within input texts. Simply moving relevant information from the beginning or end to the middle can significantly degrade performance, underscoring the unresolved need for robust retrieval and integration mechanisms.
% Although previous efforts have attempted to mitigate these issues through larger context windows, retrieval techniques, chunking, or multi-agent decomposition, these methods tend to be task-specific and input-structure-dependent, lacking a unified paradigm. 
% % Moreover, explicit reasoning approaches like Chain-of-Thought (CoT) are not universally beneficial and may even cause a detrimental "overthinking" effect in certain contexts. 
% The core challenge remains: \textbf{\textit{How can we uniformly measure and enhance the robustness of models in handling the universal Find–Fuse–Form text-processing pipeline?}}

To address this challenge, we first draw inspiration from how humans handle long textual information:
When humans comprehend lengthy texts or perform text generation tasks, an effective approach to improve quality is to perform \textbf{text structuring} by extracting key elements and clearly defining their relationships. This structuring facilitates quick retrieval of relevant information, integration of multiple textual sources, and clearer communication to others.
Based on this insight, we propose \textbf{Structure of Thought (SoT)}, a general prompting strategy that instructs models to first structure the text into key nodes and links before generating final answers. As demonstrated in Fig. \ref{fig_1}, SoT consistently and significantly improves model performance across eight mainstream text-processing tasks and three models, suggesting that text structure can serve as a universal intermediate representation (IR) to enhance various downstream tasks.

% To address this challenge, we first draw inspiration from how humans naturally handle long textual information and propose \textbf{Structure of Thought (SoT)}, a general prompting strategy that instructs models to first structure the text before generating final answers. As demonstrated in Fig. \ref{fig_1}, SoT consistently and significantly improves model performance across eight mainstream text-processing tasks and three models, suggesting that text structure can serve as a universal IR to enhance various downstream tasks.

Building on this insight, we propose \textbf{T2S-Bench}, the first comprehensive dataset designed to evaluate and enhance models' text structuring capabilities. T2S-Bench comprises a high-quality training set (T2S-Train-1.2k), a multi-hop reasoning evaluation set (T2S-Bench-MR with 500 samples), and an end-to-end structuring evaluation set (T2S-Bench-E2E with 87 samples). It covers six major scientific domains, 17 subfields, and 32 structure types, and offers several advantages:
(i) \textit{High Structural Accuracy}: By extracting text-structure pairs from rigorously vetted academic papers, T2S-Bench ensures structural correctness and reduces the inaccuracies associated with manual or model-based extraction.
(ii) \textit{Universal and Fair Evaluation}: T2S-Bench provides a broadly applicable evaluation suite across diverse text types. T2S-Bench-MR uses $4$ structural question categories and $32$ templates that require correct structuring to enable accurate multi-hop reasoning, while T2S-Bench-E2E reduces ambiguity from multiple valid structures by fixing key nodes and links and enforcing partial structural constraints for consistent, fair scoring.
(iii) \textit{High Sample Quality}: Constructing T2S-Bench involved over 6,000 model searches, six rounds of model validation, and three rounds of human quality checks spanning several months. Each sample was independently validated by at least two reviewers for structural, textual, and question accuracy.

We benchmarked 45 models from 10 families on T2S-Bench and found ample headroom: average exact match (EM) is only 52.1\% on T2S-Bench-MR. End-to-end structuring, especially node extraction, remains challenging: even state-of-art Gemini2.5-Pro reaches just 58.1\% accuracy. We further perform model fine-tuning on T2S-Train-1.2k, boosting performance at most by 8.5\% on average across eight downstream text processing tasks, showing that stronger structuring improves robustness and accuracy in downstream general text workflows. Overall, our contributions include:

% To thoroughly benchmark current model capabilities, we evaluated 45 models across 10 diverse model families using T2S-Bench. Our results reveal substantial room for improvement, with an average exact match (EM) of just 52.1\% on T2S-Bench-MR. 
% End-to-end structurization, particularly node extraction, remains highly challenging, even the state-of-the-art Gemini2.5-Pro achieved only 58.1\% accuracy. 
% We further perform RL fine-tuning on Qwen2.5-7B-Instruct and Llama3.1-8B-Instruct using T2S-Train-1.2k; the tuned models improve by 8.5\% and 7.6\% on average across eight downstream text processing tasks, reinforcing that stronger structuring ability can improve robustness and accuracy in general text workflows.
% Overall, our contributions include:
\begin{enumerate}
% \vspace{-7pt}
    \item \textbf{Proposing Structure of Thought (SoT)}, a prompting strategy that structurizes texts before answering, consistently improving performance across diverse tasks.
    
    % \vspace{-5pt}
    
    \item \textbf{Introducing T2S-Bench}, the first comprehensive dataset evaluating and improving text structuring capabilities, featuring 1.8k high-quality samples covering extensive scientific domains and structural types.
    % \vspace{-5pt}
    \item \textbf{Benchmarking 45 models using T2S-Bench}, identifying substantial room for improvement. Our findings also demonstrate that fine-tuning models on T2S-Train-1.2k significantly enhances downstream text-processing performance, underscoring the critical value and practical benefits of structured text processing.
    % \vspace{-7pt}
\end{enumerate}

\begin{figure*}
% \vspace{-5pt}
 \centering
 \centerline{\includegraphics[width=\linewidth]{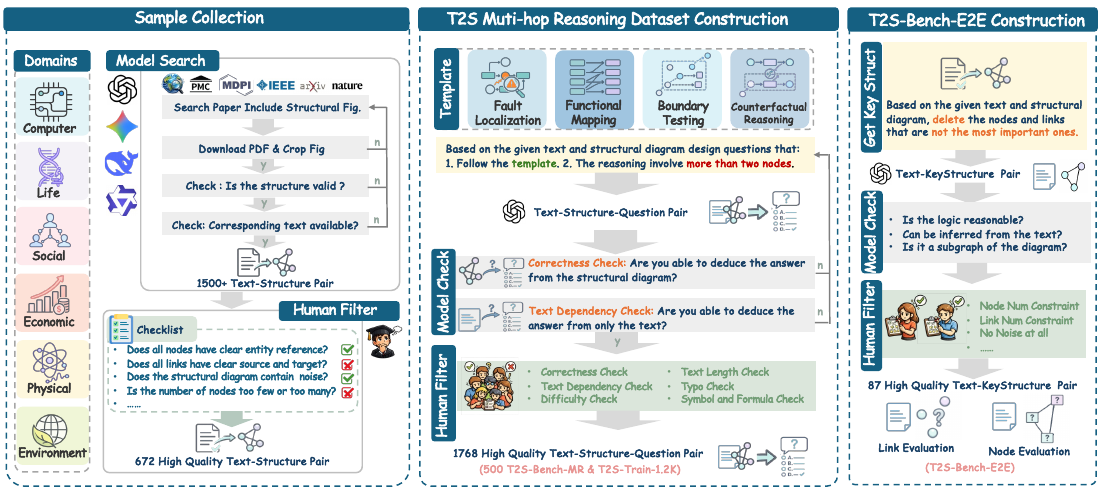}}
 % \vspace{-4pt}
 \caption{ \textbf{Construction Process of T2S-Bench,} including Sample Collection, Muti-hop Reasoning and End-to-End Dataset Construction.}
 \label{fig_2}
 % \vspace{-12pt}
\end{figure*}

% \vspace{-10pt}
\section{Motivation \& Challenges}
% \vspace{-5pt}
In this section, we explore and answer: \textit{Can explicitly structuring text improve a model’s general text-processing ability, and if so, by how much?}
We first probe the potential of text structuring by introducing SoT and evaluating its performance (Sect. \ref{sec22}).
Then we summarize three core challenges in building text-to-structure dataset (Sect. \ref{sec23}).

\subsection{Structure of Thought}
% \vspace{-5pt}
\label{sec22}
To explicitly leverage text structuring in general text processing, we introduce Structure of Thought (SoT), a universal prompting strategy. 
% SoT instructs models to first structurize text into nodes and links before answering questions. 
Specifically, SoT follows the format:

% \vspace{-5pt}
\begin{promptbox}

\textit{Based on the provided text, identify the key nodes and links between them, and provide the structure. Then give your answer based on the text and structure.}

\medskip
 Expected format:
 \vspace{-5pt}
\begin{Verbatim}[fontsize=\small]
[Structure]
{ "nodes": [
    {"id": "n1", "label": "Node1 Label"},
    {"id": "n2", "label": "Node2 Label"}],
  "links": [
    {"source": "n1", "target": "n2"}]}
[Answer]
your answer
\end{Verbatim}
\end{promptbox}
% \vspace{-5pt}
% "Based on the provided text, identify the key nodes (important entities/steps/concepts) and links (relations/flows/dependencies) between them, and output the structure. Then give your answer based on the text and structure.\\

% Expected format:
% [Structure]
% {
%   "nodes": [
%     { "id": "n1", "label": "Node Label" },
%     { "id": "n2", "label": "Another Node" }
%   ],
%   "links": [
%     { "source": "n1", "target": "n2", "label": "Link Label or empty" }
%   ]
% }
% [Answer]
% your answer"

By forcing the model to extract key nodes and links, SoT encourages models to process long texts similarly to human reasoning: first structuring textual information, then performing content retrieval, aggregation, and generation.  
Compared to Chain of Thought (CoT), SoT provides clearer task instructions, offering models a more concrete objective.
We evaluate three prompting strategies, including Direct Answer, CoT, and SoT, on eight widely used text-processing benchmarks using three different models. The results are shown in Fig. \ref{fig_1}, which we can draw three key observations:

% \vspace{-2pt}
 \textbf{1) Explicit text structuring substantially improves general text-related task performance.}
Across eight different tasks, SoT consistently delivers over 5\% performance improvement. Notably, on 2WikiMultiHopQA and MuSiQue, SoT improves performance by over 10\%.

% Across all eight tasks, SoT consistently delivers notable performance improvements without additional training. Notably, SoT achieves over 10% performance gains in tasks such as 2WikiMultiHopQA and Musique. These results highlight that explicit text structurization strongly facilitates general text-processing tasks, including Find (evidence retrieval), Fuse (multi-document aggregation), and Form (controlled downstream generation).
% \vspace{-5pt}
\textbf{2) SoT is more effective than CoT for text processing.}
While CoT is highly beneficial for domains like math and coding, it is not reliably helpful for general text tasks due to potential noise. 
% Our experiments reveal that CoT can even degrade performance in Find and Form tasks, as prolonged reasoning chains can introduce noise and negatively affect final outputs [1,2]. 
In contrast, SoT anchors reasoning to an explicit structure, leading to more consistent improvements.

% \vspace{-5pt}
\textbf{3) The benefit of structuring is both model- and task-agnostic.}
SoT consistently improves performance across all different model families and task types, suggesting that text structure can serve as a universal intermediate representation (IR) to enhance various downstream tasks. 
% This highlight the importance of improving structurization capabilities for overall task performance.

% Requires:
% \usepackage{booktabs}
% \usepackage{tabularx}
% \usepackage{array}
% \usepackage{xcolor}

% --- Domain badge background colors ---
\definecolor{CSBG}{RGB}{210,235,235}
\definecolor{LSBG}{RGB}{216,242,216}
\definecolor{SSBG}{RGB}{232,224,244}
\definecolor{ESBG}{RGB}{232,242,214}
\definecolor{EMBG}{RGB}{244,228,214}
\definecolor{PSBG}{RGB}{214,226,246}

% --- Domain badge text colors (a bit bolder, professional) ---
\definecolor{CSText}{RGB}{0,90,160}
\definecolor{LSText}{RGB}{0,120,90}
\definecolor{SSText}{RGB}{110,80,170}
\definecolor{ESText}{RGB}{140,120,0}
\definecolor{EMText}{RGB}{170,95,0}
\definecolor{PSText}{RGB}{40,70,140}

% --- Slim domain badge: [CS] [LS] ... ---
\newcommand{\domainbadge}[2]{%
\begingroup
\setlength{\fboxsep}{1.2pt}%
\colorbox{#1}{\textbf{#2}}%
\endgroup
}

% --- Compact colored ID tag like (CS1), (LS3), etc. ---
% Color matches the major domain accent (CSText, LSText, ...)
\newcommand{\areaid}[2]{\textcolor{#1}{\textbf{\texttt{(#2)}}}}

% --- Left aligned fixed width column type ---
\newcolumntype{L}[1]{>{\raggedright\arraybackslash}p{#1}}

\begin{table*}[t]
% \vspace{-5pt}
\centering
\scriptsize
\caption{\textbf{Major Science Domain Taxonomy (Compact).}
Abbreviations: \domainbadge{CSBG}{CS} Computer Science, \domainbadge{LSBG}{LS} Life Sciences,
\domainbadge{SSBG}{SS} Social Sciences, \domainbadge{ESBG}{ES} Environmental Sciences,
\domainbadge{EMBG}{EM} Economics \& Management Sciences, \domainbadge{PSBG}{PS} Physical Sciences.}
\label{tab:domain_taxonomy_two_col_badged_sleek}
% \vspace{-5pt}
\renewcommand{\arraystretch}{1.18}

\begin{tabularx}{\textwidth}{@{} L{4.5cm} X @{}}
\toprule
\textbf{Domain} & \textbf{Specific Area} \\
\midrule

\domainbadge{CSBG}{CS} \, Algorithm / AI &
\areaid{CSText}{CS1} Model Architecture
\areaid{CSText}{CS2} Training / Inference Pipeline
\areaid{CSText}{CS3} RAG / Agent System Component Diagram \\

\domainbadge{CSBG}{CS} \, System &
\areaid{CSText}{CS4} End to end Pipeline
\areaid{CSText}{CS5} System Component Architecture
\areaid{CSText}{CS6} Serving / Inference System \\

\domainbadge{CSBG}{CS} \, Computer Architecture &
\areaid{CSText}{CS7} Accelerator / Microarchitecture Block Diagram
\areaid{CSText}{CS8} Network on Chip / Interconnect Topology \\

\domainbadge{CSBG}{CS} \, Hardware &
\areaid{CSText}{CS9} EDA Toolchain / Design Flow Diagram \\
\midrule

\domainbadge{LSBG}{LS} \, Public Health \& Clinical Med. &
\areaid{LSText}{LS1} CONSORT / PRISMA Flow
\areaid{LSText}{LS2} Logic Model / Theory of Change (Health)
\areaid{LSText}{LS3} Causal DAG \\

\domainbadge{LSBG}{LS} \, Physiology &
\areaid{LSText}{LS4} Homeostatic Feedback Control Loop
\areaid{LSText}{LS5} Physiological Pathway / Axis Network \\

\domainbadge{LSBG}{LS} \, Cellular \& Molecular Bio. &
\areaid{LSText}{LS6} Signaling Pathway Schematic
\areaid{LSText}{LS7} Cell Fate / Lineage Tree \\
\midrule

\domainbadge{SSBG}{SS} \, Societal \& Institutional &
\areaid{SSText}{SS1} Institutional / Governance Framework
\areaid{SSText}{SS2} Institutional Decision / Policy Process Workflow \\

\domainbadge{SSBG}{SS} \, Individual Cognition &
\areaid{SSText}{SS3} Path Diagram / SEM / Mediation Model
\areaid{SSText}{SS4} Cognitive Architecture / Cycle Block Diagram \\
\midrule

\domainbadge{ESBG}{ES} \, Global Earth Systems &
\areaid{ESText}{ES1} Integrated Assessment Model / Nexus Modular Framework
\areaid{ESText}{ES2} Climate / Carbon Cycle Box Model \\

\domainbadge{ESBG}{ES} \, Infrastructure \& Energy &
\areaid{ESText}{ES3} Smart Grid / Microgrid Hierarchical Control Architecture \\
\midrule

\domainbadge{EMBG}{EM} \, Macroeconomics \& Policy &
\areaid{EMText}{EM1} DSGE / Macro Sector Agent Interaction Schematic
\areaid{EMText}{EM2} Logic Model / Theory of Change (Policy) \\

\domainbadge{EMBG}{EM} \, Market \& Corporate Eco. &
\areaid{EMText}{EM3} Ecosystem Map
\areaid{EMText}{EM4} Value Network / Stakeholder Exchange Map \\

\domainbadge{EMBG}{EM} \, Financial Instruments &
\areaid{EMText}{EM5} Securitization / Structured Finance Deal Structure
\areaid{EMText}{EM6} VaR / Risk Measurement Pipeline \\
\midrule

\domainbadge{PSBG}{PS} \, Physics \& Astronomy &
\areaid{PSText}{PS1} Observational / Experimental Data Processing Pipeline \\

\domainbadge{PSBG}{PS} \, Chemistry &
\areaid{PSText}{PS2} Catalytic Cycle / Mechanism State Graph \\

\domainbadge{PSBG}{PS} \, Materials Science &
\areaid{PSText}{PS3} Material Synthesis / Processing Route Schematic \\
\bottomrule
\end{tabularx}
% \vspace{-12pt}
\end{table*}

% Requires:
% \usepackage{booktabs}
% \usepackage{tabularx}
% \usepackage{multirow}
% \usepackage{array}
% \usepackage{xcolor}

% --- Major category colors (one per category) ---
\definecolor{FLColor}{RGB}{0,90,160} % Fault Localization accent
\definecolor{FMColor}{RGB}{0,120,90} % Functional Mapping accent
\definecolor{BTColor}{RGB}{170,95,0} % Boundary Testing accent
\definecolor{CRColor}{RGB}{110,80,170} % Counterfactual Reasoning accent

% --- Sleek separator between task types ---
\newcommand{\tasksep}{\hspace{0.55em}\textcolor{black!35}{\textbar}\hspace{0.55em}}

% --- Subtle inline task separator ---
\newcommand{\taskdot}{\hspace{0.55em}\textcolor{black!35}{$\bullet$}\hspace{0.55em}}

% --- Minimal ID style: colored mono text in parentheses ---
\newcommand{\taskid}[2]{\textcolor{#1}{\texttt{\textbf{(#2)}}}}

% --- Compact task item: (ID) description ---
\newcommand{\taskitem}[3]{\taskid{#1}{#2}\,#3}

% ======================================================
% Make wrapped lines align AFTER \tasksep + LEFT aligned text
% ======================================================
\newlength{\taskseplen}
\settowidth{\taskseplen}{\tasksep}

\newcommand{\taskcell}[1]{%
 #1\par
}

\begin{table*}[h]
\centering
\footnotesize
\caption{\textbf{Task Taxonomy of T2SBench.} Organized by \textit{Single choice}, \textit{Multiple choice}, and \textit{Mixed (Single+Multiple choice)}.}
\label{tab:template_taxonomy}

\newcolumntype{L}[1]{>{\raggedright\arraybackslash}p{#1}}
\renewcommand{\arraystretch}{0.95}
\setlength{\tabcolsep}{6pt}

\begin{tabularx}{\textwidth}{@{} L{1.0cm} X @{}}
\toprule

% =========================
% 1) Fault Localization
% =========================
\multicolumn{2}{@{}l@{}}{\textbf{1. Fault Localization}} \\
\midrule
\textit{Single} &
\taskcell{
\taskitem{FLColor}{FL2}{First directly affected downstream node},
\taskitem{FLColor}{FL4}{Key bottleneck / Dominator node},
\taskitem{FLColor}{FL6}{Branch isolation: who will not be affected},
\taskitem{FLColor}{FL7}{Fault amplification point in a feedback loop}
} \\

\textit{Multiple} &
\taskcell{
\taskitem{FLColor}{FL1}{Upstream root cause set},
\taskitem{FLColor}{FL3}{Minimum cut set},
\taskitem{FLColor}{FL8}{Multi fault explainability}
} \\

\textit{Mixed} &
\taskcell{
\taskitem{FLColor}{FL5}{Observe abnormal intermediate, infer upstream}
} \\
\midrule

% =========================
% 2) Functional Mapping
% =========================
\multicolumn{2}{@{}l@{}}{\textbf{2. Functional Mapping}} \\
\midrule
\textit{Single} &
\taskcell{
\taskitem{FMColor}{FM1}{Router / Selector identification},
\taskitem{FMColor}{FM2}{Aggregator / Fusion module identification},
\taskitem{FMColor}{FM3}{Buffering / Delay / Storage identification},
\taskitem{FMColor}{FM5}{Mediator vs. Direct cause},
} \\

\textit{Multiple} &
\taskcell{
\taskitem{FMColor}{FM4}{Controllers / Tuners identification},
\taskitem{FMColor}{FM7}{Parallel division of labor mapping}
} \\

\textit{Mixed} &
\taskcell{
\taskitem{FMColor}{FM6}{Measurement / Observation node identification}
} \\
\midrule

% =========================
% 3) Boundary Testing
% =========================
\multicolumn{2}{@{}l@{}}{\textbf{3. Boundary Testing}} \\
\midrule
\textit{Single} &
\taskcell{
\taskitem{BTColor}{BT1}{Conditional edge activation},
\taskitem{BTColor}{BT2}{Module with the narrowest applicability},
\taskitem{BTColor}{BT4}{Robustness under boundary conditions},
\taskitem{BTColor}{BT6}{Bypass leakage}
} \\

\textit{Multiple} &
\taskcell{
\taskitem{BTColor}{BT3}{Redundant path check},
\taskitem{BTColor}{BT5}{Invariance check}
} \\

\textit{Mixed} &
\taskcell{
\textemdash
} \\
\midrule

% =========================
% 4) Counterfactual Reasoning
% =========================
\multicolumn{2}{@{}l@{}}{\textbf{4. Counterfactual Reasoning}} \\
\midrule
\textit{Single} &
\taskcell{
\taskitem{CRColor}{CR1}{Edge removal: First downstream change}, 
\taskitem{CRColor}{CR2}{Module replacement}, 
\taskitem{CRColor}{CR4}{Disable feedback: Convergence change}, 
\taskitem{CRColor}{CR7}{Change upstream source: Unchanged downstream}
} \\

\textit{Multiple} &
\taskcell{
\taskitem{CRColor}{CR3}{Add a shortcut edge},
\taskitem{CRColor}{CR6}{Multi-point intervention: Direct vs. Indirect}
} \\

\textit{Mixed} &
\taskcell{
\taskitem{CRColor}{CR5}{Condition flip}
} \\

\bottomrule
\end{tabularx}
\end{table*}

In summary, similar to human cognition, models substantially benefit from structurized text information, leading to improved execution of downstream tasks. The benefits are significant and universally applicable, suggesting systematically assessing and enhancing models' structurization capabilities is both crucial and urgently needed.

% \vspace{-6pt}
\subsection{Challenges in Dataset Construction}
% \vspace{-5pt}
\label{sec23}

Although evaluating and training text structuring can substantially benefit models, building structures from text data faces three key challenges:
(1) \textbf{\textit{Difficult Correctness Verification:}} structuring long text is complex and time-consuming for both humans and models, making correctness verification expensive and often ambiguous.
(2) \textbf{\textit{Complex Evaluation:}} Text structures can be nested, cyclic, or disconnected; while nodes and links capture basic forms, defining a single universal standard to score generated structures is inherently difficult.
(3) \textbf{\textit{One-to-Many Structural Mapping:}} A single text may admit multiple equally valid structures (e.g., minor node edits or merges), so evaluation against a single reference structure is typically infeasible.

These challenges help explain why text structuring datasets remain scarce. 
In this work, we address all three challenges through carefully designed construction pipeline and introduce T2S-Bench. As summarized in Tab. \ref{tab:prior_benchmark_comparison}, T2S-Bench is the first dataset to comprehensively evaluate models’ text structuring capability, while also providing practical guidance and insights for future text-to-structure benchmarks.

% \vspace{-8pt}
\section{Construction Process of T2S-Bench}
% \vspace{-7pt}
In this section, we introduce construction process of T2S-Bench. The overall construction flow is shown in Fig. \ref{fig_2}. All prompts used are detailed in Appendix \ref{app:B}.

% \vspace{-7pt}
\subsection{Sample Collection}
% \vspace{-7pt}
\textbf{Academic Paper-Based Data Source.} To address the challenge of verifying structural correctness, T2S-Bench utilizes academic papers and their structural diagrams as primary data sources. Academic papers offer two significant advantages: (1) \textit{High Structural Accuracy}: Diagrams in academic papers are meticulously designed by authors and rigorously validated by reviewers, especially for well-cited articles, ensuring their structural accuracy and completeness. (2) \textit{Strong Textual Structure}: Text segments corresponding to diagrams in academic papers inherently possess high structural coherence and logical clarity, allowing readers to infer diagrammatic relationships directly from the text.

By leveraging the professionalism and correctness inherent in academic paper diagrams and the structural clarity of corresponding texts, T2S-Bench averts hallucinations from model-generated structures and significantly reduces human verification efforts. Hence, our initial goal is to collect high-quality text-structure pairs from academic paper.

\textbf{Science Topic \& Structure Type Design.} To ensure dataset diversity, as detailed in Tab. \ref{tab:domain_taxonomy_two_col_badged_sleek}, T2S-Bench includes 6 main scientific disciplines (Computer Science, Life Sciences, Social Sciences, Environmental Sciences, Economics \& Management Sciences, Physical Sciences), 17 sub-disciplines, and 32 structural types. These structural types represent commonly used diagrams within specific sub-disciplines.

% Are nodes and links clearly identified?

% Do all nodes correspond to clear entities?

% Do all links clearly define source and target nodes?

% Is the diagram free of noise?

% Is the node count appropriate (between 5 and 20)?

% Is the link count appropriate (between 5 and 40)?

% Are there duplicate node names?

% Are nodes clearly represented by phrases or multiple phrases?

% Is the diagram singular and complete, without nested or independent structures?

% Icons
\newcommand{\cmark}{\textcolor{green!55!black}{\ding{51}}}
\newcommand{\xmark}{\textcolor{red!70!black}{\ding{55}}}
\newcommand{\pmark}{\textcolor{orange!85!black}{\ding{108}}} % partial

% Cell helper
\newcommand{\smallcell}[1]{\makecell[l]{#1}}
\renewcommand{\arraystretch}{1.18}

% Requires
% \usepackage{booktabs}
% \usepackage{adjustbox}
% \usepackage{xcolor}
% \usepackage{colortbl}
% \usepackage{array}

% Symbols (if you already have these, remove duplicates)
% \newcommand{\cmark}{\ding{51}} % requires pifont
% \newcommand{\xmark}{\ding{55}} % requires pifont
% \newcommand{\smallcell}[1]{\begin{tabular}[c]{@{}l@{}}#1\end{tabular}}

\begin{table*}[t]
\centering
\small
\setlength{\tabcolsep}{5.5pt}
\rowcolors{2}{gray!6}{white}
% \vspace{-8pt}
\caption{\textbf{Comparison of long-context, evidence-grounded, and structured reasoning benchmarks.}
We summarize each dataset by its main evaluation emphasis, input and output format, data source, and whether (i) Uses high-quality real-world data (\textit{HQ Data}), (ii) Evaluation metric is Logic verifiable (\textit{Metric Verif.}), and (iii) Evaluates text structuring ability (\textit{Text Struct.}).}
\label{tab:prior_benchmark_comparison}
% \vspace{-5pt}
\begin{adjustbox}{max width=\textwidth}
\begin{tabular}{@{}l l l l l c c c@{}}
\toprule
\textbf{Dataset} &
\textbf{Primary Focus} &
\textbf{Input} &
\textbf{Output} &
\textbf{Dataset Source} &
\textbf{HQ Data} &
\textbf{Metric Verif.} &
\textbf{Text Struct.} \\
\midrule

LongBench v2~\cite{bai2025longbenchv2} &
Long-context QA &
Long documents &
MCA &
Expert-verified documents &
\cmark &
\xmark &
\xmark \\

Qasper~\cite{dasigi2021dataset} &
Evidence-based QA &
Long documents &
Mixed Answer &
NLP papers (arXiv) &
\cmark &
\xmark &
\xmark \\

LongBench Pro~\cite{chen2026longbenchpro} &
Long-context reasoning &
Long documents &
Answer/Summary &
Public web documents &
\xmark &
\xmark &
\xmark \\

QMSum~\cite{zhong2021qmsum} &
Query-focused summarization &
Meeting transcripts &
Summary &
Real-world meetings &
\cmark &
\xmark &
\xmark \\

HotpotQA~\cite{yang2018hotpotqa} &
Multi-hop QA &
Long documents &
Mixed Answer &
Wikipedia &
\xmark &
\xmark &
\xmark \\

StructEval~\cite{yang2026structeval} &
Structured language reasoning &
Structured language &
Structured language &
Synthetic &
\xmark &
\cmark &
\xmark \\

StructBench~\cite{gu2024structbench} &
Structured language reasoning &
Structured language &
Answer string &
Synthetic &
\xmark &
\xmark &
\xmark \\

HiBench~\cite{jiang2025hibench} &
Hierarchical structure reasoning &
Textualized structures &
Answer string &
Synthetic + Real-world &
\xmark &
\cmark &
\xmark \\

\rowcolor{blue!7}
\textbf{T2S Bench (Ours)} &
\textbf{Semantic structure reasoning} &
\textbf{Context paragraphs} &
\textbf{MCA/Structures} &
\textbf{Research papers} &
\cmark &
\cmark &
\cmark \\

\bottomrule
\end{tabular}
\end{adjustbox}
\end{table*}

\begin{figure*}[t]
% \vspace{-11pt}
 \centering
 \begin{subfigure}[t]{0.71\linewidth}
 \centering
 \includegraphics[width=\linewidth]{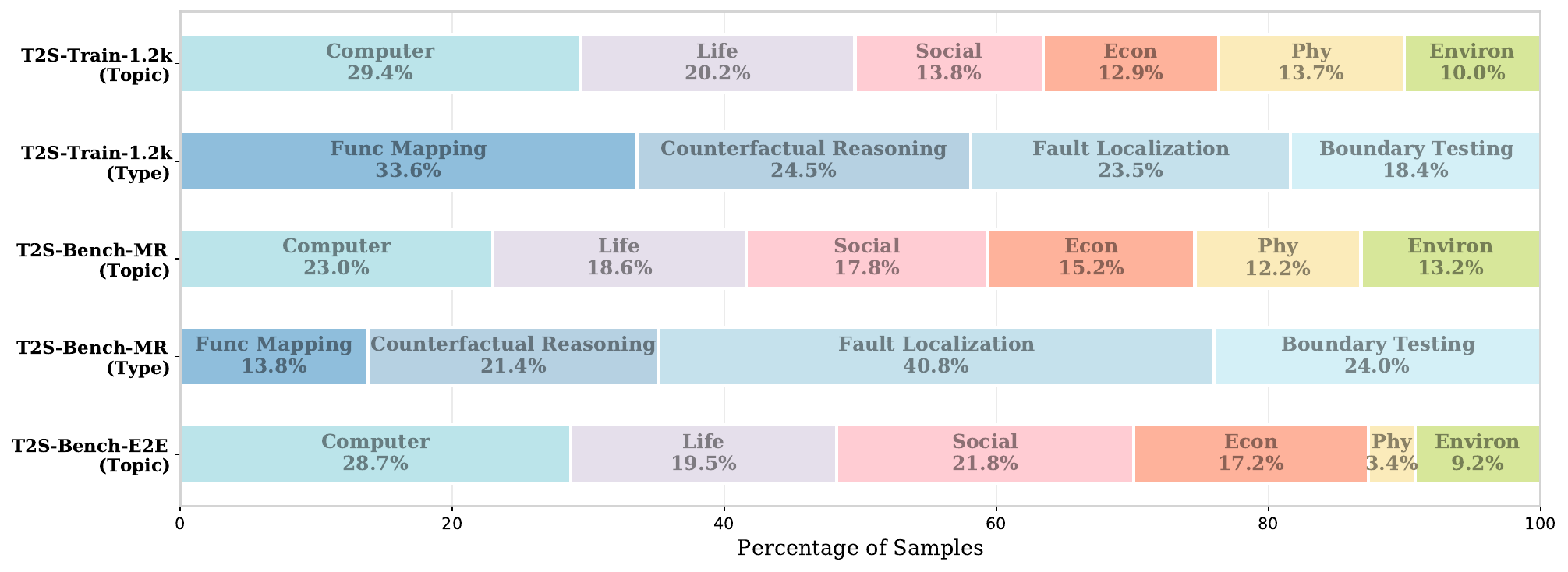}
 % \caption{Left}
 \label{fig:left}
 \end{subfigure}\hfill
 % 右边：窄一点
 \begin{subfigure}[t]{0.26\linewidth}
 \centering
 \includegraphics[width=\linewidth]{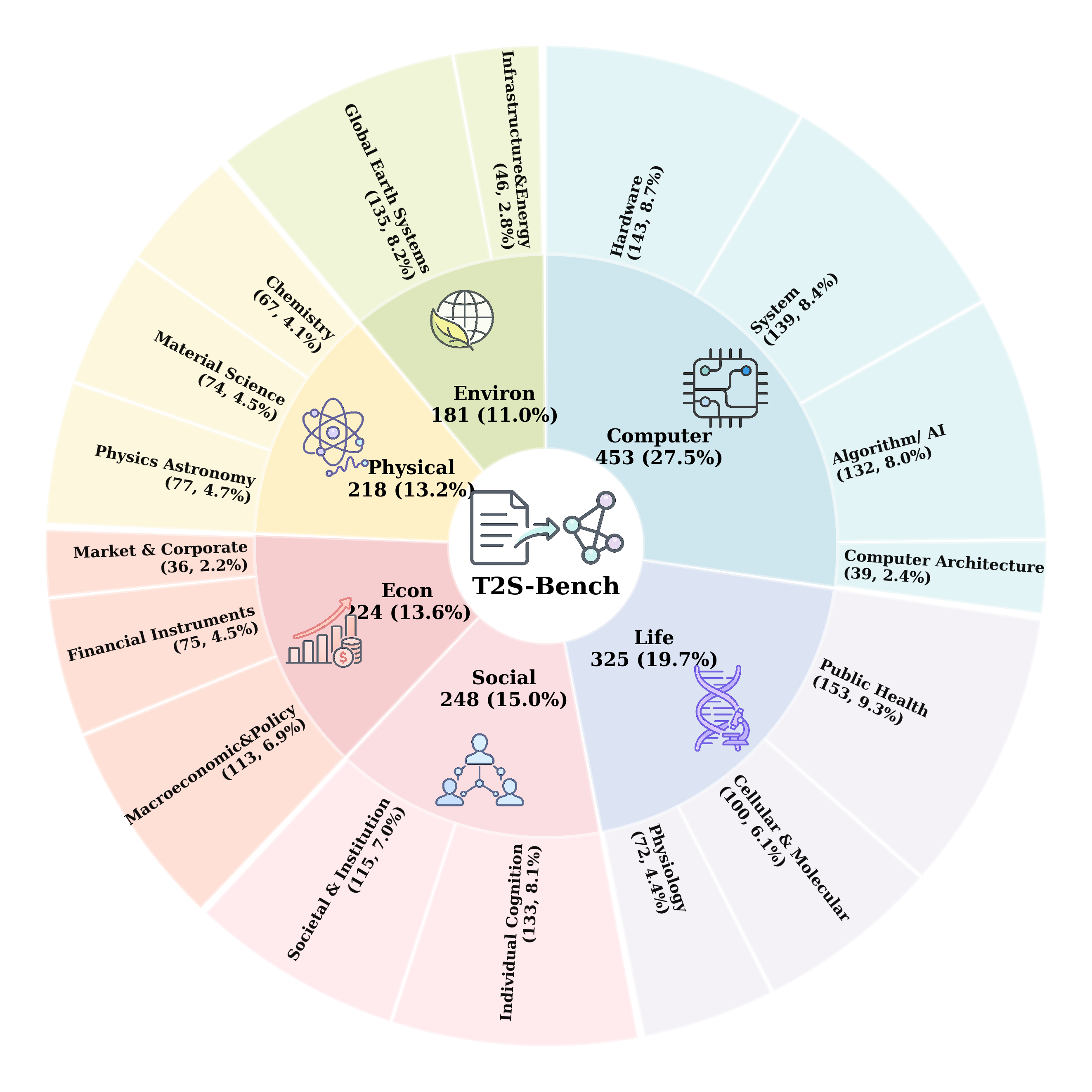}
 % \caption{Right}
 \label{fig:right}
 \end{subfigure}
% \vspace{-17pt}
 \caption{(Left) \textbf{Sample Distributions of different Dataset.} (Right) \textbf{Overview of T2S-Bench Sample Distributions.}}
% \vspace{-14pt}
 \label{fig_3}
\end{figure*}

% logo cell, vertically centered inside a multirow group
% Updated command: \cellcolor comes before \multirow
\newcommand{\GroupLogo}[2]{%
  \cellcolor{white}% <--- Moved here
  \multirow{#1}{*}{%
    \raisebox{-0.2\height}{\includegraphics[height=1.4em]{fig/logo/#2}}%
  }%
}

\begin{table*}[!t]
% \vspace{-5pt}
\caption{\textbf{Overall performance on T2S-Bench.} The leftmost two columns show overall performance (EM and F1 scores) on T2S-Bench-MR, multiple-choice dataset requiring multi-hop reasoning. The central columns represent accuracy for each question type within T2S-Bench-MR. The rightmost two columns show performance on the T2S-Bench-E2E dataset, separately evaluating node extraction (measured by average semantic similarity between predicted and reference nodes) and link extraction (measured by F1 score of correctly identified link pairs). The best-performing model within each model family on T2S-Bench-MR is highlighted with pink shading. Additionally, the top three performances in each metric are highlighted in red (first place), green (second place), and blue (third place).}
% \vspace{-5pt}
\centering
\renewcommand{\arraystretch}{1.08}
\setlength{\tabcolsep}{4pt}
\resizebox{\textwidth}{!}{
\begin{tabular}{c l||cc|cc|cc|cc|cc||cc}
\toprule[2pt]
 &  & \multicolumn{2}{>{\columncolor{gray!12}}c|}{\textbf{Multi-choice QA Overall}}
 & \multicolumn{2}{c|}{\textbf{Boundary Testing}}
 & \multicolumn{2}{c|}{\textbf{Counterfactual Reasoning}}
 & \multicolumn{2}{c|}{\textbf{Fault Localization}}
 & \multicolumn{2}{c||}{\textbf{Functional Mapping}}
 & \multicolumn{2}{>{\columncolor{gray!12}}c}{\textbf{Structure Score}} \\
\cmidrule(lr){3-4}\cmidrule(lr){5-6}\cmidrule(lr){7-8}\cmidrule(lr){9-10}\cmidrule(lr){11-12}\cmidrule(lr){13-14}
\textbf{} & \textbf{Model} & \textbf{EM} & \textbf{F1} & \textbf{EM} & \textbf{F1} & \textbf{EM} & \textbf{F1} & \textbf{EM} & \textbf{F1} & \textbf{EM} & \textbf{F1} & \textbf{Node} & \textbf{Link}\\
\midrule[1pt]

\rowcolor{colblue}
\GroupLogo{4}{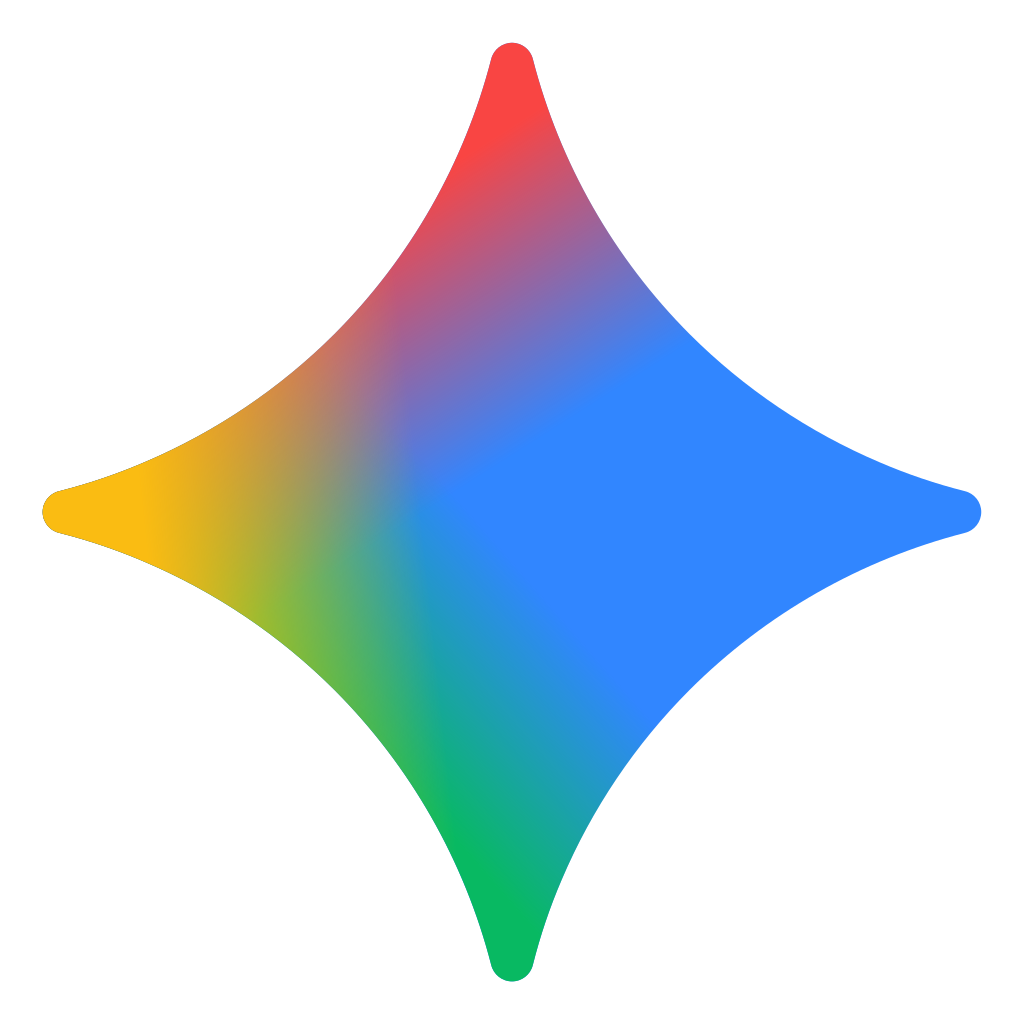} & \textbf{Gemini-2.5-Pro} & \best{81.40} & \best{91.56} & \third{80.00} & \third{91.78} & \best{90.65} & \third{93.46} & \best{74.51} & \best{90.62} & \best{89.86} & \best{91.01} & \best{58.09} & \second{84.32}\\
 & \textbf{Gemini-2.5-Flash} & 72.20 & 83.67 & 77.50 & 87.33 & 78.50 & 84.33 & 64.22 & \second{82.76} & 76.81 & 78.94 & 46.90 & 75.10\\
 & \textbf{Gemini-2.0-flash} & 61.20 & 79.63 & 72.50 & 87.03 & 75.70 & 86.76 & 41.67 & 70.99 & 76.81 & 81.26 & 42.71 & 66.42\\
 & \textbf{Gemini-2.0-flash-lite} & 53.40 & 72.41 & 65.00 & 81.08 & 79.44 & 83.74 & 29.90 & 63.35 & 62.32 & 66.57 & 39.22 & 69.18\\
\midrule

\rowcolor{colblue}
\GroupLogo{6}{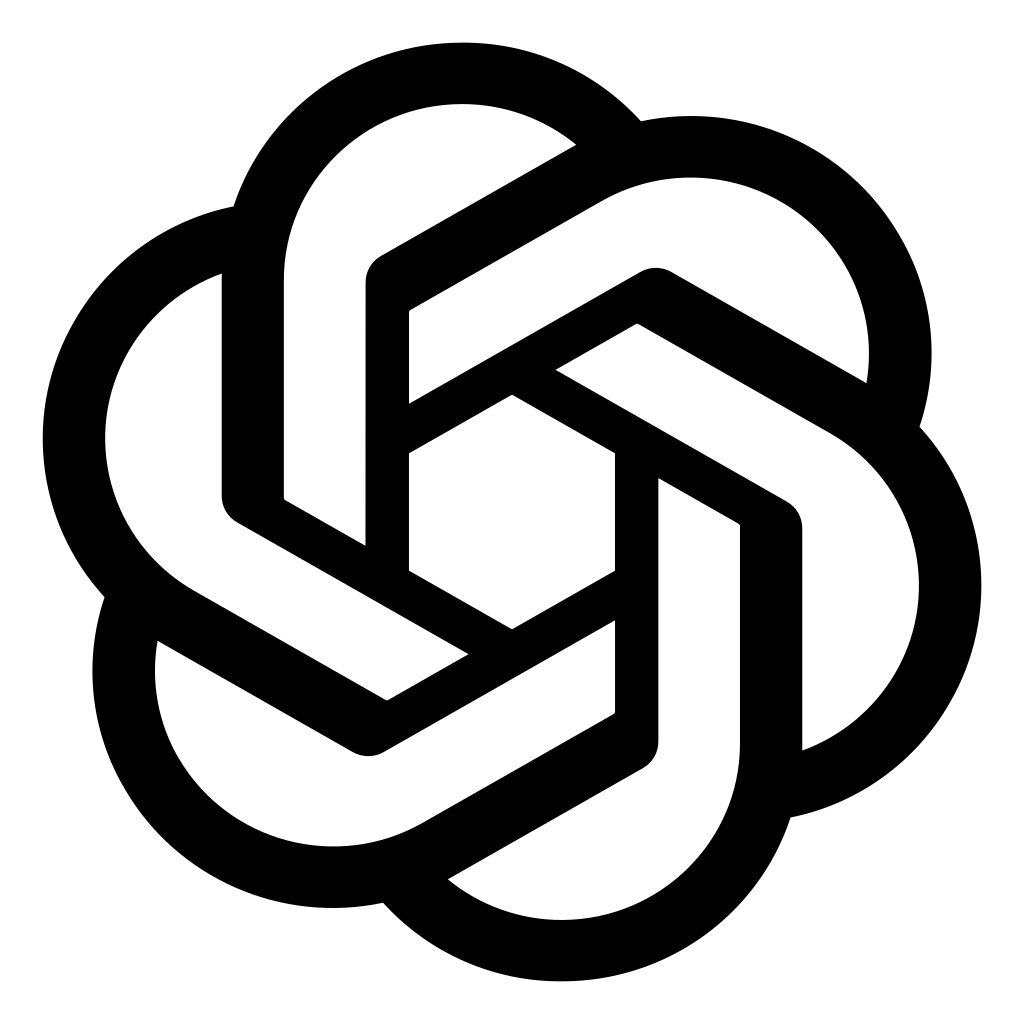} & \textbf{GPT-5.2} & 71.80 & 84.32 & 77.50 & 90.47 & 84.11 & 89.69 & 62.75 & 82.24 & 69.57 & 71.45 & 50.57 & 77.76\\
 & \textbf{GPT-5.1} & 67.80 & 83.46 & 75.83 & 89.75 & 83.18 & 89.50 & 51.47 & 77.95 & 78.26 & 79.42 & 45.36 & 79.44\\
 & \textbf{GPT-4.1-mini} & 59.00 & 76.83 & 66.67 & 84.19 & 68.22 & 82.55 & 48.04 & 72.37 & 63.77 & 68.31 & 45.55 & 74.72\\
 & \textbf{GPT-3.5-turbo} & 25.40 & 48.63 & 29.17 & 47.33 & 31.78 & 49.00 & 14.71 & 50.07 & 40.58 & 46.09 & 32.71 & 57.84\\
 & \textbf{GPT-4o} & 61.80 & 79.24 & 73.33 & 89.31 & 79.44 & 86.64 & 44.12 & 72.80 & 66.67 & 69.32 & 40.51 & 74.29\\
 & \textbf{GPT-4o-mini} & 20.40 & 54.19 & 15.83 & 58.58 & 20.56 & 51.96 & 23.04 & 61.09 & 20.29 & 29.61 & 39.83 & 66.61\\
\midrule

\rowcolor{colblue}
\GroupLogo{4}{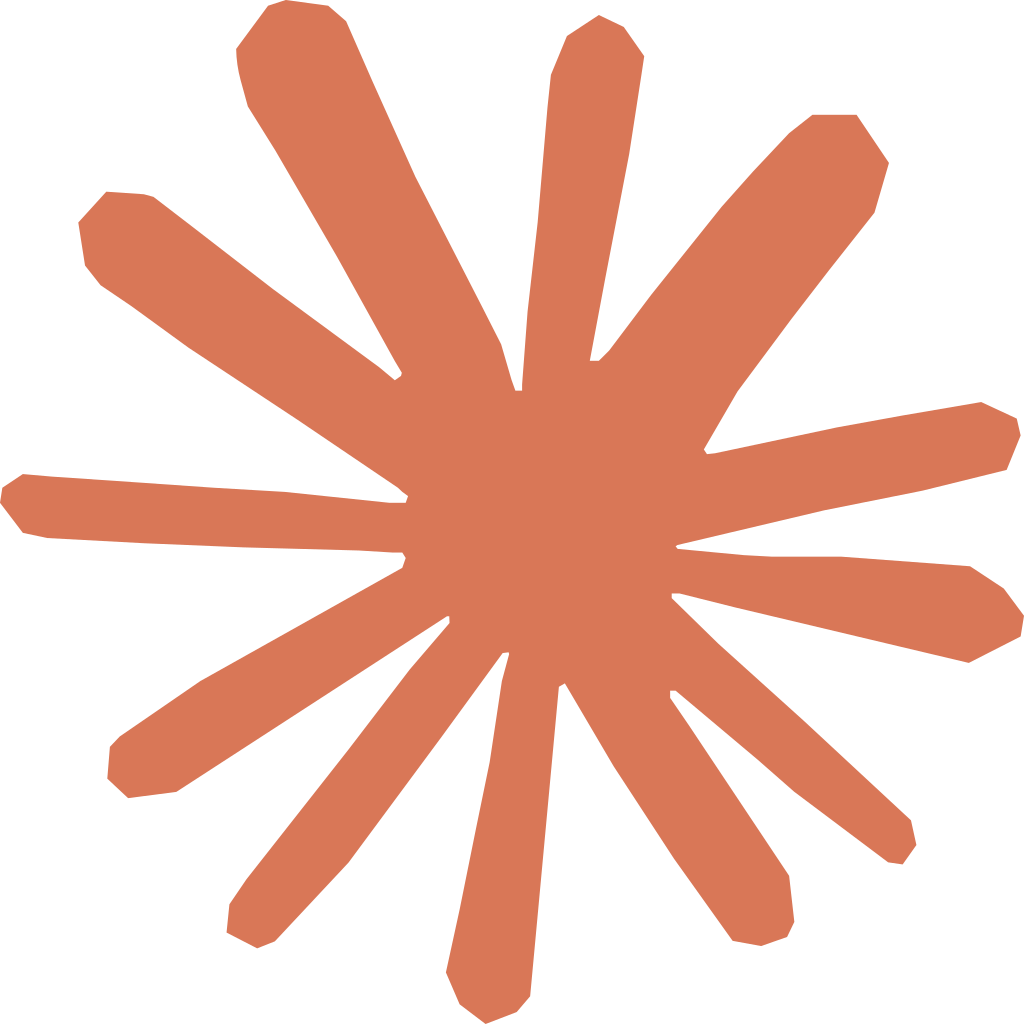} & \textbf{Claude-sonnet-4-5-20250929} & \second{76.80} & \third{86.85} & \best{84.17} & \second{92.14} & \third{88.79} & 92.49 & \third{65.20} & \third{82.71} & 79.71 & 81.16 & \second{55.97} & \best{86.91}\\
 & \textbf{Claude-haiku-4-5-20251001} & 67.40 & 80.87 & 78.33 & 90.33 & 85.98 & 89.07 & 49.51 & 73.35 & 72.46 & 73.91 & 47.06 & 79.33\\
 & \textbf{Claude-4-Sonnet-20250514} & 75.60 & 86.63 & \best{84.17} & \best{92.47} & \second{89.72} & \best{94.02} & 61.27 & 80.68 & \third{81.16} & \third{82.61} & \third{54.11} & \third{84.07}\\
 & \textbf{Claude-3-haiku-20240307} & 44.80 & 68.77 & 39.17 & 71.47 & 59.81 & 71.50 & 37.25 & 69.11 & 53.62 & 58.84 & 39.18 & 75.51\\
\midrule

\GroupLogo{6}{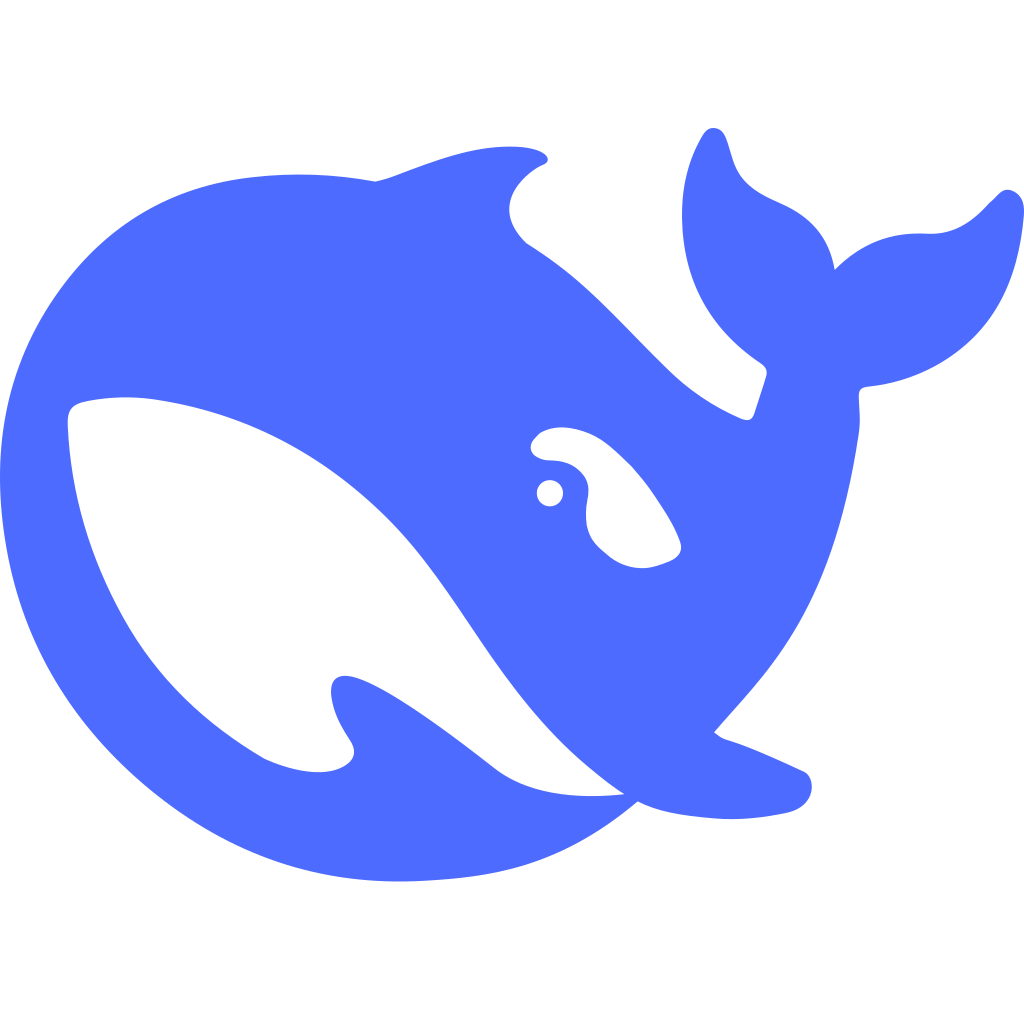} & \textbf{DeepSeek-V3.2} & 60.00 & 78.31 & 66.67 & 82.72 & 78.50 & 87.85 & 42.16 & 71.73 & 72.46 & 75.31 & 46.98 & 78.69\\
 & \textbf{DeepSeek-V3.1} & 60.40 & 78.59 & 66.67 & 82.72 & 78.50 & 87.85 & 42.65 & 72.07 & 73.91 & 76.33 & 46.59 & 77.77\\
 & \textbf{DeepSeek-R1-0528} & 57.00 & 63.76 & 58.33 & 61.08 & 65.42 & 68.50 & 46.57 & 59.11 & 72.46 & 74.78 & 49.24 & 80.31\\
\rowcolor{colblue}  \cellcolor{white} & \textbf{DeepSeek-reasoner(R1)} & \third{76.60} & \second{87.20} & \second{83.33} & 91.58 & 84.11 & \second{91.37} & \second{65.69} & 82.62 & \second{85.51} & \second{86.67} & 52.25 & 80.67\\
 & \textbf{DeepSeek-chat} & 60.20 & 78.37 & 70.00 & 83.86 & 78.50 & 88.03 & 40.69 & 71.26 & 72.46 & 74.88 & 47.58 & 78.97\\
 & \textbf{DeepSeek-V3-0324} & 60.60 & 78.75 & 65.83 & 82.31 & 78.50 & 87.85 & 43.63 & 72.71 & 73.91 & 76.33 & 47.32 & 78.17\\
\midrule

\GroupLogo{9}{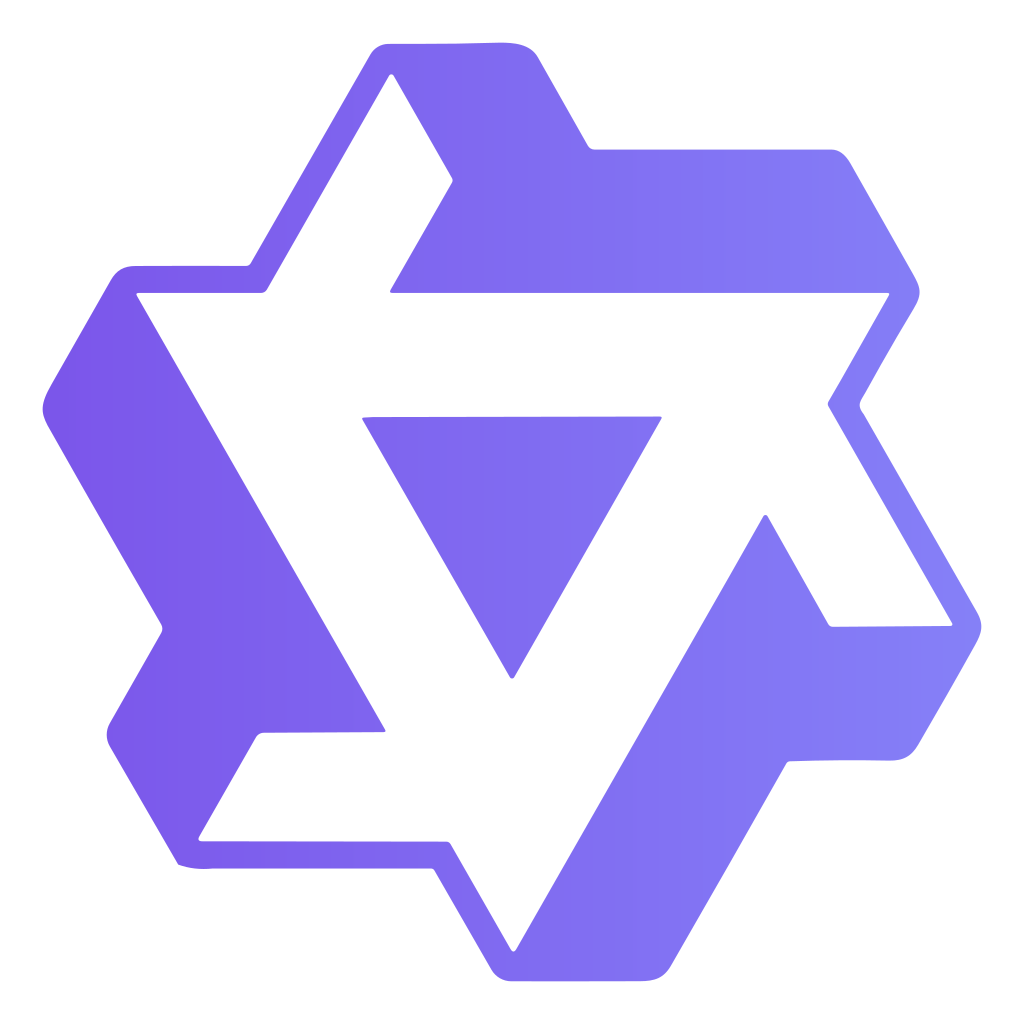} & \textbf{Qwen3-235B-A22B-Thinking-2507} & 60.80 & 79.89 & 60.00 & 82.19 & 74.77 & 85.42 & 50.00 & 76.95 & 72.46 & 76.04 & 45.97 & 76.11\\
 & \textbf{Qwen3-235B-A22B-Instruct-2507} & 62.80 & 80.25 & 61.67 & 82.72 & 78.50 & 87.54 & 50.00 & 74.61 & 78.26 & 81.30 & 49.39 & 73.54\\
 & \textbf{Qwen3-Next-80B-A3B-Thinking} & 24.40 & 36.21 & 25.83 & 36.67 & 28.97 & 37.23 & 20.59 & 37.65 & 26.09 & 29.57 & 42.58 & 77.64\\
 & \textbf{Qwen3-Next-80B-A3B-Instruct} & 46.60 & 57.00 & 57.50 & 67.64 & 59.81 & 66.60 & 28.92 & 44.24 & 59.42 & 61.30 & 45.67 & 76.11\\
 & \textbf{Qwen3-30B-A3B-Thinking-2507} & 43.20 & 64.82 & 45.00 & 64.17 & 57.01 & 75.32 & 32.84 & 63.22 & 49.28 & 54.44 & 39.51 & 71.90\\
 & \textbf{Qwen3-30B-A3B-Instruct-2507} & 47.40 & 72.15 & 46.67 & 75.17 & 65.42 & 81.84 & 36.27 & 68.58 & 53.62 & 62.46 & 45.19 & 72.16\\
\rowcolor{colblue} \cellcolor{white} & \textbf{Qwen3-32B} & 69.40 & 83.41 & 78.33 & 89.72 & \second{89.72} & \second{93.60} & 53.43 & 78.24 & 69.57 & 71.88 & 43.35 & 73.94\\
 & \textbf{Qwen3-14B} & 47.40 & 70.38 & 51.67 & 76.92 & 60.75 & 78.38 & 34.80 & 66.43 & 56.52 & 58.26 & 46.85 & 71.54\\
 & \textbf{Qwen3-8B} & 47.60 & 66.47 & 51.67 & 69.33 & 59.81 & 71.84 & 38.24 & 67.07 & 49.28 & 51.40 & 43.03 & 72.03\\
\midrule

\rowcolor{colblue}
\GroupLogo{2}{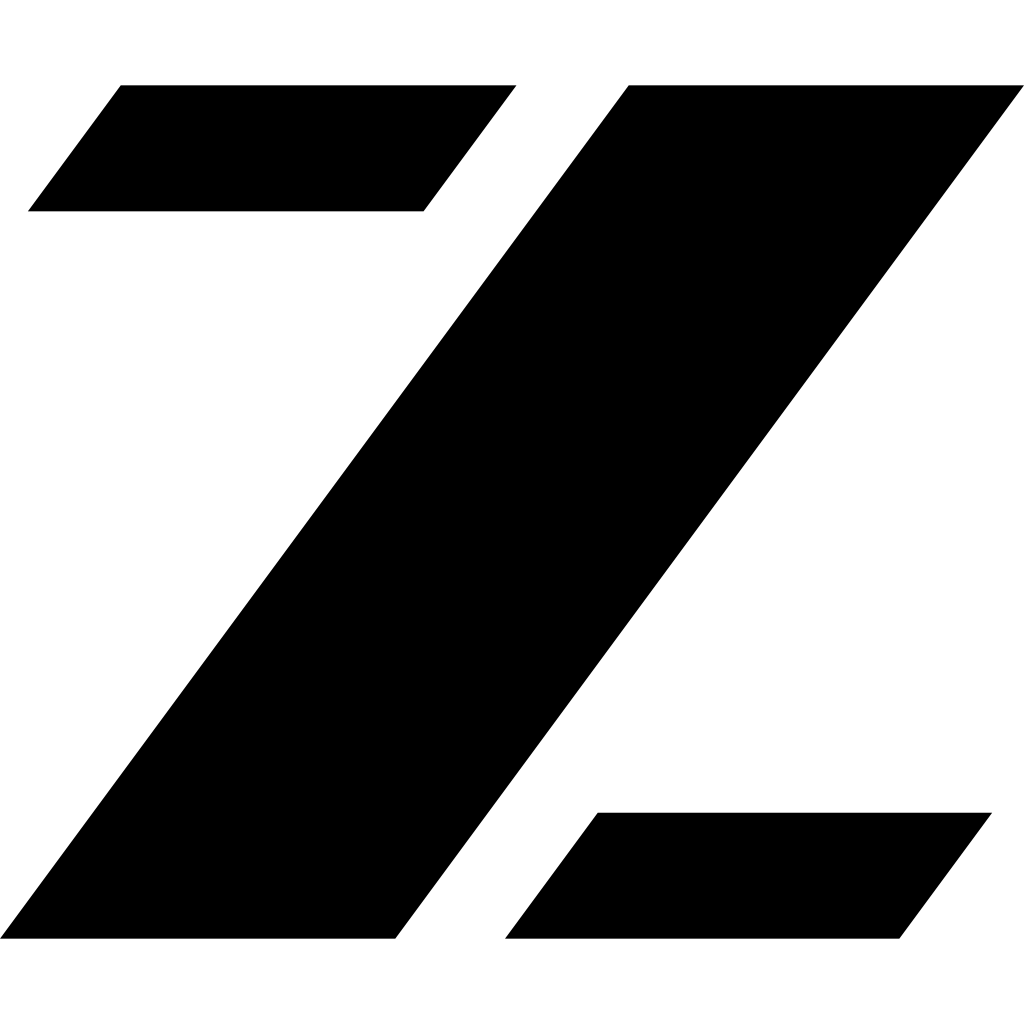} & \textbf{GLM-4.5} & 37.00 & 43.92 & 27.50 & 33.42 & 43.93 & 50.78 & 28.43 & 37.36 & 68.12 & 70.97 & 9.33 & 11.44\\
 & \textbf{GLM-4.6} & 28.80 & 33.36 & 20.00 & 22.33 & 25.23 & 27.73 & 26.96 & 35.07 & 55.07 & 56.23 & 11.84 & 11.12\\
\midrule

\GroupLogo{1}{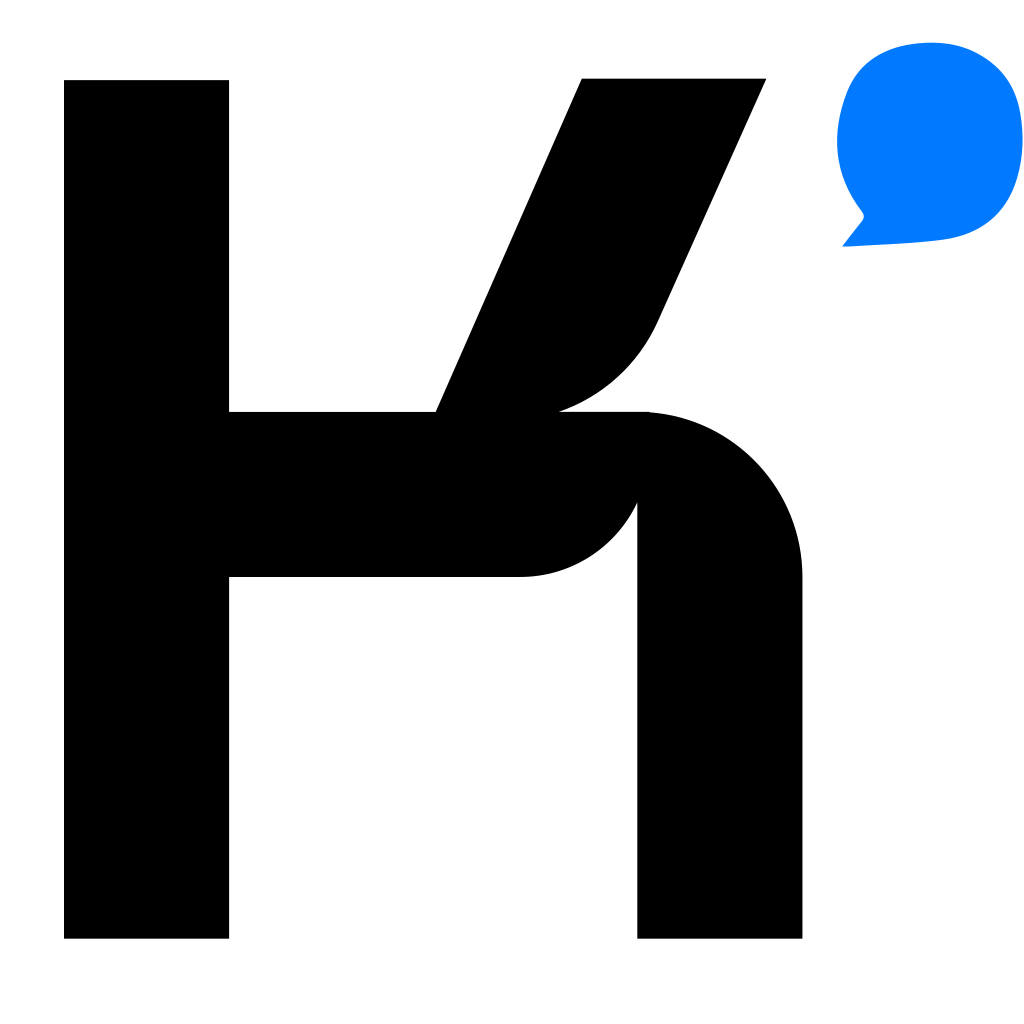} & \textbf{Kimi-K2-Instruct-0905} & 67.00 & 81.00 & 75.83 & 88.58 & 84.11 & 88.72 & 50.98 & 74.74 & 72.46 & 74.35 & 44.68 & 61.77\\
\midrule

\GroupLogo{2}{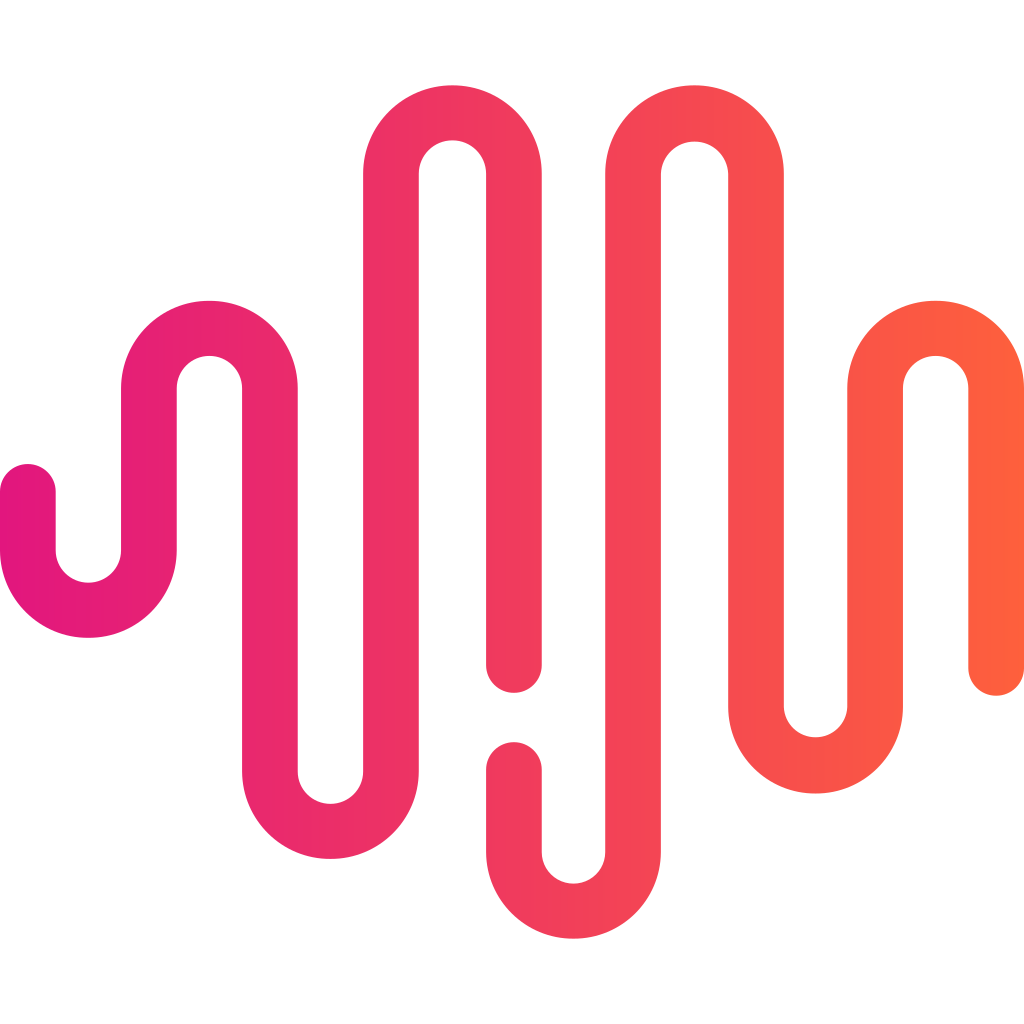} & \textbf{MiniMax-M2} & 4.20 & 4.83 & 3.33 & 4.17 & 6.54 & 7.17 & 1.96 & 2.68 & 8.70 & 8.70 & 2.52 & 2.94\\
 & \textbf{MiniMax-Text-01} & 55.80 & 74.69 & 66.67 & 80.36 & 72.90 & 81.87 & 35.78 & 68.50 & 69.57 & 71.98 & 41.88 & 73.29\\
\midrule

\GroupLogo{6}{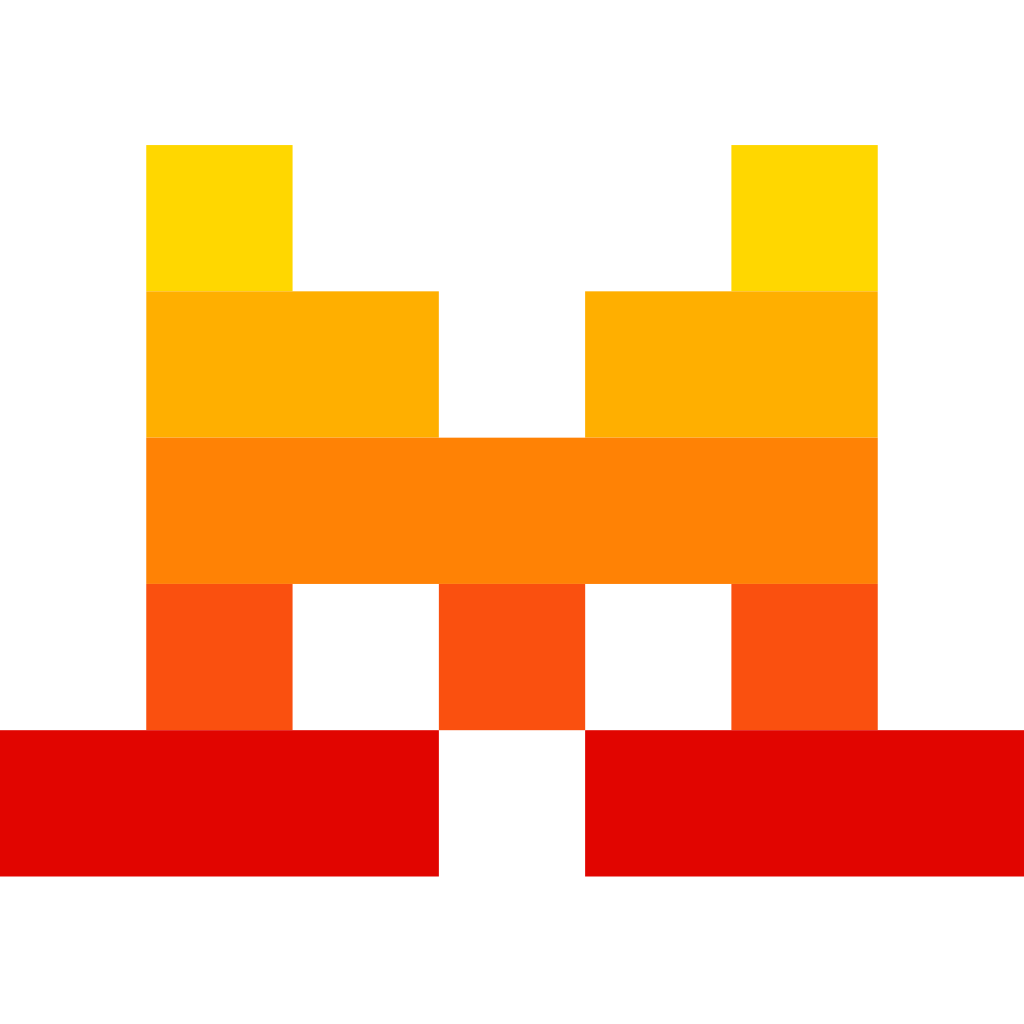} & \textbf{Ministral-3-3B-Instruct-2512} & 39.20 & 62.29 & 43.33 & 67.36 & 51.40 & 66.88 & 28.43 & 60.69 & 44.93 & 51.06 & 25.32 & 56.79\\
 & \textbf{Ministral-3-8B-Instruct-2512} & 52.40 & 71.37 & 59.17 & 75.11 & 71.96 & 78.26 & 34.31 & 67.18 & 63.77 & 66.62 & 36.70 & 62.29\\
 & \textbf{Ministral-3-14B-Instruct-2512} & 55.80 & 75.68 & 59.17 & 81.47 & 77.57 & 85.14 & 39.22 & 69.10 & 65.22 & 70.39 & 36.94 & 62.93\\
 & \textbf{Ministral-8B-Instruct-2410} & 17.40 & 45.09 & 14.17 & 49.31 & 17.76 & 40.31 & 22.55 & 55.38 & 7.25 & 14.78 & 32.47 & 59.82\\
 & \textbf{Mistral-Large-Instruct-2411} & 56.60 & 74.75 & 65.00 & 83.47 & 70.09 & 77.10 & 42.65 & 71.70 & 62.32 & 64.98 & 45.71 & 71.18\\
\rowcolor{colblue} & \textbf{Mistral-Small-3.2-24B-Instruct-2506} & 56.80 & 75.48 & 65.00 & 82.56 & 80.37 & 87.07 & 36.76 & 67.66 & 65.22 & 68.31 & 45.74 & 67.41\\
\midrule

\GroupLogo{5}{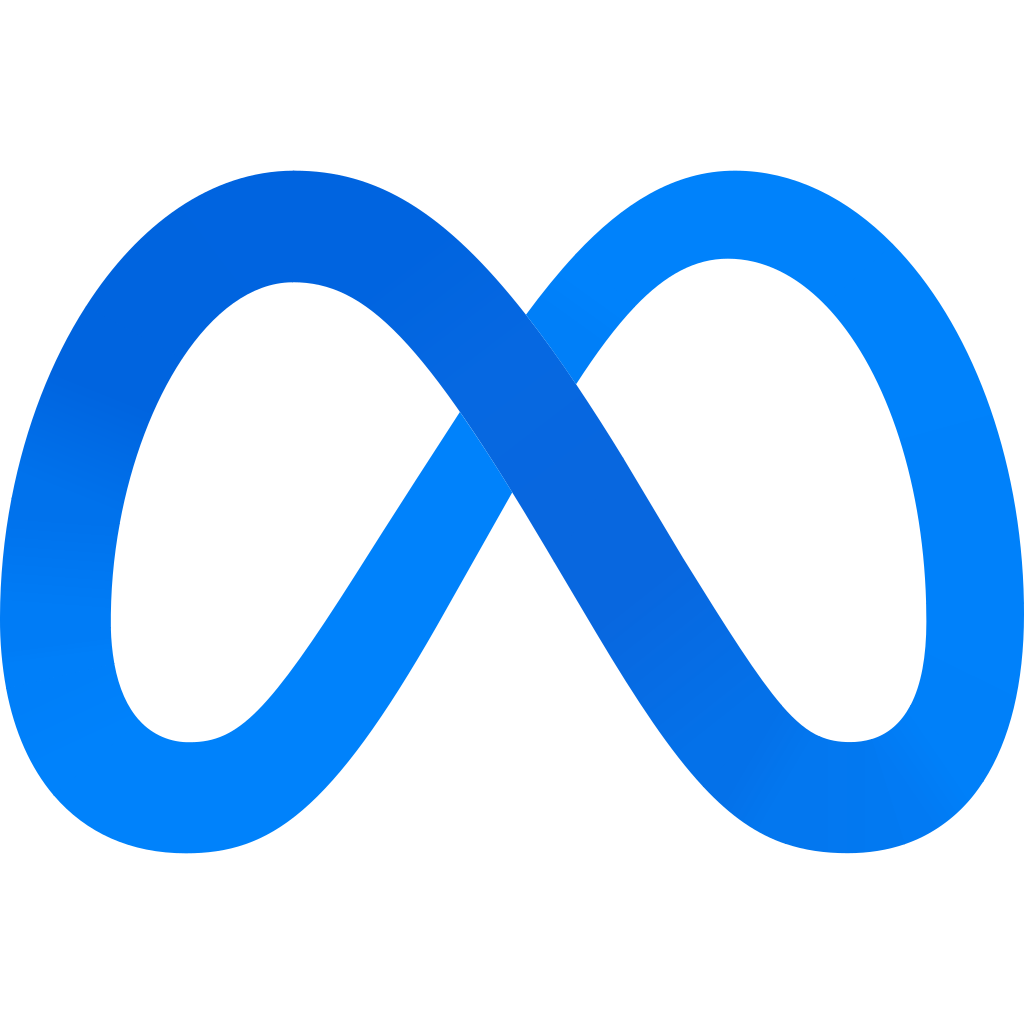} & \textbf{Llama-3.1-8B-Instruct} & 24.20 & 54.22 & 19.17 & 53.36 & 24.30 & 55.79 & 24.02 & 58.97 & 33.33 & 39.23 & 35.77 & 44.38\\
\rowcolor{colblue} \cellcolor{white} & \textbf{Llama-3.1-70B-Instruct} & 56.60 & 74.90 & 65.83 & 81.56 & 77.57 & 84.89 & 37.75 & 68.15 & 63.77 & 67.78 & 43.77 & 56.13\\
 & \textbf{Llama-3.1-405B-Instruct} & 50.40 & 60.85 & 58.33 & 66.81 & 53.27 & 57.23 & 41.18 & 59.08 & 59.42 & 61.30 & 41.51 & 59.18\\
 & \textbf{Llama-3.2-3B-Instruct} & 24.60 & 52.34 & 20.83 & 51.58 & 24.30 & 46.85 & 24.51 & 59.37 & 31.88 & 41.35 & 31.53 & 39.25\\
 & \textbf{Llama-3.3-70B-Instruct} & 54.00 & 73.10 & 59.17 & 78.81 & 72.90 & 82.55 & 40.20 & 68.80 & 56.52 & 61.26 & 40.60 & 50.99\\

\bottomrule[2pt]
\end{tabular}}
% \vspace{-14pt}
\label{tab:multiqa_category}
\end{table*}

\textbf{Model Search \& Check.} To effectively generate valid paper-structure pairs, we implement a rigorous four-module automated pipeline: (i) \textit{Paper Search}: We leverage GPT-5.2's~\cite{singh2025openai} search capability to identify papers containing structural diagrams. To improve precision and figure quality, each search is constrained to a specific subfield and structure type.
(ii) \textit{PDF Download \& Figure Cropping}: Selected papers are automatically downloaded using Python scripts. Figures are extracted via pdffigures2~\cite{pdffigures2}, and validated by GPT-4o~\cite{openai2024gpt4ocard} to confirm structural relevance. (iii) \textit{Structural Validity Check}: Cropped figures undergo validity checks using Gemini-2.5-Pro~\cite{comanici2025gemini}, ensuring diagrams can be represented as JSON structures containing nodes and links. (iv) \textit{Text Extraction \& Validity Check}: Verified figures and PDFs are cross-checked by GPT-o3 and Gemini-2.5-Pro to ensure at least three text segments correspond clearly with each structural diagram, generating coherent textual samples with clearly identified start and end sentences.

Failure at any stage triggers a restart from step (i), with only fully validated samples included.
% We performed extensive trials with multiple check model combinations, manually verifying quality and selecting the best-performing combination for final batch processing. 
We target 50 high-quality samples per structural type; ultimately, 1521 qualified text-structure pairs were obtained after approximately five searches per accepted sample.

\textbf{First-Round Human Filter.} Despite model verification, diverse presentation formats in academic diagrams often introduce significant noise (e.g., images, explanatory text, symbols). 
Therefore, we invited 11 PhD-level experts from various domains to perform manual quality checks on samples relevant to their respective fields. Each expert followed a checklist assessing structural completeness, noise presence, node and link counts, and structural singularity. The complete checklist is provided in Appendix \ref{app:B4}.
Only samples meeting all checklist criteria were labeled as high-quality. The human review took one week and resulted in 672 rigorously vetted, high-quality text-structure pairs, forming the basis of our benchmark dataset.

% \vspace{-7pt}
\subsection{T2S Multi-hop Reasoning Dataset Construction}
% \vspace{-5pt}

After collecting high-quality text–structure pairs, we construct the T2S Multi-hop Reasoning dataset. To overcome the challenge of evaluating text structuring, we use multiple-choice questions grounded in paper reference diagrams that require multi-node, multi-step reasoning. Each question must (1) Necessitate text structuring and multi-step reasoning; (2) Explicitly depend on the reference diagram; and (3) Be answerable solely from the text. Based on these criteria, we build the dataset in follow stages.

\textbf{Question Template Design.}
As shown in Tab. \ref{tab:template_taxonomy}, we developed four primary question categories including Fault Localization, Functional Mapping, Boundary Testing, and Counterfactual Reasoning, and designed eight question templates for each category (templates detail provided in Appendix \ref{app:A}). Each template is carefully crafted to make a structure-aware interpretation necessary for solving the problem.

\textbf{Question Generation \& Model Verification.}
To generate suitable multi-hop questions, we implemented a three-step process:
(i) \textit{Template-based Multi-hop Question Generation}: Using GPT-o3, questions were generated based on provided texts and structural diagrams. Each question adhered strictly to one of the predefined templates, requiring reasoning involving at least two nodes to ensure sufficient complexity.
(ii) \textit{Correctness Verification}: Generated questions and their answers were cross-validated against reference diagrams using GPT-5.2 and Gemini-2.5-Pro, ensuring logical consistency.
(iii) \textit{Text Dependency Verification}: Questions and answers were assessed to confirm they could be inferred directly from the text alone, using GPT-5.2 and Gemini-2.5-Pro models. If any check failed at steps (ii) or (iii), question generation was repeated from step (i).
We aimed to generate four questions per text-structure pair (one per category) with a maximum of three generation attempts per question. Eventually, 2,150 valid samples were collected.

\textbf{Second-Round Human Filter.}
For further verification, a manual validation of questions and answers was conducted by 15 PhD-level experts across relevant domains. 
Reviewers evaluated each item based on the correctness, difficulty, coherence and format of the question (the full checklist is provided in Appendix \ref{app:B4}).
Only samples passing all checks were included. This comprehensive human filtering process spanned two weeks, ultimately yielding approximately 1.7k high-quality text-structure-question triples.

Finally, we perform a stratified 7:3 split by domain, producing T2S-Train-1.2k (training) and T2S-Bench-MR (test, 500 samples). During evaluation, only texts and questions were provided to models, with Exact Match (EM) and F1 scores 
used as evaluation metrics. 
% This approach effectively tests models' abilities in extracting structured information from texts while circumventing evaluation complexities arising from structural diversity.

% \vspace{-5pt}
\subsection{T2S-Bench-E2E Dataset Construction}
% \vspace{-7pt}
To evaluate models’ end-to-end text-to-structure extraction, we build T2S-Bench-E2E. As mentioned in Section 2.3 (Challenge 3), E2E evaluation is inherently hard because a single text may admit multiple valid structures. To ensure fair and comparable scoring, we follow three principles to construct sample in E2E task: (1) \textit{Focus on key nodes and links} that reflect the text’s core content while filtering noise, (2) \textit{Control graph complexity} to avoid both trivial and overly ambiguous cases, and (3) \textit{Partially constrain structure generation} so models are evaluated on the remaining elements rather than unconstrained free-form graphs.
Following these principles, T2S-Bench-E2E was developed in follow steps.

\textbf{Key Structure Extraction \& Model Check.} To obtain key structural frames, we:
(i) \textit{Transformed high-quality diagrams from Section 3.1 into JSON format} using GPT-o3, clearly defining nodes and links.
(ii) \textit{Generated key structures} by removing irrelevant nodes and links via GPT-o3, resulting in concise Text-KeyStructure pairs.
(iii) \textit{Conducted rigorous model checks} using GPT-5.2 and Gemini-2.5-Pro to verify logical coherence, text dependency, and consistency with reference diagrams. Samples failing this check were discarded to ensure dataset quality.

\textbf{Third-Round Human Filter.} Finally, five PhD students independently verified each Text-KeyStructure pair by inferring structures from texts and ensuring substantial alignment with provided key structures. Samples with significant deviations or excessive complexity were excluded. After approximately two weeks of rigorous quality control, 87 high-quality Text-KeyStructure pairs were finalized.

\textbf{Partial Structure-Constrained Evaluation.}
To facilitate fair assessments and limit evaluation complexity, nodes and links were evaluated separately in T2S-Bench-E2E:
\begin{enumerate}
% \vspace{-12pt}
    \item \textit{Link Evaluation}: Models received text and all node information ("nodes": [{"id": "n1", "label": "Node Text"}]) and predicted links in JSON format. Link correctness was measured using F1 scores.
    
    % \vspace{-7pt}
    
    \item \textit{Node Evaluation}: Models were given text and all existing links ("links": [{"source": "n1", "target": "n2", "label": "Link Text" }]), predicting corresponding nodes. Node evaluation was based on average semantic similarity between predicted nodes and ground truth.

\end{enumerate}
    % \vspace{-12pt}
This partial structural constraint evaluation method standardizes outputs for fair, accurate assessments, effectively isolating and measuring node and link extraction abilities.

% \vspace{-5pt}

\subsection{Data Distribution Statistics}
% \vspace{-5pt}
% T2S-Bench is designed to be balanced by construction: we target 50 samples per structure type, and for each sample we create four questions (one for each question family). After model-based and human filtering, the final counts vary slightly across types. The overall distribution is shown in Fig. \ref{fig_3}.
% Overall, the dataset remains well balanced across domains. Computer Science accounts for the largest share (27.5\%), largely because it has more open-access papers and a higher yield of high-quality text-structure pairs. Even the smallest category, Environment Science, still contributes 11\%, highlighting the breadth and coverage of T2S-Bench.

T2S-Bench is balanced by design: we aim for 50 samples per structure type, and we create four questions per sample (one per question family). After model and human filtering, counts vary slightly across types (Fig. \ref{fig_3}). The dataset is also well balanced across domains: Computer Science is the largest (27.5\%) due to more open-access papers and higher-quality text–structure yield, while Environment Science still contributes 11\%, demonstrating broad coverage.

\begin{table*}[t]
\centering
\small
\setlength{\tabcolsep}{4pt}
\caption{\textbf{Performance improvements after fine-tuning on the T2S-Train-1.2k.}
We fine-tuned Qwen2.5-7B-Instruct and Llama3.1-8B-Instruct models for 100 epochs using GRPO on T2S-Train-1.2k (detailed training settings provided in Appendix \ref{app:E1}). Evaluations were conducted on various text-processing tasks within Longbench and SCROLLS using lm-eval.
}
% \vspace{-5pt}
%T2S-Train-1.2k数据集微调模型后的性能提升。我们用GRPO在T2S-Train-1.2k对qwen2.5-7b-instruct和llama3.1-8b-instruct分别训练100epoch（详细训练设置提供在appendix x），并在lm-eval上对Longbench和scrolls任务进行性能评估（它们包含多个不同类型的文本处理任务）。
%
\resizebox{0.9\textwidth}{!}{
\begin{tabular}{lcccccccccc}
\toprule
 & \multicolumn{2}{c}{\textbf{T2S-Bench}} 
 & \multicolumn{5}{c}{\textbf{LongBench}} 
 & \multicolumn{3}{c}{\textbf{Scrolls}} \\
\cmidrule(lr){2-3}
\cmidrule(lr){4-8}
\cmidrule(lr){9-11}
\textbf{Model} 
& \textbf{MC (EM)} 
& \textbf{MC (F1)} 
& \textbf{HotpotQA} 
& \textbf{2WikiMQA} 
& \textbf{Qasper} 
& \textbf{GovReport} 
& \textbf{QMSum} 
& \textbf{ContractNLI} 
& \textbf{Quality} 
& \textbf{Summscreen} \\
\midrule

Qwen2.5-7B-Instruct 
& 28.8 & 59.4 & 60.0 & 46.9 & 43.5 & 32.0 & 23.9 & 55.8 & 37.8 & 17.3 \\
\quad + CoT 
& 36.6 & 62.1 & 62.6 & 59.0 & 41.9 & 25.7 & 19.7 & 52.1 & 38.4 & 15.8 \\
\quad + SoT 
& 40.6 & 68.4 & 65.8 & 63.2 & 48.0 & 37.3 & 27.9 & 58.4 & 41.4 & 21.3 \\
% \cdashline{1-11}[2pt/2pt]
% \rule{0pt}{10pt}
\rowcolor{colblue} \quad \textbf{+ T2S-Train} 
& \textbf{46.1} & \textbf{73.5} & \textbf{68.2} & \textbf{65.3} & \textbf{51.2} & \textbf{41.2} & \textbf{30.9} & \textbf{60.3} & \textbf{42.8} & \textbf{24.5} \\
\midrule

LLaMA3.1-8B-Instruct 
& 24.2 & 54.2 & 59.1 & 53.7 & 44.9 & 34.3 & 25.2 & 31.5 & 39.8 & 26.8 \\
\quad + CoT 
& 27.8 & 56.4 & 55.3 & 56.2 & 43.4 & 27.4 & 22.6 & 28.7 & 41.2 & 22.1 \\

\quad + SoT 
& 32.5 & 58.2 & 65.1 & 59.1 & 49.1 & 39.3 & 30.7 & 35.1 & 44.3 & 32.1 \\
% \cdashline{1-11}[2pt/2pt]
% \rule{0pt}{10pt}

\rowcolor{colblue} \quad \textbf{+ T2S-Train} 
& \textbf{38.1} & \textbf{64.2} & \textbf{69.2} & \textbf{63.2} & \textbf{51.8} & \textbf{42.9} & \textbf{35.2} & \textbf{39.2} & \textbf{48.5} & \textbf{34.5} \\
\bottomrule
\end{tabular}}

\label{tab:t2s_longbench_scrolls}
\end{table*}

\begin{figure*}
% \vspace{-10pt}
 \centering
 \centerline{\includegraphics[width=0.9\linewidth]{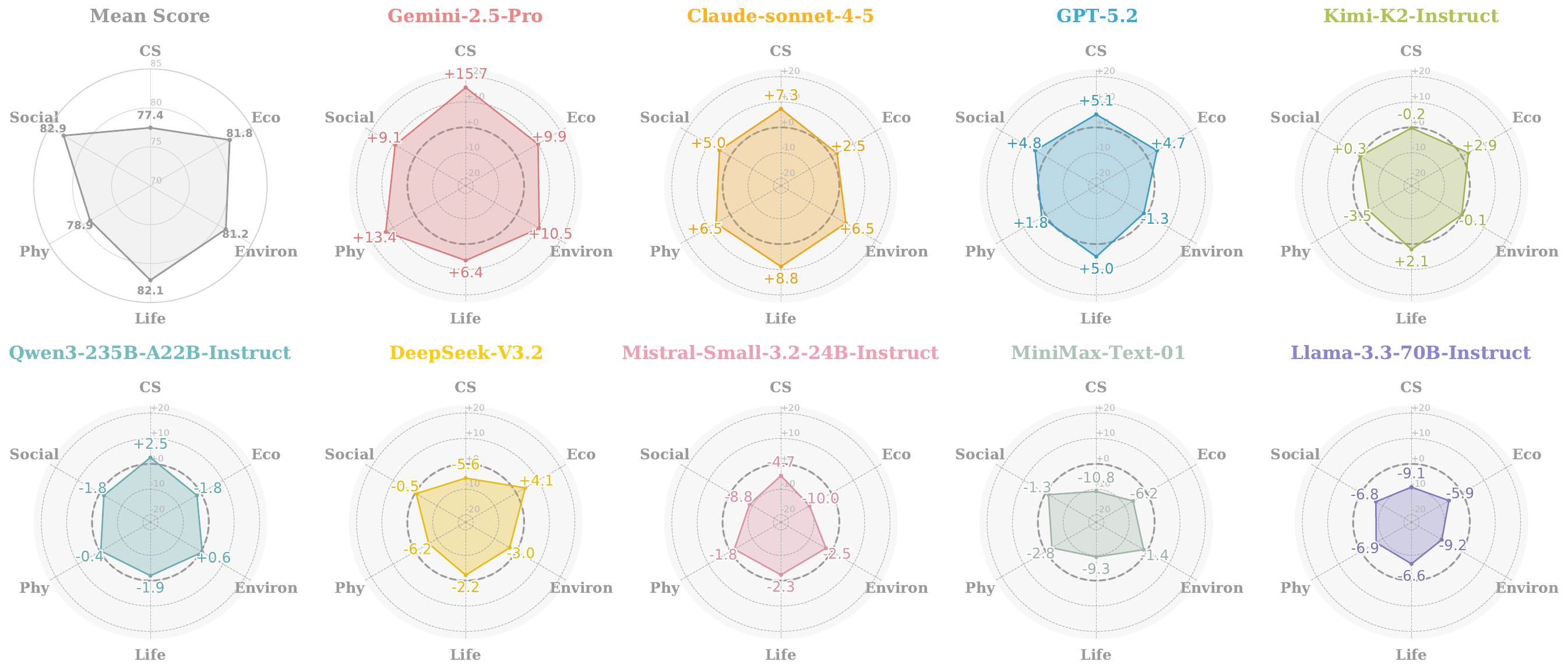}}
 % \vspace{-10pt}
 \caption{\textbf{F1 scores across different topics on T2S-Bench-MR.} We selected one representative model from each model family; The first fig shows their average F1 scores across various domains. The remains fig illustrate individual model performances per domain, with the vertical axis indicating deviations from the average performance. The dark dashed rectangle represents the average performance (set to zero). Scores outside this rectangle indicate above-average performance, while scores inside indicate below-average performance..}
 %F1 score across different Topic on T2S-Bench-MR.我们从各模型家族中选取了一个模型作为代表，图1展示了它们在各领域的f1分数均值。图2-10则展示了各模型在各领域上的表现，其中纵轴为相比于均值的性能差异，深色虚线框代表0（集均值），在虚线框外表面性能超过均值，反之则低于均值。 
 \label{fig_overall}
 % \vspace{-10pt}
\end{figure*}

\section{Evaluation}
% \vspace{-5pt}
In this section, we comprehensively report the performance of mainstream models on T2S-Bench and demonstrate the effectiveness of T2S-Train-1.2k. Detailed experimental settings are provided in Appendix \ref{app:E1}, and more extensive results and insights are included in Appendix \ref{app:E2} and \ref{app:E3}.

% \vspace{-5pt}
\subsection{General Performance on T2S‑Bench}\label{sec:gen_perf}
% \vspace{-5pt}
\textbf{Overall Results.} Tab. \,\ref{tab:multiqa_category} compares the performance of 45 mainstream language models on the T2S‑Bench benchmark, reporting the results both on T2S-Bench-MR and T2S-Bench-E2E. Several high‑level trends emerge. First, proprietary giants continue to dominate: Gemini‑2.5‑Pro tops the table with 81.40\% EM and 91.56\% F1, followed closely by Claude‑sonnet (76.80/86.85) and GPT‑5.2 (71.80/84.32). Second, instruction‑tuned open‑source models are rapidly closing the gap. Variants of Qwen3 and DeepSeek achieve overall EM in the 60\%–70\% range, illustrating that careful curation of training data and prompting can deliver competitive reasoning ability without proprietary resources. Third, older or smaller architectures such as GLM‑4.5, GLM‑4.6 and MiniMax‑M2 languish below 40\% EM, underscoring the importance of both model capacity and high‑quality instruction fine‑tuning for successful multi‑hop reasoning.
% reports the aggregate performance of representative models on the multi‑choice QA task. 
% Proprietary models such as Gemini‑2.5‑Pro and GPT‑5.2 dominate, achieving EM/F1 scores of 81.40/91.56 and 71.80/84.32, respectively. Open‑source models like Qwen3‑32B (69.40/83.41) and DeepSeek‑V3.2 (60.00/78.31) trail behind, while smaller or older models (e.g., GLM‑4.5 and MiniMax‑M2) struggle to exceed 40 EM. 
% Notably, performance varies across domains: physics and social science questions are handled relatively well, whereas environmental science and economics are more challenging for most models.

\textbf{Task Breakdown.} The granular breakdown reveals clear patterns: Boundary testing and counterfactual reasoning tend to be the easiest for strong models: top‑tier systems like Gemini‑2.5‑Pro, Claude‑sonnet and GPT‑5.2 achieve above 80\% EM on these categories. By contrast, fault localization consistently drags down performance. Even among the leaders, there is a 15\%–20\% drop between counterfactual reasoning and fault localization, reflecting the difficulty of tracing causal chains within complex graphs. Functional mapping sits between the extremes: high‑performing models solve most functional mapping questions, whereas weaker models like MiniMax‑M2, GLM‑4.6 and LLaMA‑3.2‑3B answer fewer than 10\% correctly. Overall, the per‑category results highlight that T2S‑Bench exercises a broad spectrum of reasoning skills and that models must handle a variety of inference types to succeed.

% To better understand model behaviour, Table\,\ref{tab:multiqa_category} also splits the multi‑choice QA results by reasoning category: Boundary Testing, Counterfactual Reasoning, Fault Localization and Functional Mapping. The gap between categories is pronounced. For example, Gemini‑2.5‑Pro attains 90.65 EM on counterfactual reasoning but only 74.51 EM on fault localization. Similarly, open‑source models such as DeepSeek‑V3.2 and Qwen3‑32B drop more than 20 points between their best and worst categories. These disparities highlight the diversity of reasoning skills required by T2S‑Bench.

\textbf{Structure Extraction.} Results on T2S-Bench-E2E in the last two columns of Tab.\,\ref{tab:multiqa_category} reveal that 
structure extraction remains a major bottleneck for all models. Across the board, Node similarity rarely exceeds 60\%: only Gemini‑2.5‑Pro and a handful of Claude models surpass the 55\%. Most open‑source and mid‑sized models cluster between 35\% and 50\%, and the smallest baselines (GLM‑4.5, MiniMax‑M2, LLaMA‑3.2‑3B) fall below 35\%. In contrast, Link F1 scores are uniformly higher, with leading models achieving 84\%–87\% and even weaker models often exceeding 70\%. This disparity implies that identifying the correct set of nodes is far harder than linking them once found. Since the Node score limits the potential Link score, continued advances in entity detection, co‑reference resolution, and discourse segmentation will be essential for closing the gap. 
%Detailed per‑domain structure results appear in the appendix.

% generating the correct node–link graph is substantially harder than selecting among multiple answers. Even the best model, Gemini‑2.5‑Pro, achieves only 58.09 NodeF1 and 84.32 LinkF1 overall. The gap between NodeF1 and LinkF1 suggests that inferring relations among predicted nodes is easier than identifying the nodes themselves. Open‑source models perform poorly (e.g., DeepSeek‑V3.2 at 46.98/78.69), and several models collapse on this task, highlighting the fundamental difficulty of full structure extraction.

Fig.\,\ref{fig_overall} visualises domain‑wise performance for selected models using radar charts. Each axis corresponds to a science domain, and values represent the deviation from the mean score. These plots illustrate that proprietary models maintain balanced performance across domains, whereas open‑source models exhibit larger fluctuations. For example, the Kimi‑K2 model excels in environmental science but underperforms in physics, while MiniMax‑Text‑01 struggles across all domains. The radar charts emphasise that T2S‑Bench requires broad, domain‑general reasoning skills.

\begin{figure}
% \vspace{-5pt}
 \centering
 \centerline{\includegraphics[width=\linewidth]{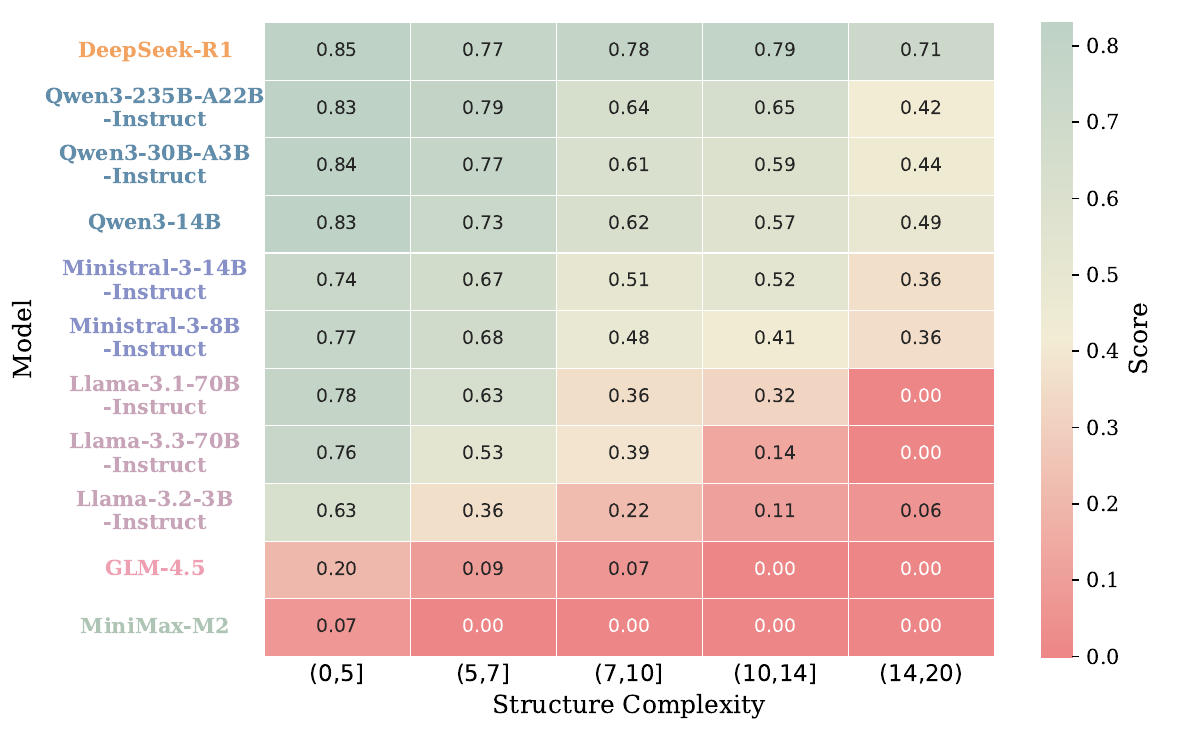}}
 % \vspace{-5pt} 
 \caption{Link F1 scores on MR-Bench-E2E across texts corresponding to reference graphs with varying node counts.}
 %link F1 score on MR-Bench-E2E across 不同节点数参考图对应的文本.
 \label{fig_heatmap}
 % \vspace{-17pt}
\end{figure}

% \vspace{-5pt}
\subsection{The Importance of Structure for Downstream Tasks}\label{sec:importance}
% \vspace{-5pt}
To evaluate whether improved structuring skills translate to better downstream task, we perform ablation experiments on Qwen2.5‑7B and LLaMA-3.1‑8B. Tab.\,\ref{tab:t2s_longbench_scrolls} summarises the effect of different prompting strategies—vanilla, CoT, SoT and T2S training—on T2S‑Bench and on long‑context tasks from LongBench and Scrolls. SoT consistently yields larger gains than CoT, and fine‑tuning on T2S‑Bench further boosts both in‑domain and out‑of‑domain performance. For example, Qwen2.5‑7B’s EM improves from 28.8\% to 46.1\% on T2S‑Bench and from 60.0\% to 68.2\% on HotpotQA, while LLaMA-3.1‑8B’s EM increases from 24.2\% to 38.1\% on T2S‑Bench and 59.1\% to 69.2\% on HotpotQA. These results demonstrate that structuring skills learned on T2S‑Bench generalise to real‑world long‑context tasks.

\begin{figure}
% \vspace{-5pt}
 \centering
 \centerline{\includegraphics[width=0.9\linewidth]{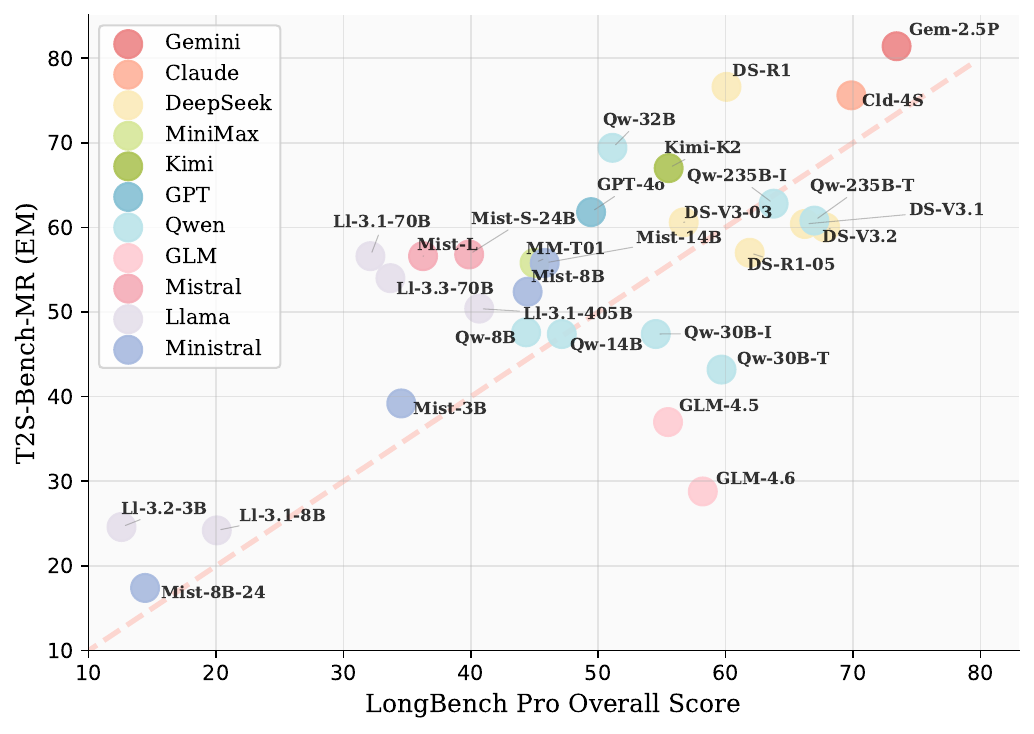}}
 % \vspace{-5pt}
 \caption{Correlation of model performance between T2S-Bench-MR and Longbench Pro Dataset.}
 \label{fig_correlation}
 % \vspace{-17pt}
\end{figure}
\textbf{Correlation analysis.} Fig.\,\ref{fig_correlation} plots T2S‑Bench-MR EM against LongBench Pro scores for a variety of models. A clear positive correlation emerges: models that perform well on T2S‑Bench also achieve high scores on LongBench. For example, Gemini‑2.5‑Pro, Claude sonnet and DeepSeek‑R1 occupy the upper right region of the plot, whereas weak models like LLaMA‑3.2‑3B and MiniMax‑M2 reside in the lower left. This relationship suggests that multi‑hop reasoning and structuring skills are indicative of general long‑context reasoning ability, reinforcing the value of structural thinking for downstream applications.
% This correlation holds not only for question‑answer accuracy but also for structure extraction. Models that produce higher NodeF1/LinkF1 scores on T2S‑Bench tend to achieve higher scores on LongBench tasks requiring document summarisation and question answering, reinforcing the value of structural thinking for downstream applications.

% \vspace{-7pt}
\subsection{Analysis Experiments}
% \vspace{-7pt}
To investigate how structure complexity affects model performance, we partition T2S-Bench-E2E by the number of nodes in the reference graph. Fig.\,\ref{fig_heatmap} shows the heatmap of LinkF1 scores across complexity bins for several models. As the number of nodes increases from 1–5 to 14–20, performance declines sharply. Models like DeepSeek‑R1 and Qwen3‑235B remain robust up to 10–14 nodes but degrade beyond that, while smaller models (e.g., LLaMA‑3.1‑8B) collapse to near‑zero when faced with more than 14 nodes. These results highlight that current models struggle to maintain accuracy as structural complexity increases, motivating future work on scalable structuring algorithms.
In summary, our evaluation demonstrates that T2S‑Bench provides a challenging testbed for assessing both reasoning and structuring capabilities of large models. The benchmark exposes significant performance gaps across domains and reasoning categories, emphasises the importance of explicit structure extraction, and shows that structuring skills transfer to downstream tasks. We hope that T2S‑Bench will spur further research into structure‑aware training and inference for long‑context language understanding.

% \vspace{-5pt}
\section{Conclusion}
% \vspace{-7pt}
In this study, we introduce T2S-Bench, the first comprehensive benchmark evaluating and improving text-to-structure capabilities. Derived from scientific literature, T2S-Bench spans six scientific domains and 32 structural types, employing carefully designed multi-hop reasoning (MR) evaluations and partially constrained end-to-end (E2E) extraction evaluation.
Benchmarking 45 mainstream models highlights significant improvement potential, notably in node extraction. Our fine-tuning experiments further show that enhanced structuring skills effectively transfer to downstream tasks. In summary, our results underscore structuring as a fundamental competence for reliable text understanding, encouraging future research in text structuring.

\section*{Impact Statement}
\textbf{Potential positive impacts.}
This paper contributes methods and resources for improving models' ability to convert long text into explicit intermediate structures. If adopted responsibly, SoT-style structuring and T2S-Bench-style evaluation may improve reliability in document-centric applications such as literature review, evidence-grounded question answering, and structured report generation. Intermediate structures can also increase \emph{auditability}: users and developers can inspect nodes/links to verify what information the model relied on, potentially reducing hallucinations and making failures easier to diagnose.

\textbf{Potential negative impacts and dual use.}
Stronger text-to-structure capability can also enable misuse. In particular, it may facilitate large-scale extraction of entities and relations from sensitive or proprietary documents, supporting surveillance, profiling, or targeted manipulation. Moreover, text structure can be repurposed to organize misleading narratives or generate persuasive but incorrect ``structured'' reports that appear authoritative. Finally, training and evaluating many large models can incur non-trivial computational and environmental costs.

\textbf{Mitigations and responsible use.}
Our benchmark is grounded in publicly available scientific writing, which reduces the likelihood of containing personal data; nevertheless, we encourage future users to apply appropriate privacy safeguards when deploying structuring techniques on private corpora (e.g., access control, data minimization, and redaction). We also recommend human-in-the-loop verification for high-stakes settings: the produced structure should be treated as an \emph{inspectable hypothesis} rather than a guaranteed faithful representation. For dataset release and downstream usage, practitioners should respect copyright/licensing constraints of source materials and follow venue policies on research ethics and data usage.

In summary, we believe the primary expected impact of this work is to provide a unified paradigm and benchmark for structure-aware text processing that improves robustness and transparency, while acknowledging the dual-use risks inherent to stronger information extraction and recommending safeguards for responsible deployment.

% \onecolumn
%%%%%%%%%%%%%%%%%%%%%%%%%%%%%%%%%%%%%%%%%%%%%%%%%%%%%%%%%%%%%%%%%%%%%%%%%%%%%%%
%%%%%%%%%%%%%%%%%%%%%%%%%%%%%%%%%%%%%%%%%%%%%%%%%%%%%%%%%%%%%%%%%%%%%%%%%%%%%%%
\bibliography{example_paper}

@inproceedings{liang2025reasoning,
  title={Reasoning rag via system 1 or system 2: A survey on reasoning agentic retrieval-augmented generation for industry challenges},
  author={Liang, Jintao and Lin, Huifeng and Wu, You and Zhao, Rui and Li, Ziyue and others},
  booktitle={Proceedings of the 14th International Joint Conference on Natural Language Processing and the 4th Conference of the Asia-Pacific Chapter of the Association for Computational Linguistics},
  pages={1954--1966},
  year={2025}
}

@article{xi2025survey,
  title={A survey of llm-based deep search agents: Paradigm, optimization, evaluation, and challenges},
  author={Xi, Yunjia and Lin, Jianghao and Xiao, Yongzhao and Zhou, Zheli and Shan, Rong and Gao, Te and Zhu, Jiachen and Liu, Weiwen and Yu, Yong and Zhang, Weinan},
  journal={arXiv preprint arXiv:2508.05668},
  year={2025}
}

@misc{zheng2025pptagent,
      title={PPTAgent: Generating and Evaluating Presentations Beyond Text-to-Slides}, 
      author={Hao Zheng and Xinyan Guan and Hao Kong and Jia Zheng and Weixiang Zhou and Hongyu Lin and Yaojie Lu and Ben He and Xianpei Han and Le Sun},
      year={2025},
      eprint={2501.03936},
      archivePrefix={arXiv},
      primaryClass={cs.AI},
      url={https://arxiv.org/abs/2501.03936}, 
}

@article{zhang2025evolving,
  title={The evolving role of large language models in scientific innovation: Evaluator, collaborator, and scientist},
  author={Zhang, Haoxuan and Li, Ruochi and Zhang, Yang and Xiao, Ting and Chen, Jiangping and Ding, Junhua and Chen, Haihua},
  journal={arXiv preprint arXiv:2507.11810},
  year={2025}
}

@misc{li2023sheetcopilot,
      title={SheetCopilot: Bringing Software Productivity to the Next Level through Large Language Models}, 
      author={Hongxin Li and Jingran Su and Yuntao Chen and Qing Li and Zhaoxiang Zhang},
      year={2023},
      eprint={2305.19308},
      archivePrefix={arXiv},
      primaryClass={cs.SE},
      url={https://arxiv.org/abs/2305.19308}, 
}

@inproceedings{fu2022doc2ppt,
  title={Doc2ppt: Automatic presentation slides generation from scientific documents},
  author={Fu, Tsu-Jui and Wang, William Yang and McDuff, Daniel and Song, Yale},
  booktitle={Proceedings of the AAAI Conference on Artificial Intelligence},
  volume={36},
  number={1},
  pages={634--642},
  year={2022}
}

@article{song2025evaluating,
  title={Evaluating large language models in scientific discovery},
  author={Song, Zhangde and Lu, Jieyu and Du, Yuanqi and Yu, Botao and Pruyn, Thomas M and Huang, Yue and Guo, Kehan and Luo, Xiuzhe and Qu, Yuanhao and Qu, Yi and others},
  journal={arXiv preprint arXiv:2512.15567},
  year={2025}
}

@inproceedings{du2025enhancing,
  title={Enhancing Long Document Long Form Summarisation with Self-Planning},
  author={Du, Xiaotang and Saxena, Rohit and Perez-Beltrachini, Laura and Minervini, Pasquale and Titov, Ivan},
  booktitle={Proceedings of the 14th International Joint Conference on Natural Language Processing and the 4th Conference of the Asia-Pacific Chapter of the Association for Computational Linguistics},
  pages={317--332},
  year={2025}
}

@misc{lin2025srag,
      title={SRAG: Structured Retrieval-Augmented Generation for Multi-Entity Question Answering over Wikipedia Graph}, 
      author={Teng Lin and Yizhang Zhu and Yuyu Luo and Nan Tang},
      year={2025},
      eprint={2503.01346},
      archivePrefix={arXiv},
      primaryClass={cs.CL},
      url={https://arxiv.org/abs/2503.01346}, 
}

@inproceedings{yang2018hotpotqa,
  title={{HotpotQA}: A Dataset for Diverse, Explainable Multi-hop Question Answering},
  author={Yang, Zhilin and Qi, Peng and Zhang, Saizheng and Bengio, Yoshua and Cohen, William W. and Salakhutdinov, Ruslan and Manning, Christopher D.},
  booktitle={Conference on Empirical Methods in Natural Language Processing ({EMNLP})},
  year={2018}
}

@article{trivedi2022musique,
  title={MuSiQue: Multihop Questions via Single-hop Question Composition},
  author={Trivedi, Harsh and Balasubramanian, Niranjan and Khot, Tushar and Sabharwal, Ashish},
  journal={Transactions of the Association for Computational Linguistics},
  volume={10},
  pages={539--554},
  year={2022},
  publisher={MIT Press One Broadway, 12th Floor, Cambridge, Massachusetts 02142, USA~…}
}

@misc{bai2024longbench,
      title={LongBench: A Bilingual, Multitask Benchmark for Long Context Understanding}, 
      author={Yushi Bai and Xin Lv and Jiajie Zhang and Hongchang Lyu and Jiankai Tang and Zhidian Huang and Zhengxiao Du and Xiao Liu and Aohan Zeng and Lei Hou and Yuxiao Dong and Jie Tang and Juanzi Li},
      year={2024},
      eprint={2308.14508},
      archivePrefix={arXiv},
      primaryClass={cs.CL},
      url={https://arxiv.org/abs/2308.14508}, 
}

@article{dasigi2021dataset,
  title={A dataset of information-seeking questions and answers anchored in research papers},
  author={Dasigi, Pradeep and Lo, Kyle and Beltagy, Iz and Cohan, Arman and Smith, Noah A and Gardner, Matt},
  journal={arXiv preprint arXiv:2105.03011},
  year={2021}
}

@inproceedings{huang-etal-2021-efficient,
    title = "Efficient Attentions for Long Document Summarization",
    author = "Huang, Luyang  and
      Cao, Shuyang  and
      Parulian, Nikolaus  and
      Ji, Heng  and
      Wang, Lu",
    booktitle = "Proceedings of the 2021 Conference of the North American Chapter of the Association for Computational Linguistics: Human Language Technologies",
    month = jun,
    year = "2021",
    address = "Online",
    publisher = "Association for Computational Linguistics",
    url = "https://aclanthology.org/2021.naacl-main.112",
    doi = "10.18653/v1/2021.naacl-main.112",
    pages = "1419--1436",
}

@misc{zhong2021qmsum,
      title={QMSum: A New Benchmark for Query-based Multi-domain Meeting Summarization}, 
      author={Ming Zhong and Da Yin and Tao Yu and Ahmad Zaidi and Mutethia Mutuma and Rahul Jha and Ahmed Hassan Awadallah and Asli Celikyilmaz and Yang Liu and Xipeng Qiu and Dragomir Radev},
      year={2021},
      eprint={2104.05938},
      archivePrefix={arXiv},
      primaryClass={cs.CL},
      url={https://arxiv.org/abs/2104.05938}, 
}

@article{pdffigures2,
     title = {PDFFigures 2.0: Mining Figures from Research Papers},
     author = {Christopher Clark and Santosh Divvala},
     booktitle = {JCDL},
     year = {2016}
}

@misc{bai2025longbenchv2,
      title={LongBench v2: Towards Deeper Understanding and Reasoning on Realistic Long-context Multitasks}, 
      author={Yushi Bai and Shangqing Tu and Jiajie Zhang and Hao Peng and Xiaozhi Wang and Xin Lv and Shulin Cao and Jiazheng Xu and Lei Hou and Yuxiao Dong and Jie Tang and Juanzi Li},
      year={2025},
      eprint={2412.15204},
      archivePrefix={arXiv},
      primaryClass={cs.CL},
      url={https://arxiv.org/abs/2412.15204}, 
}

@misc{chen2026longbenchpro,
      title={LongBench Pro: A More Realistic and Comprehensive Bilingual Long-Context Evaluation Benchmark}, 
      author={Ziyang Chen and Xing Wu and Junlong Jia and Chaochen Gao and Qi Fu and Debing Zhang and Songlin Hu},
      year={2026},
      eprint={2601.02872},
      archivePrefix={arXiv},
      primaryClass={cs.CL},
      url={https://arxiv.org/abs/2601.02872}, 
}

@misc{yang2026structeval,
      title={StructEval: Benchmarking LLMs' Capabilities to Generate Structural Outputs}, 
      author={Jialin Yang and Dongfu Jiang and Lipeng He and Sherman Siu and Yuxuan Zhang and Disen Liao and Zhuofeng Li and Huaye Zeng and Yiming Jia and Haozhe Wang and Benjamin Schneider and Chi Ruan and Wentao Ma and Zhiheng Lyu and Yifei Wang and Yi Lu and Quy Duc Do and Ziyan Jiang and Ping Nie and Wenhu Chen},
      year={2026},
      eprint={2505.20139},
      archivePrefix={arXiv},
      primaryClass={cs.SE},
      url={https://arxiv.org/abs/2505.20139}, 
}

@misc{gu2024structbench,
      title={StrucText-Eval: Evaluating Large Language Model's Reasoning Ability in Structure-Rich Text}, 
      author={Zhouhong Gu and Haoning Ye and Xingzhou Chen and Zeyang Zhou and Hongwei Feng and Yanghua Xiao},
      year={2024},
      eprint={2406.10621},
      archivePrefix={arXiv},
      primaryClass={cs.CL},
      url={https://arxiv.org/abs/2406.10621}, 
}

@misc{jiang2025hibench,
      title={HiBench: Benchmarking LLMs Capability on Hierarchical Structure Reasoning}, 
      author={Zhuohang Jiang and Pangjing Wu and Ziran Liang and Peter Q. Chen and Xu Yuan and Ye Jia and Jiancheng Tu and Chen Li and Peter H. F. Ng and Qing Li},
      year={2025},
      eprint={2503.00912},
      archivePrefix={arXiv},
      primaryClass={cs.CL},
      url={https://arxiv.org/abs/2503.00912}, 
}

@article{team2023gemini,
  title={Gemini: a family of highly capable multimodal models},
  author={Team, Gemini and Anil, Rohan and Borgeaud, Sebastian and Alayrac, Jean-Baptiste and Yu, Jiahui and Soricut, Radu and Schalkwyk, Johan and Dai, Andrew M and Hauth, Anja and Millican, Katie and others},
  journal={arXiv preprint arXiv:2312.11805},
  year={2023}
}

@article{comanici2025gemini,
  title={Gemini 2.5: Pushing the frontier with advanced reasoning, multimodality, long context, and next generation agentic capabilities},
  author={Comanici, Gheorghe and Bieber, Eric and Schaekermann, Mike and Pasupat, Ice and Sachdeva, Noveen and Dhillon, Inderjit and Blistein, Marcel and Ram, Ori and Zhang, Dan and Rosen, Evan and others},
  journal={arXiv preprint arXiv:2507.06261},
  year={2025}
}

@article{liu2024deepseek,
  title={Deepseek-v3 technical report},
  author={Liu, Aixin and Feng, Bei and Xue, Bing and Wang, Bingxuan and Wu, Bochao and Lu, Chengda and Zhao, Chenggang and Deng, Chengqi and Zhang, Chenyu and Ruan, Chong and others},
  journal={arXiv preprint arXiv:2412.19437},
  year={2024}
}

@article{guo2025deepseek,
  title={Deepseek-r1: Incentivizing reasoning capability in llms via reinforcement learning},
  author={Guo, Daya and Yang, Dejian and Zhang, Haowei and Song, Junxiao and Zhang, Ruoyu and Xu, Runxin and Zhu, Qihao and Ma, Shirong and Wang, Peiyi and Bi, Xiao and others},
  journal={arXiv preprint arXiv:2501.12948},
  year={2025}
}

@article{team2025kimi,
  title={Kimi k2: Open agentic intelligence},
  author={Team, Kimi and Bai, Yifan and Bao, Yiping and Chen, Guanduo and Chen, Jiahao and Chen, Ningxin and Chen, Ruijue and Chen, Yanru and Chen, Yuankun and Chen, Yutian and others},
  journal={arXiv preprint arXiv:2507.20534},
  year={2025}
}

@article{yang2025qwen3,
  title={Qwen3 technical report},
  author={Yang, An and Li, Anfeng and Yang, Baosong and Zhang, Beichen and Hui, Binyuan and Zheng, Bo and Yu, Bowen and Gao, Chang and Huang, Chengen and Lv, Chenxu and others},
  journal={arXiv preprint arXiv:2505.09388},
  year={2025}
}

@article{zeng2025glm,
  title={Glm-4.5: Agentic, reasoning, and coding (arc) foundation models},
  author={Zeng, Aohan and Lv, Xin and Zheng, Qinkai and Hou, Zhenyu and Chen, Bin and Xie, Chengxing and Wang, Cunxiang and Yin, Da and Zeng, Hao and Zhang, Jiajie and others},
  journal={arXiv preprint arXiv:2508.06471},
  year={2025}
}

@article{li2025minimax,
  title={Minimax-01: Scaling foundation models with lightning attention},
  author={Li, Aonian and Gong, Bangwei and Yang, Bo and Shan, Boji and Liu, Chang and Zhu, Cheng and Zhang, Chunhao and Guo, Congchao and Chen, Da and Li, Dong and others},
  journal={arXiv preprint arXiv:2501.08313},
  year={2025}
}

@misc{jiang2023mistral7b,
      title={Mistral 7B}, 
      author={Albert Q. Jiang and Alexandre Sablayrolles and Arthur Mensch and Chris Bamford and Devendra Singh Chaplot and Diego de las Casas and Florian Bressand and Gianna Lengyel and Guillaume Lample and Lucile Saulnier and Lélio Renard Lavaud and Marie-Anne Lachaux and Pierre Stock and Teven Le Scao and Thibaut Lavril and Thomas Wang and Timothée Lacroix and William El Sayed},
      year={2023},
      eprint={2310.06825},
      archivePrefix={arXiv},
      primaryClass={cs.CL},
      url={https://arxiv.org/abs/2310.06825}, 
}

@article{touvron2023llama,
  title={Llama: Open and efficient foundation language models},
  author={Touvron, Hugo and Lavril, Thibaut and Izacard, Gautier and Martinet, Xavier and Lachaux, Marie-Anne and Lacroix, Timoth{\'e}e and Rozi{\`e}re, Baptiste and Goyal, Naman and Hambro, Eric and Azhar, Faisal and others},
  journal={arXiv preprint arXiv:2302.13971},
  year={2023}
}

@article{grattafiori2024llama,
  title={The llama 3 herd of models},
  author={Grattafiori, Aaron and Dubey, Abhimanyu and Jauhri, Abhinav and Pandey, Abhinav and Kadian, Abhishek and Al-Dahle, Ahmad and Letman, Aiesha and Mathur, Akhil and Schelten, Alan and Vaughan, Alex and others},
  journal={arXiv preprint arXiv:2407.21783},
  year={2024}
}

@article{achiam2023gpt,
  title={Gpt-4 technical report},
  author={Achiam, Josh and Adler, Steven and Agarwal, Sandhini and Ahmad, Lama and Akkaya, Ilge and Aleman, Florencia Leoni and Almeida, Diogo and Altenschmidt, Janko and Altman, Sam and Anadkat, Shyamal and others},
  journal={arXiv preprint arXiv:2303.08774},
  year={2023}
}

@misc{openai2024gpt4ocard,
      title={GPT-4o System Card}, 
      author={OpenAI and : and Aaron Hurst and Adam Lerer and Adam P. Goucher and Adam Perelman and Aditya Ramesh and Aidan Clark and AJ Ostrow and Akila Welihinda and Alan Hayes and Alec Radford and Aleksander Madry and Alex Baker-Whitcomb and Alex Beutel and Alex Borzunov and Alex Carney and Alex Chow and Alex Kirillov and Alex Nichol and Alex Paino and Alex Renzin and Alex Tachard Passos and Alexander Kirillov and Alexi Christakis and Alexis Conneau and Ali Kamali and Allan Jabri and Allison Moyer and Allison Tam and Amadou Crookes and Amin Tootoochian and Amin Tootoonchian and Ananya Kumar and Andrea Vallone and Andrej Karpathy and Andrew Braunstein and Andrew Cann and Andrew Codispoti and Andrew Galu and Andrew Kondrich and Andrew Tulloch and Andrey Mishchenko and Angela Baek and Angela Jiang and Antoine Pelisse and Antonia Woodford and Anuj Gosalia and Arka Dhar and Ashley Pantuliano and Avi Nayak and Avital Oliver and Barret Zoph and Behrooz Ghorbani and Ben Leimberger and Ben Rossen and Ben Sokolowsky and Ben Wang and Benjamin Zweig and Beth Hoover and Blake Samic and Bob McGrew and Bobby Spero and Bogo Giertler and Bowen Cheng and Brad Lightcap and Brandon Walkin and Brendan Quinn and Brian Guarraci and Brian Hsu and Bright Kellogg and Brydon Eastman and Camillo Lugaresi and Carroll Wainwright and Cary Bassin and Cary Hudson and Casey Chu and Chad Nelson and Chak Li and Chan Jun Shern and Channing Conger and Charlotte Barette and Chelsea Voss and Chen Ding and Cheng Lu and Chong Zhang and Chris Beaumont and Chris Hallacy and Chris Koch and Christian Gibson and Christina Kim and Christine Choi and Christine McLeavey and Christopher Hesse and Claudia Fischer and Clemens Winter and Coley Czarnecki and Colin Jarvis and Colin Wei and Constantin Koumouzelis and Dane Sherburn and Daniel Kappler and Daniel Levin and Daniel Levy and David Carr and David Farhi and David Mely and David Robinson and David Sasaki and Denny Jin and Dev Valladares and Dimitris Tsipras and Doug Li and Duc Phong Nguyen and Duncan Findlay and Edede Oiwoh and Edmund Wong and Ehsan Asdar and Elizabeth Proehl and Elizabeth Yang and Eric Antonow and Eric Kramer and Eric Peterson and Eric Sigler and Eric Wallace and Eugene Brevdo and Evan Mays and Farzad Khorasani and Felipe Petroski Such and Filippo Raso and Francis Zhang and Fred von Lohmann and Freddie Sulit and Gabriel Goh and Gene Oden and Geoff Salmon and Giulio Starace and Greg Brockman and Hadi Salman and Haiming Bao and Haitang Hu and Hannah Wong and Haoyu Wang and Heather Schmidt and Heather Whitney and Heewoo Jun and Hendrik Kirchner and Henrique Ponde de Oliveira Pinto and Hongyu Ren and Huiwen Chang and Hyung Won Chung and Ian Kivlichan and Ian O'Connell and Ian O'Connell and Ian Osband and Ian Silber and Ian Sohl and Ibrahim Okuyucu and Ikai Lan and Ilya Kostrikov and Ilya Sutskever and Ingmar Kanitscheider and Ishaan Gulrajani and Jacob Coxon and Jacob Menick and Jakub Pachocki and James Aung and James Betker and James Crooks and James Lennon and Jamie Kiros and Jan Leike and Jane Park and Jason Kwon and Jason Phang and Jason Teplitz and Jason Wei and Jason Wolfe and Jay Chen and Jeff Harris and Jenia Varavva and Jessica Gan Lee and Jessica Shieh and Ji Lin and Jiahui Yu and Jiayi Weng and Jie Tang and Jieqi Yu and Joanne Jang and Joaquin Quinonero Candela and Joe Beutler and Joe Landers and Joel Parish and Johannes Heidecke and John Schulman and Jonathan Lachman and Jonathan McKay and Jonathan Uesato and Jonathan Ward and Jong Wook Kim and Joost Huizinga and Jordan Sitkin and Jos Kraaijeveld and Josh Gross and Josh Kaplan and Josh Snyder and Joshua Achiam and Joy Jiao and Joyce Lee and Juntang Zhuang and Justyn Harriman and Kai Fricke and Kai Hayashi and Karan Singhal and Katy Shi and Kavin Karthik and Kayla Wood and Kendra Rimbach and Kenny Hsu and Kenny Nguyen and Keren Gu-Lemberg and Kevin Button and Kevin Liu and Kiel Howe and Krithika Muthukumar and Kyle Luther and Lama Ahmad and Larry Kai and Lauren Itow and Lauren Workman and Leher Pathak and Leo Chen and Li Jing and Lia Guy and Liam Fedus and Liang Zhou and Lien Mamitsuka and Lilian Weng and Lindsay McCallum and Lindsey Held and Long Ouyang and Louis Feuvrier and Lu Zhang and Lukas Kondraciuk and Lukasz Kaiser and Luke Hewitt and Luke Metz and Lyric Doshi and Mada Aflak and Maddie Simens and Madelaine Boyd and Madeleine Thompson and Marat Dukhan and Mark Chen and Mark Gray and Mark Hudnall and Marvin Zhang and Marwan Aljubeh and Mateusz Litwin and Matthew Zeng and Max Johnson and Maya Shetty and Mayank Gupta and Meghan Shah and Mehmet Yatbaz and Meng Jia Yang and Mengchao Zhong and Mia Glaese and Mianna Chen and Michael Janner and Michael Lampe and Michael Petrov and Michael Wu and Michele Wang and Michelle Fradin and Michelle Pokrass and Miguel Castro and Miguel Oom Temudo de Castro and Mikhail Pavlov and Miles Brundage and Miles Wang and Minal Khan and Mira Murati and Mo Bavarian and Molly Lin and Murat Yesildal and Nacho Soto and Natalia Gimelshein and Natalie Cone and Natalie Staudacher and Natalie Summers and Natan LaFontaine and Neil Chowdhury and Nick Ryder and Nick Stathas and Nick Turley and Nik Tezak and Niko Felix and Nithanth Kudige and Nitish Keskar and Noah Deutsch and Noel Bundick and Nora Puckett and Ofir Nachum and Ola Okelola and Oleg Boiko and Oleg Murk and Oliver Jaffe and Olivia Watkins and Olivier Godement and Owen Campbell-Moore and Patrick Chao and Paul McMillan and Pavel Belov and Peng Su and Peter Bak and Peter Bakkum and Peter Deng and Peter Dolan and Peter Hoeschele and Peter Welinder and Phil Tillet and Philip Pronin and Philippe Tillet and Prafulla Dhariwal and Qiming Yuan and Rachel Dias and Rachel Lim and Rahul Arora and Rajan Troll and Randall Lin and Rapha Gontijo Lopes and Raul Puri and Reah Miyara and Reimar Leike and Renaud Gaubert and Reza Zamani and Ricky Wang and Rob Donnelly and Rob Honsby and Rocky Smith and Rohan Sahai and Rohit Ramchandani and Romain Huet and Rory Carmichael and Rowan Zellers and Roy Chen and Ruby Chen and Ruslan Nigmatullin and Ryan Cheu and Saachi Jain and Sam Altman and Sam Schoenholz and Sam Toizer and Samuel Miserendino and Sandhini Agarwal and Sara Culver and Scott Ethersmith and Scott Gray and Sean Grove and Sean Metzger and Shamez Hermani and Shantanu Jain and Shengjia Zhao and Sherwin Wu and Shino Jomoto and Shirong Wu and Shuaiqi and Xia and Sonia Phene and Spencer Papay and Srinivas Narayanan and Steve Coffey and Steve Lee and Stewart Hall and Suchir Balaji and Tal Broda and Tal Stramer and Tao Xu and Tarun Gogineni and Taya Christianson and Ted Sanders and Tejal Patwardhan and Thomas Cunninghman and Thomas Degry and Thomas Dimson and Thomas Raoux and Thomas Shadwell and Tianhao Zheng and Todd Underwood and Todor Markov and Toki Sherbakov and Tom Rubin and Tom Stasi and Tomer Kaftan and Tristan Heywood and Troy Peterson and Tyce Walters and Tyna Eloundou and Valerie Qi and Veit Moeller and Vinnie Monaco and Vishal Kuo and Vlad Fomenko and Wayne Chang and Weiyi Zheng and Wenda Zhou and Wesam Manassra and Will Sheu and Wojciech Zaremba and Yash Patil and Yilei Qian and Yongjik Kim and Youlong Cheng and Yu Zhang and Yuchen He and Yuchen Zhang and Yujia Jin and Yunxing Dai and Yury Malkov},
      year={2024},
      eprint={2410.21276},
      archivePrefix={arXiv},
      primaryClass={cs.CL},
      url={https://arxiv.org/abs/2410.21276}, 
}

@article{singh2025openai,
  title={Openai gpt-5 system card},
  author={Singh, Aaditya and Fry, Adam and Perelman, Adam and Tart, Adam and Ganesh, Adi and El-Kishky, Ahmed and McLaughlin, Aidan and Low, Aiden and Ostrow, AJ and Ananthram, Akhila and others},
  journal={arXiv preprint arXiv:2601.03267},
  year={2025}
}

@article{brown2020language,
  title={Language models are few-shot learners},
  author={Brown, Tom and Mann, Benjamin and Ryder, Nick and Subbiah, Melanie and Kaplan, Jared D and Dhariwal, Prafulla and Neelakantan, Arvind and Shyam, Pranav and Sastry, Girish and Askell, Amanda and others},
  journal={Advances in neural information processing systems},
  volume={33},
  pages={1877--1901},
  year={2020}
}

@inproceedings{TheC3,
  title={The Claude 3 Model Family: Opus, Sonnet, Haiku},
  author={},
  url={https://api.semanticscholar.org/CorpusID:268232499}
}

@misc{ho2020multihopqa,
      title={Constructing A Multi-hop QA Dataset for Comprehensive Evaluation of Reasoning Steps}, 
      author={Xanh Ho and Anh-Khoa Duong Nguyen and Saku Sugawara and Akiko Aizawa},
      year={2020},
      eprint={2011.01060},
      archivePrefix={arXiv},
      primaryClass={cs.CL},
      url={https://arxiv.org/abs/2011.01060}, 
}

@inproceedings{saad2024pdftriage,
  title={Pdftriage: Question answering over long, structured documents},
  author={Saad-Falcon, Jon and Barrow, Joe and Siu, Alexa and Nenkova, Ani and Yoon, Seunghyun and Rossi, Ryan A and Dernoncourt, Franck},
  booktitle={Proceedings of the 2024 Conference on Empirical Methods in Natural Language Processing: Industry Track},
  pages={153--169},
  year={2024}
}

@misc{cheng2024structureguidedprompt,
      title={Structure Guided Prompt: Instructing Large Language Model in Multi-Step Reasoning by Exploring Graph Structure of the Text}, 
      author={Kewei Cheng and Nesreen K. Ahmed and Theodore Willke and Yizhou Sun},
      year={2024},
      eprint={2402.13415},
      archivePrefix={arXiv},
      primaryClass={cs.CL},
      url={https://arxiv.org/abs/2402.13415}, 
}

@misc{yao2023treethoughts,
      title={Tree of Thoughts: Deliberate Problem Solving with Large Language Models}, 
      author={Shunyu Yao and Dian Yu and Jeffrey Zhao and Izhak Shafran and Thomas L. Griffiths and Yuan Cao and Karthik Narasimhan},
      year={2023},
      eprint={2305.10601},
      archivePrefix={arXiv},
      primaryClass={cs.CL},
      url={https://arxiv.org/abs/2305.10601}, 
}

@article{Besta_2024,
   title={Graph of Thoughts: Solving Elaborate Problems with Large Language Models},
   volume={38},
   ISSN={2159-5399},
   url={http://dx.doi.org/10.1609/aaai.v38i16.29720},
   DOI={10.1609/aaai.v38i16.29720},
   number={16},
   journal={Proceedings of the AAAI Conference on Artificial Intelligence},
   publisher={Association for the Advancement of Artificial Intelligence (AAAI)},
   author={Besta, Maciej and Blach, Nils and Kubicek, Ales and Gerstenberger, Robert and Podstawski, Michal and Gianinazzi, Lukas and Gajda, Joanna and Lehmann, Tomasz and Niewiadomski, Hubert and Nyczyk, Piotr and Hoefler, Torsten},
   year={2024},
   month=mar, pages={17682–17690} }

@article{wei2022chain,
  title={Chain-of-thought prompting elicits reasoning in large language models},
  author={Wei, Jason and Wang, Xuezhi and Schuurmans, Dale and Bosma, Maarten and Xia, Fei and Chi, Ed and Le, Quoc V and Zhou, Denny and others},
  journal={Advances in neural information processing systems},
  volume={35},
  pages={24824--24837},
  year={2022}
}

@article{wang2023dgl,
  title={DGL: Device generic latency model for neural architecture search on mobile devices},
  author={Wang, Qinsi and Zhang, Sihai},
  journal={IEEE Transactions on Mobile Computing},
  volume={23},
  number={2},
  pages={1954--1967},
  year={2023},
  publisher={IEEE}
}

@article{wang2023mathnas,
  title={Mathnas: if blocks have a role in mathematical architecture design},
  author={Wang*, Qinsi and Ke*, Jinghan and Liang, Zhi and Zhang, Sihai},
  journal={Advances in Neural Information Processing Systems},
  volume={36},
  pages={47475--47486},
  year={2023}
}

@article{wang2024coreinfer,
  title={Coreinfer: Accelerating large language model inference with semantics-inspired adaptive sparse activation},
  author={Wang, Qinsi and Vahidian, Saeed and Ye, Hancheng and Gu, Jianyang and Zhang, Jianyi and Chen, Yiran},
  journal={arXiv preprint arXiv:2410.18311},
  year={2024}
}

@article{kuo2025h,
  title={H-cot: Hijacking the chain-of-thought safety reasoning mechanism to jailbreak large reasoning models, including openai o1/o3, deepseek-r1, and gemini 2.0 flash thinking},
  author={Kuo, Martin and Zhang, Jianyi and Ding, Aolin and Wang, Qinsi and DiValentin, Louis and Bao, Yujia and Wei, Wei and Li, Hai and Chen, Yiran},
  journal={arXiv preprint arXiv:2502.12893},
  year={2025}
}

@inproceedings{wang2025dobi,
  title={Dobi-SVD: Differentiable SVD for LLM Compression and Some New Perspectives},
  author={Wang*, Qinsi and Ke*, Jinghan and Tomizuka, Masayoshi and Keutzer, Kurt and Xu, Chenfeng},
  booktitle={The Thirteenth International Conference on Learning Representations},
  year={2025}
}

@inproceedings{wang2025corematching,
  title={CoreMatching: A Co-adaptive Sparse Inference Framework with Token and Neuron Pruning for Comprehensive Acceleration of Vision-Language Models},
  author={Wang, Qinsi and Ye, Hancheng and Chung, Ming-Yu and Liu, Yudong and Lin, Yueqian and Kuo, Martin and Ma, Mingyuan and Zhang, Jianyi and Chen, Yiran},
  booktitle={Forty-second International Conference on Machine Learning},
  year={2025}
}

@article{wang2025angles,
  title={Angles Don't Lie: Unlocking Training-Efficient RL Through the Model's Own Signals},
  author={Wang, Qinsi and Ke, Jinghan and Ye, Hancheng and Lin, Yueqian and Fu, Yuzhe and Zhang, Jianyi and Keutzer, Kurt and Xu, Chenfeng and Chen, Yiran},
  journal={NeurIPS 2025 Spotlight},
  year={2025}
}

@article{shao2025flashsvd,
  title={Flashsvd: Memory-efficient inference with streaming for low-rank models},
  author={Shao, Zishan and Wang, Yixiao and Wang, Qinsi and Jiang, Ting and Du, Zhixu and Ye, Hancheng and Zhuo, Danyang and Chen, Yiran and Li, Hai},
  journal={arXiv preprint arXiv:2508.01506},
  year={2025}
}

@article{wang2025vision,
  title={Vision-zero: Scalable vlm self-improvement via strategic gamified self-play},
  author={Wang, Qinsi and Liu, Bo and Zhou, Tianyi and Shi, Jing and Lin, Yueqian and Chen, Yiran and Li, Hai Helen and Wan, Kun and Zhao, Wentian},
  journal={arXiv preprint arXiv:2509.25541},
  year={2025}
}

@article{ye2025kvcomm,
  title={Kvcomm: Online cross-context kv-cache communication for efficient llm-based multi-agent systems},
  author={Ye, Hancheng and Gao, Zhengqi and Ma, Mingyuan and Wang, Qinsi and Fu, Yuzhe and Chung, Ming-Yu and Lin, Yueqian and Liu, Zhijian and Zhang, Jianyi and Zhuo, Danyang and others},
  journal={arXiv preprint arXiv:2510.12872},
  year={2025}
}
\bibliographystyle{icml2026}

%%%%%%%%%%%%%%%%%%%%%%%%%%%%%%%%%%%%%%%%%%%%%%%%%%%%%%%%%%%%%%%%%%%%%%%%%%%%%%%
%%%%%%%%%%%%%%%%%%%%%%%%%%%%%%%%%%%%%%%%%%%%%%%%%%%%%%%%%%%%%%%%%%%%%%%%%%%%%%%
% APPENDIX
%%%%%%%%%%%%%%%%%%%%%%%%%%%%%%%%%%%%%%%%%%%%%%%%%%%%%%%%%%%%%%%%%%%%%%%%%%%%%%%
%%%%%%%%%%%%%%%%%%%%%%%%%%%%%%%%%%%%%%%%%%%%%%%%%%%%%%%%%%%%%%%%%%%%%%%%%%%%%%%
\clearpage
\newpage
\appendix
\onecolumn

% \section*{Organization}
% In this Appendix, we provide in depth descriptions of the materials that are not covered in the main paper, and report additional experimental results. The document is organized as follows:

% Dot leader
\newcommand{\TOCdots}{\leaders\hbox{.\kern0.6pt}\hfill}

% One macro for both levels
\newcommand{\TOCentry}[3][0em]{%
  \noindent\hspace*{#1}%
  \hyperref[#2]{#3}%
  \TOCdots
  \hyperref[#2]{\pageref{#2}}%
  \par
}

% Convenience wrappers
\newcommand{\TOClineAuto}[2]{\TOCentry[0em]{#2}{#1}}
\newcommand{\TOCsublineAuto}[2]{\TOCentry[1.5em]{#2}{#1}}

% In document (where you want the manual ToC)
\section*{Appendix Contents}
\TOClineAuto{A Background}{app:background}

\TOClineAuto{B Task Description and Examples}{app:A}
\TOCsublineAuto{B.1 Fault Localization Task Examples}{app:A1}
\TOCsublineAuto{B.2 Functional Mapping Task Examples}{app:A2}
\TOCsublineAuto{B.3 Boundary Testing Task Examples}{app:A3}
\TOCsublineAuto{B.4 Counterfactual Reasoning Task Examples}{app:A4}

\TOClineAuto{C Dataset Curation}{app:B}
\TOCsublineAuto{C.1 Sample Collection Prompts}{app:B1}
\TOCsublineAuto{C.2 T2S Muti-hop Reasoning Datatset Construction Prompts}{app:B2}
\TOCsublineAuto{C.3 Question Type Templates}{app:B3}
\TOCsublineAuto{C.4 Human Evaluation Checklist}{app:B4}

\TOClineAuto{D Evaluation Details}{app:C}
\TOCsublineAuto{D.1 Experiment setup}{app:C1}
\TOCsublineAuto{D.2 API evaluation}{app:C2}
\TOCsublineAuto{D.3 Open source model evaluation}{app:C3}

\TOClineAuto{E Sample Examples}{app:D}
\TOCsublineAuto{E.1 Structure diagram examples}{app:D1}
\TOCsublineAuto{E.2 Full sample examples}{app:D2}

\TOClineAuto{F Additional Results and Analysis}{app:E}
\TOCsublineAuto{F.1 Experimental Setting}{app:E1}
\TOCsublineAuto{F.2 Additional Results}{app:E2}
\TOCsublineAuto{F.3 Observtaion and Insight}{app:E3}

% \begin{itemize}
%     \item \textbf{\ref{sec_appn_relatedWork}}- Related Work
%     \item \textbf{\ref{sec_appn_theoretical}}- Theoretical Supplement
%     \begin{itemize}
%         \item \textbf{\ref{sec_appn_angleimportance}} Theoretical Justification of Angle-Dependent Effects in LLMs
%         \item \textbf{\ref{sec_appn_attention}} Attention-Based Explanation of Layer-wise Angle Concentration Patterns
%         \item \textbf{\ref{sec_appn_neuron}} Neuron-Based Explanation of Data-wise Angle Concentration Patterns
%     \end{itemize}
%     \item \textbf{\ref{sec_appn_visual}}- Visualization Results
%     \begin{itemize}
%         \item \textbf{\ref{sec_appn_layer}}  Layer-wise Angle Concentration Patterns
%         \item \textbf{\ref{sec_appn_data}} Data-wise Angle Concentration Patterns
%     \end{itemize}
%     \item \textbf{\ref{sec_appn_gainrl}}- GAIN-RL Algorithm
%     \item \textbf{\ref{sec_appn_expersetup}}- Experimental Setup
%     \item \textbf{\ref{sec_appn_addExper}}- Additional Experimental Results
%     \begin{itemize}
%         \item \textbf{\ref{sec_appn_addExper_weightedsignal}} Performance of Weighted Signals
%         \item \textbf{\ref{sec_appn_addExper_single}} Model Performance on Single Task
%     \end{itemize}
%     \item \textbf{\ref{discussion}}- Discussion and Future Work
%     \end{itemize}

\section{Background}
\label{app:background}

\paragraph{Text-processing benchmarks.} Recent evaluations oriented toward real-world text workflows can generally be categorized into three types: Find, Fuse, and Form. Find-type datasets, such as MultiFieldQA~\cite{bai2024longbench} and Qasper~\cite{dasigi2021dataset}, focus on locating specific information within lengthy or specialized documents. Fuse-type datasets, including HotpotQA~\cite{yang2018hotpotqa}, 2WikiMultiHopQA~\cite{ho2020multihopqa}, and Musique~\cite{trivedi2022musique}, emphasize integrating and reasoning across multiple paragraphs or documents. Form-type datasets, exemplified by Qmsum~\cite{zhong2021qmsum} and GovReport~\cite{huang-etal-2021-efficient}, require generating specific outputs after reading texts. Although these benchmarks reflect realistic text-processing scenarios, tasks are commonly modeled as end-to-end "direct generation," lacking a unified, stable, and verifiable intermediate representation (IR). This limitation results in unstable evidence retrieval, difficult-to-control integration across evidence, and less-auditable generation outcomes.

\paragraph{Information structuring.} Indeed, prior research has demonstrated that introducing structured information or structured intermediate representations can significantly enhance model stability and performance in specific text-processing or reasoning tasks. For instance, Structure Guided Prompt~\cite{cheng2024structureguidedprompt} explicitly converts unstructured texts into graph structures, guiding models through graph-based multi-step reasoning and improving multi-step inference capabilities in zero-shot scenarios. Similarly, PDFTriage~\cite{saad2024pdftriage} targets structured documents (e.g., PDFs with chapters, tables, and layouts), emphasizing retrieval and question-answering guided by structural or content clues. However, these methods have primarily shown effectiveness in specific tasks or structural types, with inconsistent structural definitions and evaluation protocols, leaving a gap for a comprehensive evaluation framework covering a broader range of text types and structural forms.

\paragraph{Chain-of-Thought (CoT).} CoT prompting~\cite{wei2022chain, wang2025vision} showed that LLMs can be steered to produce a sequence of intermediate reasoning steps, which often improves performance on hard reasoning tasks by making the inference path explicit, and has been shown to be more effective than traditional acceleration approaches~\cite{wang2024coreinfer,kuo2025h,wang2025corematching} that rely on model-level modifications such as compression~\cite{wang2025dobi,wang2023mathnas,wang2023dgl}, quantization, or system-level optimizations~\cite{ye2025kvcomm,shao2025flashsvd,wang2025angles}. Follow-up work improved CoT’s reliability by sampling multiple reasoning paths and aggregating them, e.g., self-consistency selects the most consistent answer across diverse chains rather than trusting a single greedy trace.
More recent lines of work treat inference as structured search over intermediate states: Tree-of-Thoughts~\cite{yao2023treethoughts} explores and evaluates multiple candidate “thought” states with lookahead and backtracking, and Graph-of-Thoughts (GoT)~\cite{Besta_2024} further generalizes this to an arbitrary graph, enabling operations like merging partial solutions and using feedback loops via an explicit execution graph (generate/score/aggregate/transform) over nodes.
Importantly, these paradigms primarily structure the model’s reasoning process—their nodes represent solution states and edges represent dependencies among them—whereas Structure of Thought (SoT) (as you define it) structures the input text content into a graph of salient nodes and typed links to serve as a stable, task-agnostic intermediate representation for downstream text processing; thus, SoT is largely orthogonal to CoT/GoT and composable with them (e.g., GoT can search over multiple candidate SoT structures, while SoT can ground CoT/GoT reasoning in explicit evidence structure).

\section{Task Description and Examples}
\label{app:A}

\vspace{0.5em}
In this section, we provide a detailed guide to our task taxonomy. As shown in Tab.~\ref{tab:template_taxonomy}, we organize the questions into four categories: Fault Localization, Functional Mapping, Boundary Testing, and Counterfactual Reasoning. We then present representative question examples for each category to illustrate how benchmark items are written and formatted in practice. Each example is shown in the same boxed layout, including the question prompt, options when applicable, the reference answer, and a short rationale. These examples serve as a reference to understand the style and structure of the benchmark questions.

% ========= Appendix: FL Examples (Subsection + Box Pattern) =========
% Preamble requirements:
% \usepackage[most]{tcolorbox}
% \usepackage{xcolor}
% \usepackage{enumitem}
% \usepackage{tabularx}
% \usepackage{booktabs}

% --- Palette ---
\definecolor{BoxBorder}{RGB}{210,210,210}
\definecolor{BoxBG}{RGB}{250,250,250}
\definecolor{BoxTitleBG}{RGB}{245,245,245}
\definecolor{AccentFL}{RGB}{0,90,160}
\definecolor{AccentFM}{RGB}{182,132,0}    % mustard
\definecolor{AccentBD}{RGB}{0,132,92}     % green
\definecolor{AccentCF}{RGB}{118,86,176}   % purple

% --- Compact option lists ---
\setlist[itemize]{leftmargin=1.2em, itemsep=1.0pt, topsep=1.5pt}
\setlist[enumerate]{leftmargin=1.3em, itemsep=1.5pt, topsep=1.5pt}

% --- Small label chips inside box ---
\newcommand{\qtag}{\textbf{Question}}
\newcommand{\atag}{\textbf{Answer}}
\newcommand{\rtag}{\textbf{Reasoning}}

% Generic QA box style, accent color is passed in as #1
\tcbset{
 qabox/.style={
  enhanced,
  colback=BoxBG,
  colframe=BoxBorder,
  boxrule=0.6pt,
  arc=1.2mm,
  left=5mm, right=4mm, top=2mm, bottom=2mm,
  colbacktitle=BoxTitleBG,
  fonttitle=\bfseries,
  coltitle=black,
  breakable,
  before skip=6pt,
  after skip=10pt,
  fontupper=\scriptsize,
 },
 qaboxaccent/.style n args={1}{
  qabox,
  borderline west={1.2pt}{0pt}{#1},
 }
}

% --- Inside-box QA layout helper ---
\newcommand{\qaBlock}[3]{%
\noindent\qtag\par
#1
\vspace{6pt}

\noindent\atag\par
\textbf{#2}
\vspace{6pt}

\noindent\rtag\par
#3
}

% ========= Usage Pattern =========
% subsection (plain text) -> box (question/answer/reasoning)
% repeat

\subsection{Fault Localization Task Examples}
\label{app:A1}
Fault Localization focuses on pinpointing the minimal text span(s) that cause an error, inconsistency, or wrong output, such as the specific sentence/claim that breaks correctness. Its key characteristic is fine-grained evidence attribution: models must localize where the fault is and often distinguish true evidence from distractors or near-miss statements.

% ============================================================
\subsubsection*{\textbf{\color{AccentFL}{[FL1]}} Upstream Root Cause Identification}
\begin{itemize}
 \item \textbf{Task:} Given an observed failure at a node, identify which upstream components could plausibly be the root cause by tracing dependency paths backward from the anomaly.
\end{itemize}

\begin{tcolorbox}[qaboxaccent={AccentFL}, title={FL1 Example}]
\qaBlock
{If recent samples suddenly stop appearing in the Leaves of a zone, which of the following components could plausibly be the upstream root cause of that data loss (select all that apply)?

\medskip
\textbf{Options}
\begin{itemize}
 \item A) Leaf Routers
 \item B) Ingestion Routers
 \item C) Range Assigner
 \item D) Root Mixers
\end{itemize}}
{A, B, C}
{Data reach the Leaves through the ingestion path: Ingestion Routers forward records to Leaf Routers, which write them into the Leaves. The Range Assigner controls how data are mapped onto Leaves. Failures in any of these can prevent samples from being stored. Root Mixers only issue queries downstream of the Leaves and cannot cause missing stored samples.}
\end{tcolorbox}

% ============================================================
% =========================
% FL2
% =========================
\subsubsection*{\textbf{\color{AccentFL}{[FL2]}} Downstream Propagation Ordering}
\begin{itemize}
 \item \textbf{Task:} If a component is removed or fails, identify which downstream node will be affected at a specific position in the propagation chain by following the directed edges and counting hops.
\end{itemize}

\begin{tcolorbox}[qaboxaccent={AccentFL}, title={FL2 Example}]
\qaBlock
{Suppose strict treatment compliance eliminates the Non-adherence node from the causal pathway. Which component becomes the second downstream element to reflect this change after the initial effect on prolonged hyperglycemia?

\medskip
\textbf{Options}
\begin{itemize}
 \item A) Prolonged upward-arrow[glucose]
 \item B) Beta islet failure
 \item C) Ketosis
 \item D) T2DM + prior ketosis
\end{itemize}}
{B}
{Removing Non-adherence first prevents prolonged hyperglycemia as the immediate downstream effect. The next node in the chain is beta islet failure, reached after two hops, making it the second downstream element. Ketosis is one step further downstream, while T2DM + prior ketosis lies upstream and is unaffected.}
\end{tcolorbox}

% =========================
% FL3
% =========================
\subsubsection*{\textbf{\color{AccentFL}{[FL3]}} Minimal Cut Set Identification}
\begin{itemize}
 \item \textbf{Task:} Identify the smallest set or sets of nodes whose removal blocks all directed paths from a source to a target, ensuring no causal influence can pass through any remaining branch.
\end{itemize}

\begin{tcolorbox}[qaboxaccent={AccentFL}, title={FL3 Example}]
\qaBlock
{To guarantee that a change in the Keyrate can no longer affect Inflation through any of the model's three transmission channels, which of the following triplets of variables form approximately minimal cut sets (select all that apply)?

\medskip
\textbf{Options}
\begin{itemize}
 \item A) Real rate + Exchange rate + Inflation expectations
 \item B) Exchange rate + Output gap + Inflation expectations
 \item C) Real rate + Output gap + Exchange rate
 \item D) Real rate + Output gap + Inflation expectations
\end{itemize}}
{A, B}
{All influence from Keyrate reaches Inflation via three independent branches: (i) through Real rate, (ii) through Exchange rate, and (iii) through Inflation expectations. Option A removes one pivotal node from each branch. Option B blocks both shortcut branches and also breaks remaining longer chains by cutting Output gap. Options C and D each leave at least one shortcut path intact.}
\end{tcolorbox}

% =========================
% FL4
% =========================
\subsubsection*{\textbf{\color{AccentFL}{[FL4]}} Single Bottleneck or Mandatory Path Identification}
\begin{itemize}
 \item \textbf{Task:} Identify the unique component that lies on every path between request sources and the target node, making it a mandatory bottleneck that all flows must traverse.
\end{itemize}

\begin{tcolorbox}[qaboxaccent={AccentFL}, title={FL4 Example}]
\qaBlock
{Every write request, whether it originates in the master region or in a remote slave region, must traverse a particular component before the data can be recorded in any database. Which component acts as this mandatory bottleneck in the write path?

\medskip
\textbf{Options}
\begin{itemize}
 \item A) Master DB
 \item B) Leader Cache in the master region
 \item C) Leader Cache in the slave region
 \item D) Replication stream from the Master DB to the Slave DB
\end{itemize}}
{B}
{Local writes go through the master region leader cache before reaching the Master DB. Remote writes are first forwarded from the slave region leader cache to the same master region leader cache and only then committed. Therefore, the master region leader cache lies on every write path and is the mandatory bottleneck.}
\end{tcolorbox}

% =========================
% FL5
% =========================
\subsubsection*{\textbf{\color{AccentFL}{[FL5]}} Partial Failure with Compensation}
\begin{itemize}
 \item \textbf{Task:} When one expected intermediate outcome is missing but a related final outcome still holds, identify which upstream components in the failing branch could explain the observed pattern while allowing other branches to compensate.
\end{itemize}

\begin{tcolorbox}[qaboxaccent={AccentFL}, title={FL5 Example}]
\qaBlock
{Monitoring shows that consumers are not decreasing their consumption of the targeted foods, yet the population's overall diet quality is still improving. Which of the following upstream components could most plausibly account for this pattern (select all that apply)?

\medskip
\textbf{Options}
\begin{itemize}
 \item A) Taxes for food and beverages
 \item B) Target foods become relatively more expensive for consumers
 \item C) Target foods become relatively less expensive for consumers
 \item D) Consumers increase consumption of target foods
\end{itemize}}
{A, B}
{A failure in the tax pathway, either the tax itself or its immediate price increase effect, can prevent the decrease in consumption signal. Meanwhile, other pathways can still contribute to improved overall diet quality, allowing the final outcome to appear normal. Nodes in the subsidy branch do not explain the missing decrease consumption outcome.}
\end{tcolorbox}

% =========================
% FL6
% =========================
\subsubsection*{\textbf{\color{AccentFL}{[FL6]}} Independent Branch Identification}
\begin{itemize}
 \item \textbf{Task:} If one edge fails, identify which other relation remains unaffected because it lies on an independent branch that does not rely on the failed dependency.
\end{itemize}

\begin{tcolorbox}[qaboxaccent={AccentFL}, title={FL6 Example}]
\qaBlock
{If the switching: power adaption message sent by the DSO never reaches the EVSE owner, which of the following relations will remain unaffected because it belongs to an independent branch?

\medskip
\textbf{Options}
\begin{itemize}
 \item A) Flexibility transfer from the EVSE owner to the DSO
 \item B) Compensation payment from the energy supplier to the EVSE owner
 \item C) Flexibility-potential information from the EVSE owner to the DSO
 \item D) The switching: power adaption link itself
\end{itemize}}
{B}
{The failure lies on the control path DSO $\rightarrow$ EVSE owner. The monetary path energy supplier $\rightarrow$ EVSE owner (compensation payment) bypasses the DSO entirely, so it is independent of the disrupted control branch and remains unaffected.}
\end{tcolorbox}

% =========================
% FL7
% =========================
\subsubsection*{\textbf{\color{AccentFL}{[FL7]}} Feedback Loop Malfunction Diagnosis}
\begin{itemize}
 \item \textbf{Task:} Given an observed symptom of inefficient or degraded system behavior, identify which component in a feedback or control loop is most likely malfunctioning by reasoning about how the loop maintains system performance.
\end{itemize}

\begin{tcolorbox}[qaboxaccent={AccentFL}, title={FL7 Example}]
\qaBlock
{In the described coverage guided fuzzing process, if the system is observed to be repeatedly testing the same code paths without discovering new vulnerabilities, this suggests a failure in the exploration strategy. Which part of the process is most likely malfunctioning to cause this specific inefficiency?

\medskip
\textbf{Options}
\begin{itemize}
 \item A) The module responsible for mutating inputs.
 \item B) The mechanism that executes a single test case against the target application.
 \item C) The feedback loop that analyzes code coverage to guide input selection.
 \item D) The component responsible for detecting and logging crashes.
\end{itemize}}
{C}
{Coverage guided fuzzing continuously monitors and analyzes code coverage, then prioritizes inputs that lead to unexplored or less covered code paths. If this feedback mechanism fails, the system stops prioritizing new paths and repeatedly tests the same known paths, causing the observed inefficiency. Mutation issues would reduce input diversity but do not necessarily cause repeated path selection, execution failure would stop testing entirely, and crash detection failure would miss vulnerabilities but would not prevent exploration.}
\end{tcolorbox}

% =========================
% FL8
% =========================
\subsubsection*{\textbf{\color{AccentFL}{[FL8]}} Common Ancestor of Multiple Outputs}
\begin{itemize}
 \item \textbf{Task:} Given multiple output nodes with simultaneous anomalies, identify upstream components that could serve as a single common cause by having directed paths to all anomalous outputs.
\end{itemize}

\begin{tcolorbox}[qaboxaccent={AccentFL}, title={FL8 Example}]
\qaBlock
{Both the mass balance of the Greenland ice sheet and the sea-level contribution from thermal expansion are observed to shift unexpectedly. Which of the following components could serve as a single common upstream cause for both of these anomalies (select all that apply)?

\medskip
\textbf{Options}
\begin{itemize}
 \item A) Upper-ocean temperature anomaly
 \item B) Deeper-ocean temperature anomaly
 \item C) Anthropogenic CO$_2$ emissions
 \item D) SO$_2$ injections
\end{itemize}}
{A, C, D}
{Upper-ocean temperature anomalies directly drive both Greenland melting and thermal expansion. Anthropogenic CO$_2$ emissions influence atmospheric CO$_2$, which raises upper-ocean temperatures, giving a shared upstream path to both outputs. SO$_2$ injections also feed the climate sub-model by altering upper-ocean temperatures, affecting both outputs. Deeper-ocean temperature anomaly only feeds the thermal expansion branch and has no path to Greenland.}
\end{tcolorbox}

\subsection{Functional Mapping Task Examples}
\label{app:A2}

Functional Mapping asks models to map pieces of text to explicit functions/roles in a structured workflow, e.g., linking a requirement to the responsible component, step, or API, or mapping an observation to the operation that should handle it. It emphasizes role alignment and relationship grounding: correct answers require consistent many-to-many linking between textual units and predefined function slots.

% ============================================================
\subsubsection*{\textbf{\color{AccentFM}{[FM1]}} Alternative Implementations Identification}
\begin{itemize}
 \item \textbf{Task:} Identify which functional stage has multiple alternative tool or component implementations available.
\end{itemize}

\begin{tcolorbox}[qaboxaccent={AccentFM}, title={FM1 Example}]
\qaBlock
{Paper: \textit{DATC RDF 2019: Towards a Complete Academic Reference Design Flow}

\smallskip
Topology: EDA Toolchain and Design Flow Diagram

\medskip
In the described RDF 2019 flow, several stages are enhanced with specific tool integrations. For which of the following functions does the flow provide multiple alternative tool implementations, thereby implying that a selection between them is necessary?

\medskip
\textbf{Options}
\begin{itemize}
 \item A) Floorplan
 \item B) Detailed Placement
 \item C) Gate Sizing
 \item D) Clock Tree Synthesis
\end{itemize}}
{C}
{The flow lists multiple tools for Gate Sizing, namely Resizer and TritonSizer, which implies an explicit choice. The other functions are each associated with a single named tool in the description, so they do not indicate alternative implementations.}
\end{tcolorbox}

% ============================================================
\subsubsection*{\textbf{\color{AccentFM}{[FM2]}} Aggregation Point Identification}
\begin{itemize}
 \item \textbf{Task:} Identify the node that serves as a central aggregation point collecting inputs from multiple sources.
\end{itemize}

\begin{tcolorbox}[qaboxaccent={AccentFM}, title={FM2 Example}]
\qaBlock
{Paper: \textit{The Hadoop Distributed File System}

\smallskip
Topology: Storage or Networked System

\medskip
After the client pipelines data through several DataNodes, every DataNode sends a ``Blocks Received'' acknowledgment upstream. Which single component serves as the central aggregation point for all of these acknowledgments?

\medskip
\textbf{Options}
\begin{itemize}
 \item A) HDFS Client
 \item B) NameNode
 \item C) First DataNode in the pipeline
 \item D) Last DataNode in the pipeline
\end{itemize}}
{B}
{Each DataNode independently reports its received blocks status to the NameNode. Since multiple acknowledgments converge to the same node, the NameNode is the aggregator.}
\end{tcolorbox}

% ============================================================
\subsubsection*{\textbf{\color{AccentFM}{[FM3]}} Intermediate Buffer or Storage Identification}
\begin{itemize}
 \item \textbf{Task:} Identify components that act as intermediate storage or buffers between two stages of processing.
\end{itemize}

\begin{tcolorbox}[qaboxaccent={AccentFM}, title={FM3 Example}]
\qaBlock
{Paper: \textit{AccelTran: A Sparsity Aware Accelerator for Dynamic Inference}

\smallskip
Topology: Accelerator and Microarchitecture Block Diagram

\medskip
Which of the following components act as the on chip storage buffers that sit between the DMA controller fetches and the Processing Elements computations? Select all that apply.

\medskip
\textbf{Options}
\begin{itemize}
 \item A) Weight Buffer
 \item B) Control Block
 \item C) Mask Buffer
 \item D) Activation Buffer
\end{itemize}}
{A, C, D}
{The DMA brings weights, activations, and masks onto the chip, and these are held in the corresponding Weight Buffer, Activation Buffer, and Mask Buffer before being consumed by the Processing Elements. The Control Block schedules and coordinates, but it does not provide data storage between stages.}
\end{tcolorbox}

% ============================================================
\subsubsection*{\textbf{\color{AccentFM}{[FM4]}} Controller or Constraint Role Identification}
\begin{itemize}
 \item \textbf{Task:} Identify which components primarily constrain or govern other components rather than directly producing outputs or allocating resources.
\end{itemize}

\begin{tcolorbox}[qaboxaccent={AccentFM}, title={FM4 Example}]
\qaBlock
{Paper: \textit{The Governance and Implementation of the National Action Plan on AMR}

\smallskip
Topology: Institutional and Governance Framework Diagram

\medskip
Within the antimicrobial resistance governance framework, which governance areas mainly act as controllers, shaping or constraining the activities of other areas rather than directly delivering interventions or funds? Select all that apply.

\medskip
\textbf{Options}
\begin{itemize}
 \item A) Sustainability
 \item B) Policy Design
 \item C) Monitoring and Evaluation
 \item D) One Health Engagement
\end{itemize}}
{B, D}
{Policy Design sets rules, strategy, and coordination mechanisms that guide how other areas operate. One Health Engagement is positioned as a cross cutting influence that shapes multiple other governance areas. Sustainability is primarily about resources, while Monitoring and Evaluation primarily measures performance rather than constraining activities as the main role.}
\end{tcolorbox}

% ============================================================
\subsubsection*{\textbf{\color{AccentFM}{[FM5]}} Mediator or Conduit Identification}
\begin{itemize}
 \item \textbf{Task:} Identify the node that serves as the main conduit through which one component influence reaches another.
\end{itemize}

\begin{tcolorbox}[qaboxaccent={AccentFM}, title={FM5 Example}]
\qaBlock
{Paper: \textit{Leaning against the Wind Policies on Vietnam Economy with DSGE Model}

\smallskip
Topology: DSGE Sector and Agent Interaction Schematic

\medskip
Which agent serves as the main conduit through which lending by Banks ultimately influences Capital good producers?

\medskip
\textbf{Options}
\begin{itemize}
 \item A) Households
 \item B) Retailers
 \item C) Entrepreneurs
 \item D) Capital good producers
\end{itemize}}
{C}
{Banks lend to Entrepreneurs, and Entrepreneurs use those funds to purchase capital from Capital good producers. That path makes Entrepreneurs the mediating node that transmits the effect of bank lending to the producers.}
\end{tcolorbox}

% ============================================================
\subsubsection*{\textbf{\color{AccentFM}{[FM6]}} Monitoring or Evaluation Metric Identification}
\begin{itemize}
 \item \textbf{Task:} Identify which element functions primarily as a monitoring or assessment metric within the system flow.
\end{itemize}

\begin{tcolorbox}[qaboxaccent={AccentFM}, title={FM6 Example}]
\qaBlock
{Paper: \textit{The Utility of Combining the IAD and SES Frameworks}

\smallskip
Topology: Institutional and Governance Framework Diagram

\medskip
Which element in the IAD framework functions primarily as a monitoring or assessment metric, receiving the results of interactions and channeling feedback without itself directly generating new actions or outcomes?

\medskip
\textbf{Options}
\begin{itemize}
 \item A) Action situation
 \item B) Interactions
 \item C) Evaluative criteria
 \item D) Outcomes
\end{itemize}}
{C}
{Outcomes feed into Evaluative criteria, where results are judged, and those evaluations then feed back into contextual factors. This role is measurement and assessment rather than direct action generation or outcome production.}
\end{tcolorbox}

% ============================================================
\subsubsection*{\textbf{\color{AccentFM}{[FM7]}} Parallel Complementary Sub Functions Identification}
\begin{itemize}
 \item \textbf{Task:} Identify nodes that act as parallel complementary sub functions feeding directly into a target outcome.
\end{itemize}

\begin{tcolorbox}[qaboxaccent={AccentFM}, title={FM7 Example}]
\qaBlock
{Paper: \textit{Digital Public Infrastructure and Development: A World Bank Approach}

\smallskip
Topology: Logic Model and Theory of Change Diagram

\medskip
In the causal structure, which of the following nodes act as parallel complementary sub functions that directly channel their influence into the ``Financial Inclusion'' outcome, rather than acting as upstream precursors or downstream results? Select all that apply.

\medskip
\textbf{Options}
\begin{itemize}
 \item A) More competition among DFS providers
 \item B) Digital and financial literacy
 \item C) Participation in the digital economy
 \item D) Catalyze demand for DFS by individuals and MSMEs
\end{itemize}}
{B, D}
{Both Digital and financial literacy and Catalyze demand for DFS by individuals and MSMEs connect directly into Financial Inclusion as separate incoming branches, making them parallel contributors. More competition among DFS providers sits one step upstream, acting through the catalyze demand node, while Participation in the digital economy appears as a downstream consequence rather than a direct input to Financial Inclusion.}
\end{tcolorbox}

\subsection{Boundary Testing Task Examples}
\label{app:A3}
% ============================================================

Boundary Testing targets edge cases and threshold conditions in natural-language specifications, such as identifying when a rule applies, when it fails, and what inputs sit just inside/outside valid ranges. Its hallmark is precision around constraints: subtle wording, numeric ranges, exceptions, and conditional logic matter, and small changes can flip the correct outcome.

\subsubsection*{\textbf{\color{AccentBD}{[BT1]}} Feedback Link Elimination}
\begin{itemize}
 \item \textbf{Task:} Determine which specific feedback or regulatory link would be eliminated if a key intermediate component is removed.
\end{itemize}

\begin{tcolorbox}[qaboxaccent={AccentBD}, title={BT1 Example}]
\qaBlock
{Paper: \textit{Inclusion of the glucocorticoid receptor in a hypothalamic pituitary adrenal axis model reveals bistability}

\smallskip
Topology: Physiological Pathway / Axis Network

\medskip
In the HPA axis network, cortisol (O) inhibits downstream hormones only after binding to the glucocorticoid receptor (R). If R is completely absent so that the OR complex cannot form, which specific negative feedback link would be eliminated?

\medskip
\textbf{Options}
\begin{itemize}
 \item A) Inhibition of CRH (C) synthesis in the hypothalamus by cortisol (O)
 \item B) Inhibition of ACTH (A) production in the pituitary by cortisol (O)
 \item C) Up regulation of glucocorticoid receptor (R) synthesis driven by OR complex
 \item D) Activation of cortisol (O) release from the adrenal by ACTH (A)
\end{itemize}}
{B}
{Cortisol must bind R to form the OR complex before it can inhibit ACTH. Without R, this mediated feedback path cannot operate. The CRH inhibition does not require R, the OR driven receptor synthesis is a positive regulation rather than a negative feedback link, and ACTH stimulating cortisol is upstream of R.}
\end{tcolorbox}

% ============================================================
\subsubsection*{\textbf{\color{AccentBD}{[BT2]}} Narrowest Prerequisites or Conditional Gate}
\begin{itemize}
 \item \textbf{Task:} Identify which stage in a workflow is subject to the narrowest set of prerequisites or the most restrictive conditional gate.
\end{itemize}

\begin{tcolorbox}[qaboxaccent={AccentBD}, title={BT2 Example}]
\qaBlock
{Paper: \textit{The Federal Rulemaking Process: An Overview}

\smallskip
Topology: Institutional Decision / Policy Process Workflow

\medskip
Considering the exceptions described for different kinds of agencies and rules, which of the following steps in the federal rulemaking workflow is subject to the narrowest set of prerequisites?

\medskip
\textbf{Options}
\begin{itemize}
 \item A) Publication of Notice of Proposed Rulemaking
 \item B) OMB OIRA review of draft proposed rule
 \item C) Public comments
 \item D) Response to comments and development of draft final rule
\end{itemize}}
{B}
{OMB OIRA review occurs only when the rule is significant and the issuing body is not an independent regulatory agency, so it has a more restrictive gate than the other stages. The NPRM, comment, and response stages may be skipped in some cases, but they are not additionally limited by the same significance and agency type conditions.}
\end{tcolorbox}

% ============================================================
\subsubsection*{\textbf{\color{AccentBD}{[BT3]}} Bypass Path Identification}
\begin{itemize}
 \item \textbf{Task:} If a specific transition is blocked, identify which alternative pathways remain valid to reach the same destination.
\end{itemize}

\begin{tcolorbox}[qaboxaccent={AccentBD}, title={BT3 Example}]
\qaBlock
{Paper: \textit{Synergistic Catalysis in Heterobimetallic Complexes for CO2 Hydrogenation}

\smallskip
Topology: Catalytic Cycle / Mechanism State Graph

\medskip
Suppose the inner sphere transition state TS3 is completely blocked. Which of the following describe a remaining valid pathway by which the hydride complex [IrIII H] can still be converted into the formate complex [IrIII HCO2]? Select all that apply.

\medskip
\textbf{Options}
\begin{itemize}
 \item A) [IrIII H] $\to$ TS3 $\to$ [IrIII HCO2]
 \item B) [IrIII H] $\to$ TS2 $\to$ [IrIII HCO2]* $\to$ [IrIII HCO2]
 \item C) [IrIII H] $\to$ [IrIII(Base)]+ $\to$ [IrIII HCO2]
 \item D) [IrIII H] $\to$ [IrIII(eta2 H2)]PF6 $\to$ TS1 $\to$ [IrIII HCO2]
\end{itemize}}
{B}
{Blocking TS3 removes the direct inner sphere conversion. A remaining route is described via TS2 that forms the adduct [IrIII HCO2]*, which then rearranges to [IrIII HCO2]. The other sequences do not provide a complete directionally consistent path once TS3 is unavailable.}
\end{tcolorbox}

% ============================================================
\subsubsection*{\textbf{\color{AccentBD}{[BT4]}} Hedged or Stable Flow Identification}
\begin{itemize}
 \item \textbf{Task:} Identify which payment flow or connection is structurally protected against external shocks due to a stabilizing mechanism in the system.
\end{itemize}

\begin{tcolorbox}[qaboxaccent={AccentBD}, title={BT4 Example}]
\qaBlock
{Paper: \textit{Modeling Multiple Event Catastrophe Bond Prices Involving the Trigger Event Correlation, Interest, and Inflation Rates}

\smallskip
Topology: Securitization / Structured Finance

\medskip
A sudden spike in market interest rates threatens the returns on the short term securities held for the bond. Thanks to the SPV's LIBOR based floating swap, which of the following payment lines is most likely to keep its value stable despite that shock?

\medskip
\textbf{Options}
\begin{itemize}
 \item A) Principal repayment from Special Purpose Vehicle to Investor
 \item B) Premium transfer from Sponsor to Special Purpose Vehicle
 \item C) Investment income remittance from Trust Account to Special Purpose Vehicle
 \item D) Coupon payments from Special Purpose Vehicle to Investor
\end{itemize}}
{C}
{The LIBOR linked swap converts the Trust Account returns into floating payments tied to LIBOR before those proceeds are remitted onward. This buffering happens before the money reaches the SPV, so the Trust Account to SPV investment income flow is the most protected against an interest rate shock. The other flows are fixed transfers or are halted under certain conditions, and they are not directly stabilized by the swap mechanism.}
\end{tcolorbox}

% ============================================================
\subsubsection*{\textbf{\color{AccentBD}{[BT5]}} Invariant Connections Across Modes}
\begin{itemize}
 \item \textbf{Task:} Identify which connections are structurally present in both of two different operating modes.
\end{itemize}

\begin{tcolorbox}[qaboxaccent={AccentBD}, title={BT5 Example}]
\qaBlock
{Paper: \textit{M3DocRAG: Multi modal Retrieval for Multi page Multi document Understanding}

\smallskip
Topology: RAG Agent Tool Use Component Architecture

\medskip
Which of the following connections are active in both the closed domain and open domain operating modes of M3DocRAG? Select all that apply.

\medskip
\textbf{Options}
\begin{itemize}
 \item A) Visual Encoder (ColPali) $\to$ page visual embeddings
 \item B) Text Encoder (ColPali) $\to$ MaxSim relevance scorer
 \item C) Page visual embeddings $\to$ IVF (IVFFlat) approximate index
 \item D) Exact exhaustive search over a single document's pages $\to$ top K page selection
\end{itemize}}
{A, B}
{Both modes embed pages with the Visual Encoder and compute similarity using MaxSim, so A and B are present in both settings. The IVF based approximate index is specific to open domain retrieval, while the exact exhaustive search is specific to closed domain retrieval.}
\end{tcolorbox}

% ============================================================
\subsubsection*{\textbf{\color{AccentBD}{[BT6]}} Bypass Path After Edge Removal}
\begin{itemize}
 \item \textbf{Task:} After a direct link is eliminated, identify which remaining pathway most plausibly allows influence to flow between the same endpoints.
\end{itemize}

\begin{tcolorbox}[qaboxaccent={AccentBD}, title={BT6 Example}]
\qaBlock
{Paper: \textit{Governance and everyday adaptations? Examining the disconnect between planned and autonomous adaptation}

\smallskip
Topology: Institutional / Governance Framework Diagram

\medskip
Suppose the direct link in which Governance systems set procedural and distributive conditions for Adaptation action situations is eliminated. Which remaining pathway most plausibly allows Governance systems to continue exerting influence on the eventual Justice outcomes?

\medskip
\textbf{Options}
\begin{itemize}
 \item A) Governance systems $\to$ Adaptation action situations $\to$ Justice outcomes
 \item B) Governance systems $\to$ Actors $\to$ Adaptation action situations $\to$ Justice outcomes
 \item C) Governance systems $\to$ Ecological and technological systems $\to$ Environmental and technological parameters $\to$ Adaptation action situations $\to$ Justice outcomes
 \item D) Governance systems $\leftarrow$ Adaptation action situations $\to$ Actors $\to$ Justice outcomes
\end{itemize}}
{B}
{Even without the direct Governance to Adaptation edge, Governance systems still set rules for Actors. Those Actors participate in Adaptation action situations, which then produce Justice outcomes. This provides a plausible bypass route that preserves influence from Governance to the eventual outcome without relying on the removed link.}
\end{tcolorbox}

\subsection{Counterfactual Reasoning Task Examples}
\label{app:A4}

Counterfactual Reasoning evaluates whether models can reason under “what-if” modifications—changing a fact, constraint, or event—and predict how conclusions should change while keeping everything else fixed. The key feature is causal/structural sensitivity: models must separate invariant background from the altered premise and update only the consequences entailed by the counterfactual.

% ============================================================
\subsubsection*{\textbf{\color{AccentCF}{[CR1]}} Downstream Consequence of Broken Link}
\begin{itemize}
 \item \textbf{Task:} If a causal link fails to transmit its effect, identify which subsequent outcome in the pathway would be most directly undermined.
\end{itemize}

\begin{tcolorbox}[qaboxaccent={AccentCF}, title={CR1 Example}]
\qaBlock
{Paper: \textit{Healthy food prescription incentive programme for adults with type 2 diabetes who are experiencing food insecurity}

\smallskip
Topology: Logic Model / Theory of Change

\medskip
According to the described theory of change, reduced food insecurity and improved diet quality are key mediators for improving blood glucose levels. If a patient's diet quality improved but this specific change failed to translate into improved blood glucose levels (A1C), which subsequent outcome in the causal pathway would be most directly undermined?

\medskip
\textbf{Options}
\begin{itemize}
 \item A) A reduction in food insecurity
 \item B) A reduction in diabetes complications
 \item C) The integration of data from modelling studies
 \item D) An improvement in blood glucose levels (A1C)
\end{itemize}}
{B}
{The model implies a chain where improved diet quality should improve blood glucose levels, which then helps reduce diabetes complications. If the link from improved diet quality to improved blood glucose is broken, the most directly downstream outcome that loses its support is the reduction in complications. Food insecurity is an upstream parallel mediator, integration of data is not part of the patient outcome chain, and improved blood glucose is the intermediate step that is failing rather than the subsequent outcome.}
\end{tcolorbox}

% ============================================================
\subsubsection*{\textbf{\color{AccentCF}{[CR2]}} Largest Drop in Association Strength}
\begin{itemize}
 \item \textbf{Task:} If a specific edge is removed from a mediation model, identify which other association would show the largest drop in overall strength.
\end{itemize}

\begin{tcolorbox}[qaboxaccent={AccentCF}, title={CR2 Example}]
\qaBlock
{Paper: \textit{Structural equation modeling of social media addiction with eating behavior}

\smallskip
Topology: Path Diagram / SEM / Mediation Model

\medskip
Suppose researchers replace the Body Image construct with a version that keeps all of its original connections except that it no longer sends any influence to Restrained Eating (EB RE). With every other pathway left intact, which linkage in the model would you expect to show the largest drop in overall strength as a consequence of this modification?

\medskip
\textbf{Options}
\begin{itemize}
 \item A) Social Media Addiction $\to$ Emotional Eating (EB EE)
 \item B) Body Image $\to$ External Stimuli (EB ES)
 \item C) Social Media Addiction $\to$ Restrained Eating (EB RE)
 \item D) Social Media Addiction $\to$ Body Image
\end{itemize}}
{C}
{Removing the Body Image to Restrained Eating edge eliminates the indirect route from Social Media Addiction to Restrained Eating that passes through Body Image. That deletion reduces the total association for the Social Media Addiction to Restrained Eating linkage the most. The other listed connections do not rely on the removed edge, so their strength is not structurally diminished in the same way.}
\end{tcolorbox}

% ============================================================
\subsubsection*{\textbf{\color{AccentCF}{[CR3]}} Structural Consequences of Adding an Edge}
\begin{itemize}
 \item \textbf{Task:} If a new direct link is introduced, identify which structural consequences necessarily follow.
\end{itemize}

\begin{tcolorbox}[qaboxaccent={AccentCF}, title={CR3 Example}]
\qaBlock
{Paper: \textit{SuperFlow: A Fully Customized RTL to GDS Design Automation Flow}

\smallskip
Topology: EDA Toolchain / Design Flow Diagram

\medskip
Suppose a new direct link is introduced that allows the AQFP based netlist to feed straight into the circuit layout generation stage, bypassing the normal place and route step. Based purely on this topological change, which of the following statements are necessarily true? Select all that apply.

\medskip
\textbf{Options}
\begin{itemize}
 \item A) The circuit layout generation stage can now be reached even if the place and route step is skipped.
 \item B) The place and route step is completely eliminated from every possible path that leads to the final GDSII layout.
 \item C) The Design Rule Check (DRC) may now receive a layout that never passed through the place and route step.
 \item D) Logic synthesis (Yosys) is no longer required to obtain the GDSII layout.
\end{itemize}}
{A, C}
{The new shortcut creates an additional path where circuit layout generation is reachable directly from the AQFP netlist, so A becomes true. Because DRC is downstream of layout generation, C also becomes true since DRC can now be reached through a path that never traverses place and route. The original path that includes place and route still exists, so B does not necessarily follow. Logic synthesis is still required to produce the netlist inputs, so D is not implied by the added edge.}
\end{tcolorbox}

% ============================================================
\subsubsection*{\textbf{\color{AccentCF}{[CR4]}} Consequence of Blocking Parallel Feedback Path}
\begin{itemize}
 \item \textbf{Task:} If a feedback pathway operating in parallel with a primary mechanism is blocked, identify the most direct structural consequence.
\end{itemize}

\begin{tcolorbox}[qaboxaccent={AccentCF}, title={CR4 Example}]
\qaBlock
{Paper: \textit{Human Investigations into the Arterial and Cardiopulmonary Baroreflexes during Exercise}

\smallskip
Topology: Homeostatic Feedback Control Loop

\medskip
The text identifies the exercise pressor reflex as a feedback mechanism. If this entire feedback pathway were selectively blocked, what would be the most direct structural consequence on the system that adjusts cardiovascular function during exercise?

\medskip
\textbf{Options}
\begin{itemize}
 \item A) Central Command would be unable to regulate baroreflex resetting.
 \item B) Cardiopulmonary baroreceptors would take over as the primary regulatory input.
 \item C) The feed forward regulation from Central Command would continue, but without the modulatory influence from the reflex.
 \item D) The entire system for neural cardiovascular adjustment to exercise would cease to function.
\end{itemize}}
{C}
{Central Command is described as a feed forward driver that can continue to regulate cardiovascular responses. The exercise pressor reflex is a parallel feedback pathway that modulates the response. Blocking it removes that modulatory feedback while leaving the primary feed forward route intact, making C the most direct structural consequence.}
\end{tcolorbox}

% ============================================================
\subsubsection*{\textbf{\color{AccentCF}{[CR5]}} Paths Blocked by Removing Conditioning}
\begin{itemize}
 \item \textbf{Task:} If a collider node is no longer conditioned on, identify which influence chains would no longer transmit association.
\end{itemize}

\begin{tcolorbox}[qaboxaccent={AccentCF}, title={CR5 Example}]
\qaBlock
{Paper: \textit{Use of directed acyclic graphs (DAGs) to identify confounders}

\smallskip
Topology: Causal DAG

\medskip
Conditioning on the Mediator (M) currently opens the collider at M. If we instead leave M unconditioned, which of the following influence chains would no longer transmit association? Select all that apply.

\medskip
\textbf{Options}
\begin{itemize}
 \item A) The mediated causal path Exposure $\to$ Mediator $\to$ Outcome
 \item B) The collider path Exposure $\to$ Mediator $\leftarrow$ MOC $\to$ Outcome
 \item C) The confounding path Observed confounder $\to$ Exposure
 \item D) The collider link Exposure $\to$ Mediator $\leftarrow$ MOC (association between Exposure and MOC)
\end{itemize}}
{B, D}
{When M is not conditioned on, it functions as a collider and blocks any association path that must pass through the converging arrows at M. The paths in B and D rely on traversing the collider structure Exposure $\to$ M $\leftarrow$ MOC, so they are closed without conditioning. The standard mediated path Exposure $\to$ M $\to$ Outcome and the confounding link to Exposure do not require opening the collider, so they remain available.}
\end{tcolorbox}

% ============================================================
\subsubsection*{\textbf{\color{AccentCF}{[CR6]}} Surviving Routes When Node Is Clamped}
\begin{itemize}
 \item \textbf{Task:} If a mediator node is experimentally held constant, identify which structural routes can still convey an effect to the outcome.
\end{itemize}

\begin{tcolorbox}[qaboxaccent={AccentCF}, title={CR6 Example}]
\qaBlock
{Paper: \textit{The chain mediating role of interest and physical activity level}

\smallskip
Topology: Path Diagram / SEM / Mediation Model

\medskip
Researchers raise PE teacher autonomy support but keep pupils' physical activity levels fixed at a constant value. According to the mediation model, through which of the following routes can this manipulation still influence pupils' physical and mental health? Select all that apply.

\medskip
\textbf{Options}
\begin{itemize}
 \item A) PE teacher autonomy support $\to$ physical activity levels $\to$ physical and mental health
 \item B) PE teacher autonomy support $\to$ physical and mental health
 \item C) PE teacher autonomy support $\to$ interest $\to$ physical activity levels $\to$ physical and mental health
 \item D) PE teacher autonomy support $\to$ interest $\to$ physical and mental health
\end{itemize}}
{B, D}
{Clamping physical activity levels prevents any causal effect from propagating through that node. Routes A and C necessarily pass through physical activity levels, so they are blocked. The direct path from autonomy support to health and the route through interest alone both avoid the fixed node, so B and D can still transmit an effect.}
\end{tcolorbox}

% ============================================================
\subsubsection*{\textbf{\color{AccentCF}{[CR7]}} Invariant Output Under Source Change}
\begin{itemize}
 \item \textbf{Task:} If the source of a component's input is changed to a different upstream pathway, identify which downstream component's activation would remain essentially unchanged.
\end{itemize}

\begin{tcolorbox}[qaboxaccent={AccentCF}, title={CR7 Example}]
\qaBlock
{Paper: \textit{The Renin Angiotensin System in Cardiovascular Autonomic Control}

\smallskip
Topology: Physiological Pathway / Axis Network

\medskip
Suppose the pathway is re engineered so that all Angiotensin (1 7) is produced exclusively from Angiotensin I (via Angiotensin (1 9)) instead of from Angiotensin II. Assuming no other reactions are altered, which of the following components is expected to show the least change in its level of activation?

\medskip
\textbf{Options}
\begin{itemize}
 \item A) Angiotensin (1 9)
 \item B) AT1R
 \item C) Mas R
 \item D) Angiotensin A
\end{itemize}}
{C}
{Both the original and modified production routes converge at Angiotensin (1 7). Mas R is directly downstream of Angiotensin (1 7), so its activation depends primarily on the presence of Angiotensin (1 7) rather than on which upstream precursor supplied it. By contrast, Angiotensin (1 9) is introduced in the new route, AT1R is driven mainly by Angiotensin II and related peptides, and Angiotensin A formation depends on Angiotensin II, so those quantities change under the source swap.}
\end{tcolorbox}

\section{Dataset Curation}
In this section, we provide a detailed breakdown of the data curation pipeline and prompts used in this process. We first present the specific prompts and multi-step workflows used by our automated 'Dataset Builder' to source papers, normalize data schemas, and extract structural graphs (Sec.~\ref{app:B1}). Next, we document the prompts used for question generation and automated quality control (Sec.~\ref{app:B2}). Additionally, we organize the question type templates used for question generation (Sec.~\ref{app:B3}). Finally, we outline the human evaluation checklists and protocols used to ensure the validity and accuracy of the dataset (Sec.~\ref{app:B4}). 

\label{app:B}

% ============================================================
% Prompt extraction for (1) dataset builder and (2) QA generator
% ============================================================
% Requires:
% \usepackage{xcolor}
% \usepackage{tcolorbox}
% \tcbuselibrary{breakable}

% --- Colors (reuse yours if already defined) ---
\definecolor{BoxBG}{RGB}{250,250,250}
\definecolor{BoxBorder}{RGB}{210,210,210}
\definecolor{BoxTitleBG}{RGB}{245,245,245}

\definecolor{AccentSearch}{RGB}{64,64,64}
\definecolor{AccentNorm}{RGB}{64,64,64}
\definecolor{AccentVerify}{RGB}{64,64,64}
\definecolor{AccentVision}{RGB}{64,64,64}
\definecolor{AccentQA}{RGB}{64,64,64}
\definecolor{AccentQC}{RGB}{64,64,64}

% --- Box style (smaller font inside) ---
\tcbset{
 promptbox/.style={
  enhanced,
  colback=BoxBG,
  colframe=BoxBorder,
  boxrule=0.6pt,
  arc=1.2mm,
  left=5mm, right=4mm, top=2mm, bottom=2mm,
  borderline west={1.2pt}{0pt}{AccentSearch},
  colbacktitle=BoxTitleBG,
  fonttitle=\bfseries,
  coltitle=black,
  breakable,
  before skip=6pt,
  after skip=10pt,
  fontupper=\scriptsize,
  title={#1}
 }
}

% Optional helper: per box accent override
\newcommand{\setpromptaccent}[1]{%
  \tcbset{promptbox/.style={
    enhanced,
    colback=BoxBG,
    colframe=BoxBorder,
    boxrule=0.6pt,
    arc=1.2mm,
    left=5mm, right=4mm, top=2mm, bottom=2mm,
    borderline west={1.2pt}{0pt}{#1},
    colbacktitle=BoxTitleBG,
    fonttitle=\bfseries,
    coltitle=black,
    breakable,
    before skip=6pt,
    after skip=10pt,
    fontupper=\scriptsize,
    title={##1}
  }}%
}

% ============================================================
\subsection{Sample Collection Prompts}
\label{app:B1}
This section details the automated pipeline used to source and structure the raw data for our dataset. The process is divided into four sequential steps: (1) Candidate Search, where the model identifies relevant papers and figures; (2) Schema Normalization, which converts raw outputs into strict JSON formats; (3) Caption Consistency, which verifies that retrieved metadata matches the specific figure; and (4) Graph Extraction, where the visual diagram is parsed into a structured node-link graph representation. The specific system prompts for each stage are provided below.
% ============================================================

\subsection*{Step 1: Paper Search}
\begin{itemize}
 \item \textbf{Goal:} Find one strong paper and one exact figure identifier for a topology category.
\end{itemize}

\setpromptaccent{AccentSearch}
\begin{tcolorbox}[promptbox={Step 1 Prompt: Paper Search}]
\noindent\textbf{\color{AccentSearch}Task}\par
You are curating ONE high quality sample for the SciStruct benchmark.

\medskip
\noindent\textbf{\color{AccentSearch}Goal}\par
Find ONE strong Algorithm or AI research paper that contains a structural node link diagram matching the category:
\begin{itemize}
 \item Topology category: \texttt{\{category\}}
 \item Category description: \texttt{\{cat\_desc\}}
\end{itemize}

\medskip
\noindent\textbf{\color{AccentSearch}Hard constraints}\par
\begin{enumerate}
 \item \textbf{Deduplication:} Do not select any paper in \texttt{[\{already\_collected\_str\}]}

 \item \textbf{Top tier source:} Prefer strong venues and widely cited reports. Provide a stable link.

 \item \textbf{Strict diagram validity:} The chosen figure must be representable as a single connected node link graph $G=(V,E)$.
 Nodes must be labeled blocks with readable text. Links must be explicit connectors indicating flow or dependency.
 Reject plots, tables, photos, qualitative samples, or disconnected mini diagrams.

 \item \textbf{Category semantic match:} The diagram must match the node and link semantics in \texttt{\{cat\_desc\}}.

 \item \textbf{Exact figure ID:} Verify the caption explicitly starts with \texttt{Figure X} or \texttt{Fig.\ X}. If uncertain, discard.
\end{enumerate}

\medskip
\noindent\textbf{\color{AccentSearch}Output format (JSON only)}\par
Return exactly one JSON object (no markdown, no extra commentary):
\begin{quote}\ttfamily
\{\\
\ \ "paper\_metadata": \{\\
\ \ \ \ "title": "Full Paper Title",\\
\ \ \ \ "paper\_link": "Direct link to the paper",\\
\ \ \ \ "topology\_type": "\{category\}"\\
\ \ \ \},\\
\ \ "figure": \{\\
\ \ \ \ "figure\_id": "Figure X",\\
\ \ \ \ "figure\_caption\_snippet": "Copy one to two caption sentences (30 to 80 words)"\\
\ \ \ \},\\
\ \ "selection\_rationale": "One to three sentences: validity and semantics match",\\
\ \ "figure\_id\_verification": "One to two sentences: how the figure ID was verified"\\
\}\\
\end{quote}
\end{tcolorbox}

\subsection*{Step 2: Strict Schema Normalization}
\begin{itemize}
 \item \textbf{Goal:} Rewrite the draft object into a strict JSON schema.
\end{itemize}

\setpromptaccent{AccentNorm}
\begin{tcolorbox}[promptbox={Step 2 Prompt: Strict JSON Rewriter}]
\noindent You are a strict JSON rewriter.

\medskip
\noindent Input is a draft object that may miss fields or contain noisy strings.
Rewrite into exactly one JSON object with this schema:

\begin{quote}\ttfamily
\{\\
\ \ "paper\_metadata": \{\\
\ \ \ \ "title": "Full Paper Title",\\
\ \ \ \ "paper\_link": "Direct link to the paper",\\
\ \ \ \ "topology\_type": "\{category\}"\\
\ \ \ \},\\
\ \ "figure": \{\\
\ \ \ \ "figure\_id": "Figure X",\\
\ \ \ \ "figure\_caption\_snippet": "short snippet containing the label and diagram context"\\
\ \ \ \},\\
\ \ "selection\_rationale": "One to three sentences: validity and semantics match",\\
\ \ "figure\_id\_verification": "One to two sentences: how the figure ID was verified"\\
\}\\
\end{quote}

\medskip
\noindent Rules:
\begin{itemize}
 \item Keep URLs as is. Do not invent a new paper.
 \item Ensure \texttt{topology\_type} equals \texttt{\{category\}} exactly.
 \item Return JSON only.
\end{itemize}

\medskip
\noindent Draft:
\begin{quote}\ttfamily
\{draft\_obj\_json\}
\end{quote}
\end{tcolorbox}

\subsection*{Step 3: Caption Consistency Check}
\begin{itemize}
 \item \textbf{Goal:} Confirm the web snippet plausibly refers to the same caption as the PDF extracted caption text.
\end{itemize}

\setpromptaccent{AccentVerify}
\begin{tcolorbox}[promptbox={Step 3 Prompt: Lenient Caption Consistency Checker}]
\noindent You are a lenient caption consistency checker.

\medskip
\noindent We have:
\begin{itemize}
 \item \texttt{fig\_id}: \texttt{\{fig\_id\}}
 \item \texttt{model\_claim\_snippet}: a short or partial snippet produced during web search
 \item \texttt{true\_caption}: text extracted near the real \texttt{Figure} or \texttt{Fig.} label in the PDF
\end{itemize}

\medskip
\noindent Goal:\par
Decide whether \texttt{model\_claim\_snippet} is plausibly referring to the same figure caption as \texttt{true\_caption}.
Do not penalize incompleteness. If unsure, choose match true.

\medskip
\noindent Return mismatch only if clearly unrelated, for example:
\begin{enumerate}
 \item The snippet refers to a different figure or table number
 \item Topics are obviously different with near zero shared technical terms
 \item The snippet is clearly from a different paper and not about the same caption
\end{enumerate}

\medskip
\noindent Return JSON only:
\begin{quote}\ttfamily
\{\\
\ \ "match": true or false,\\
\ \ "reason": "short explanation"\\
\}\\
\end{quote}

\medskip
\noindent model\_claim\_snippet:\par
\texttt{\{model\_snippet\}}

\medskip
\noindent true\_caption:\par
\texttt{\{true\_caption\}}
\end{tcolorbox}

\subsection*{Step 4: Structural Validity Check}
\begin{itemize}
 \item \textbf{Goal:} Decide if the diagram is representable as a clean node link graph, then extract nodes and links.
\end{itemize}

\setpromptaccent{AccentVision}
\begin{tcolorbox}[promptbox={Step 4 Prompt: Structural Validity Check}]
\noindent Role: You are a professional Visual Graph Analyzer and Data Engineer.

\medskip
\noindent Task:
\begin{enumerate}
 \item Decide whether the provided structural diagram can be represented well as a simple node link graph JSON.
 \item If feasible, convert it into the required JSON format.
\end{enumerate}

\medskip
\noindent Feasibility criteria (strict):
\begin{itemize}
 \item Primarily a node link structural diagram such as a flowchart, architecture, pipeline, or logic map
 \item Most nodes have meaningful labels, not anonymous placeholders
 \item Links reflect real relations or flows and direction can be inferred
 \item One connected component
 \item Reject plots, tables, photos, qualitative grids, or math only figures
 \item Reject if unreadable or too overloaded for reliable extraction
\end{itemize}

\medskip
\noindent If the diagram fails feasibility, return only:
\begin{quote}\ttfamily
\{\\
\ \ "ok": false,\\
\ \ "reason": "one to two sentences"\\
\}\\
\end{quote}

\medskip
\noindent If the diagram passes, extraction rules:
\begin{itemize}
 \item Nodes: labeled blocks or entities with text, each with a unique id and label
 \item Links: directed edges from source to target, with optional label (empty string if none)
 \item Ignore decorative elements and background noise
 \item If a link is bidirectional, emit two directed links
\end{itemize}

\medskip
\noindent Output format for pass case (JSON only):
\begin{quote}\ttfamily
\{\\
\ \ "ok": true,\\
\ \ "reason": "one to two sentences",\\
\ \ "graph": \{\\
\ \ \ \ "nodes": [\{ "id": "n1", "label": "Node Text" \}],\\
\ \ \ \ "links": [\{ "source": "n1", "target": "n2", "label": "" \}]\\
\ \ \ \}\\
\}\\
\end{quote}
\end{tcolorbox}

% ============================================================
\subsection{T2S Muti-hoop Reasoning Dataset Construction Prompts}
\label{app:B2}
Once the structural graph is extracted, we employ a three-stage prompting strategy to create and validate benchmark questions. First, the Question Generation prompt instantiates a specific reasoning template (see Sec.~\ref{app:B3}) grounded in the text and diagram. This is followed by two rigorous quality control steps: a Diagram Consistency Check to ensure the question and answer are supported by the visual topology, and a Text-Only Solvability Check to verify that the answer can be derived via reasoning from the text description alone, without relying on visual cues.
% ============================================================

\subsection*{Step 1: Class Specific Question Generation}
\begin{itemize}
 \item \textbf{Goal:} Generate one multiple choice question for a target reasoning class grounded in the diagram topology and text paragraph.
\end{itemize}

\setpromptaccent{AccentQA}
\begin{tcolorbox}[promptbox={QA Prompt: General Rules and Required JSON Schema}]
\noindent You are designing benchmark questions for SciStruct (text only evaluation about a system diagram).

\medskip
\noindent You are given:
\begin{itemize}
 \item A reference diagram image used only to align the question to the true topology. Do not mention the image.
 \item A text paragraph extracted from the paper describing that diagram.
\end{itemize}

\medskip
\noindent Core requirements:
\begin{itemize}
 \item Multiple choice only, single select or multi select, exactly four options A to D
 \item Grounding is strict: stem and all options must reference concrete nodes, edges, relations, or topology structures present
 \item Text only solvability: the correct answer must be derivable from the text alone via structural reasoning
 \item Non triviality is strict: require at least two reasoning steps, not a one hop lookup
 \item Distractors must be plausible but topologically wrong
 \item Randomize answer position across A to D
 \item Do not mention Figure, image, or as shown
 \item Math must use Markdown math like \texttt{\$x \\to y\$}
\end{itemize}

\medskip
\noindent Template compliance:
\begin{itemize}
 \item Generate exactly one question for the target class
 \item Choose exactly one template among eight templates for that class
 \item The chosen \texttt{template\_id} must match the reasoning structure in \texttt{analysis\_plan}
\end{itemize}

\medskip
\noindent Return JSON only with schema:
\begin{quote}\ttfamily
\{\\
\ \ "question\_class": "\{TARGET\_CLASS\}",\\
\ \ "template\_id": "\{ONE\_TEMPLATE\_ID\}",\\
\ \ "type": "single" or "multi",\\
\ \ "analysis\_plan": \{\\
\ \ \ \ "target\_query": "...",\\
\ \ \ \ "key\_nodes": ["..."],\\
\ \ \ \ "key\_edges\_or\_relations": ["..."],\\
\ \ \ \ "two\_hop\_witness": "...",\\
\ \ \ \ "why\_not\_one\_hop": "..."\\
\ \ \ \},\\
\ \ "question": "...",\\
\ \ "options": [\\
\ \ \ \ \{ "id":"A", "text":"..." \},\\
\ \ \ \ \{ "id":"B", "text":"..." \},\\
\ \ \ \ \{ "id":"C", "text":"..." \},\\
\ \ \ \ \{ "id":"D", "text":"..." \}\\
\ \ \ \ ],\\
\ \ "answer": ["B"] or ["B","D"],\\
\ \ "reason": "one to three sentences"\\
\}\\
\end{quote}

\medskip
\noindent Hard constraints on \texttt{analysis\_plan}:
\begin{itemize}
 \item Include at least two distinct relations, or a branch merge bypass or condition argument
 \item Include an explicit two hop witness such as $A \\to M \\to Z$, or an equivalent multi step structural argument
 \item If you cannot find a two hop witness from the text, regenerate a different question
\end{itemize}
\end{tcolorbox}

\subsection*{Step 2: Diagram Grounded Quality Control}
\begin{itemize}
 \item \textbf{Goal:} Verify the question stem and provided answer are fully supported by the diagram topology and remain non trivial.
\end{itemize}

\setpromptaccent{AccentQC}
\begin{tcolorbox}[promptbox={QC Prompt: Image Grounded Diagram Consistency Checker}]
\noindent You are a high strictness diagram consistency checker for SciStruct.

\medskip
\noindent You are given:
\begin{itemize}
 \item A reference diagram image
 \item A candidate multiple choice question JSON including the provided answer
\end{itemize}

\medskip
\noindent Primary goal:
Verify the question and the provided correct answer are fully supported by the diagram.
If pass true, explain the diagram evidence.
If pass false, explain the mismatch.

\medskip
\noindent Checks:
\begin{enumerate}
 \item Stem grounding: stem refers to relations that exist in the diagram
 \item Answer verification: each answer option is correct under the diagram
 \item Anti triviality: fail if it is a one hop neighbor lookup
\end{enumerate}

\medskip
\noindent Return JSON only:
\begin{quote}\ttfamily
\{\\
"pass": true or false,\\
"verdict": "one to two sentences",\\
"diagram\_alignment": \{\\
\ \ "stem\_supported": true or false,\\
\ \ "stem\_evidence": ["..."],\\
\ \ "answer\_supported": true or false,\\
\ \ "answer\_evidence": ["..."]\\
\ \ \},\\
"errors": [] or ["..."]\\
\}\\
\end{quote}

\medskip
\noindent Candidate:\par
\texttt{\{question\_json\}}
\end{tcolorbox}

\subsection*{Step 3: Text Only Solvability Quality Control}
\begin{itemize}
 \item \textbf{Goal:} Verify the question and answer are derivable from the text paragraph alone using multi step structural reasoning.
\end{itemize}

\setpromptaccent{AccentQC}
\begin{tcolorbox}[promptbox={QC Prompt: Text Only Solvability Checker}]
\noindent You are a high strictness text only solvability checker for SciStruct.

\medskip
\noindent You are given:
\begin{itemize}
 \item A text paragraph describing a system
 \item A candidate multiple choice question JSON including the provided answer
\end{itemize}

\medskip
\noindent Primary goal:
Verify the question and provided answer can be derived from the text alone.
If unsure, set pass false.

\medskip
\noindent Checks:
\begin{enumerate}
 \item Answer derivability with at least two step reasoning over relations
 \item Not keyword lookup: fail if answer is a direct phrase match without structural inference
 \item Visual only dependency: fail if required facts are only in the diagram
 \item Anti guessing: fail if correct answer is guessable by surface semantics alone
 \item Anti leakage: fail if ordinal cues like Stage or Step reveal topology ordering
\end{enumerate}

\medskip
\noindent Return JSON only:
\begin{quote}\ttfamily
\{\\
"pass": true or false,\\
"verdict": "one to two sentences",\\
"text\_support": \{\\
\ \ "answer\_supported": true or false,\\
\ \ "conclusion": "reasoning chain or missing evidence"\\
\ \ \},\\
"nontrivial\_ok": true or false,\\
"nontrivial\_analysis": "...",\\
"errors": [] or ["..."]\\
\}\\
\end{quote}

\medskip
\noindent TEXT (markdown):\par
\texttt{\{extracted\_paragraph\_md\}}

\medskip
\noindent Candidate:\par
\texttt{\{question\_json\}}
\end{tcolorbox}

% ============================================================
\subsection{Question Type Templates}
\label{app:B3}
To ensure diverse and rigorous reasoning requirements, we utilize a library of parameterized templates corresponding to specific reasoning classes (e.g., Fault Localization, Functional Mapping). Each class contains a set of logic templates (e.g., 'Bottleneck Identification' or 'Feedback Loop Failure') that define the core reasoning task. The prompt blocks below list the logic definitions and constraints used to guide the model during the question generation phase described in Sec.~\ref{app:B2}.

\setpromptaccent{AccentQA}

% ------------------------------------------------------------
% Fault Localization (FL)
% ------------------------------------------------------------
\begin{tcolorbox}[promptbox={Fault Localization Template}]
\noindent This block is provided only when \texttt{target\_class = "Fault Localization"}.

\medskip
\noindent Allowed template ids:
[FL-1,FL-2,FL-3,FL-4,FL-5,FL-6,FL-7,FL-8]

\medskip
\noindent Templates (choose exactly one \texttt{template\_id}):

\medskip
\noindent \textbf{FL-1 Upstream Root-Cause Set (Multi-select)}\par
\noindent \textbf{Core logic:} Given an abnormality in a key output [Z] (too high/too low/missing), ask which components or variables could plausibly be upstream on a directed causal or dependency path leading to Z.\par
\noindent \textbf{Correct options:} Select 2 to 4 nodes that have at least one directed path to Z (mix direct parents and more distant ancestors). Ensure evidence is spread across multiple sentences or clauses.\par
\noindent \textbf{Distractors:}
\begin{enumerate}
 \item parallel branch nodes that do not reach Z
 \item downstream effects of Z
 \item nodes only connected under a condition not satisfied in the stem
 \item same domain nodes with similar wording but wrong topology
\end{enumerate}

\medskip
\noindent \textbf{FL-2 Secondly or Thirdly Affected Downstream (Single- or Multi-select)}\par
\noindent \textbf{Core logic:} If [A] fails, stops responding, or is removed, which components are the secondly or thirdly impacted ?\par
\noindent \textbf{Correct option:} the second or third component is affected.\par
\noindent \textbf{Distractors:} direct successor, parallel branch, bypass protected node.

\medskip
\noindent \textbf{FL-3 Minimal Cut Set for Failure (Multi-select, Hard)}\par
\noindent \textbf{Core logic:} To make [Z] impossible to produce (or guaranteed to fail), which option sets are sufficient and approximately minimal cut sets that disconnect all paths to Z?\par
\noindent \textbf{Correct options:} 2 to 3 small node sets or edge sets that, if removed, break every path from relevant sources to Z (true cut sets).\par
\noindent \textbf{Distractors:} sets that break only one path while leaving an alternative; sets that include redundant nodes (not minimal); sets containing only downstream nodes that do not prevent generation.

\medskip
\noindent \textbf{FL-4 Bottleneck or Dominator Node (Single-select)}\par
\noindent \textbf{Core logic:} Which node is the strongest bottleneck between [S] and [Z] (most paths must go through it; its failure knocks out multiple routes)?\par
\noindent \textbf{Correct option:} a dominator or articulation like node shared by many paths.\par
\noindent \textbf{Distractors:} nodes in only one branch; nodes at same depth but not shared; nodes in a feedback loop that does not dominate routes to Z.

\medskip
\noindent \textbf{FL-5 Abnormal Intermediate but Output Still Normal (Single- or Multi-select)}\par
\noindent \textbf{Core logic:} You observe intermediate [M] is abnormal but [Z] is not yet abnormal. Which upstream nodes are plausible root causes consistent with that pattern (e.g., buffering, redundancy, delay)?\par
\noindent \textbf{Correct options:} ancestors of M that can explain Z not yet impacted via redundancy, lag, or bypasses described in text.\par
\noindent \textbf{Distractors:} downstream of M; unrelated parallel modules; nodes that would necessarily affect Z immediately (contradicting the observation).

\medskip
\noindent \textbf{FL-6 Branch Isolation: Who Is Unaffected (Single-select)}\par
\noindent \textbf{Core logic:} Given flow from [S], if [M] fails, which node will not be affected because it is in an independent parallel branch (no propagation path)?\par
\noindent \textbf{Correct option:} a node topologically disconnected from M's influence (or only connected via inactive conditional edge).\par
\noindent \textbf{Distractors:} descendants of M; nodes after a shared bottleneck; nodes in the same feedback subgraph.

\medskip
\noindent \textbf{FL-7 Failure Amplification in a Feedback Loop (Single-select)}\par
\noindent \textbf{Core logic:} The text describes a feedback loop. If anomalies amplify or fail to converge, which component is the most likely amplification point (gain point in positive feedback or failed inhibitor)?\par
\noindent \textbf{Correct option:} a node or edge in the loop associated with strengthening or positive feedback or a missing inhibitor.\par
\noindent \textbf{Distractors:} loop external nodes; pure output or visualization nodes; downstream nodes not feeding back.

\medskip
\noindent \textbf{FL-8 Shared Upstream Cause for Two Abnormal Outputs (Multi-select)}\par
\noindent \textbf{Core logic:} If both [Z1] and [Z2] are abnormal, which nodes are common upstream ancestors that could explain a single shared cause?\par
\noindent \textbf{Correct options:} nodes in the intersection of ancestors of Z1 and Z2.\par
\noindent \textbf{Distractors:} nodes unique to only one output's ancestry; nodes that are purely downstream of either output; parallel branch nodes.
\end{tcolorbox}

% ------------------------------------------------------------
% Functional Mapping (FM)
% ------------------------------------------------------------
\begin{tcolorbox}[promptbox={Functional Mapping Template}]
\noindent This block is provided only when \texttt{target\_class = "Functional Mapping"}.

\medskip
\noindent Allowed template ids:
[FM-1,FM-2,FM-3,FM-4,FM-5,FM-6,FM-7,FM-8]

\medskip
\noindent Templates (choose exactly one \texttt{template\_id}):

\medskip
\noindent \textbf{FM-1 Router or Selector Identification (Single-select)}\par
\noindent \textbf{Core logic:} Which component best matches a routing or selection role (chooses among alternatives or sends inputs to different branches, often condition or gating driven)?\par
\noindent \textbf{Correct option:} node with branching out degree and conditional or gating language.\par
\noindent \textbf{Distractors:} 1 in 1 out transformation nodes; multi in 1 out aggregators; terminal outputs.

\medskip
\noindent \textbf{FM-2 Aggregator or Fusion Node (Single-select)}\par
\noindent \textbf{Core logic:} Which node best matches an aggregation or fusion role (merges multiple upstream sources into a unified representation or decision)?\par
\noindent \textbf{Correct option:} multi in 1 out join node with combine, integrate, or aggregate semantics.\par
\noindent \textbf{Distractors:} routers (1 in many out); monitoring only nodes; leaf outputs.

\medskip
\noindent \textbf{FM-3 Buffer or Storage or Delay Role (Single-select)}\par
\noindent \textbf{Core logic:} Which component most plausibly acts as buffer, storage, or state (absorbs fluctuations, supports delayed consistency, caches history)?\par
\noindent \textbf{Correct option:} node described with cache, memory, state, or history and positioned on a main path.\par
\noindent \textbf{Distractors:} pure compute or transform modules; terminal outputs; controllers with no state persistence.

\medskip
\noindent \textbf{FM-4 Controller or Constraint or Tuner Nodes (Multi-select)}\par
\noindent \textbf{Core logic:} Which components are controllers, constraints, or tuners (modify other modules' behavior rather than directly producing Z)?\par
\noindent \textbf{Correct options:} nodes with modulatory edges to multiple components (set thresholds, gate, regularize, optimize, constrain).\par
\noindent \textbf{Distractors:} passive processing nodes; purely downstream effect nodes; metrics that only observe.

\medskip
\noindent \textbf{FM-5 Mediator vs Direct Cause (Single-select)}\par
\noindent \textbf{Core logic:} Text says [A] influences [Z]. Which node is the best mediator (A primarily affects Z through it: A→M→Z)?\par
\noindent \textbf{Correct option:} node on the primary path between A and Z where removing or holding it would block A's effect.\par
\noindent \textbf{Distractors:} nodes connected to A and Z but not on a directed path; parallel nodes; downstream only nodes.

\medskip
\noindent \textbf{FM-6 Measurement or Metric Nodes (Single- or Multi-select)}\par
\noindent \textbf{Core logic:} Which elements are measurement, evaluation, or monitoring outputs rather than mechanistic drivers?\par
\noindent \textbf{Correct options:} loss, score, assay, or metric nodes, often terminal or side channel observers.\par
\noindent \textbf{Distractors:} mechanistic nodes that drive changes; controllers; main transformation stages.

\medskip
\noindent \textbf{FM-7 Domain Transformation Core Step (Single-select)}\par
\noindent \textbf{Core logic:} Treat the system as mapping [input domain] → [output domain]. Which node is the key representation or domain transformation step?\par
\noindent \textbf{Correct option:} central transform on the main information path where type or representation changes.\par
\noindent \textbf{Distractors:} purely controlling or gating nodes; terminal outputs; minor preprocessing nodes not central.

\medskip
\noindent \textbf{FM-8 Parallel Complementary Subfunctions (Multi-select)}\par
\noindent \textbf{Core logic:} The text describes parallel branches contributing to [Z]. Which branches are complementary subfunctions (different subproblems) rather than redundant backups?\par
\noindent \textbf{Correct options:} branches with distinct inputs or subtasks that later converge.\par
\noindent \textbf{Distractors:} clearly redundant branches performing the same role; branches that never converge to Z; downstream only duplications.
\end{tcolorbox}

% ------------------------------------------------------------
% Boundary Testing (BT)
% ------------------------------------------------------------
\begin{tcolorbox}[promptbox={Boundary Testing Template}]
\noindent This block is provided only when \texttt{target\_class = "Boundary Testing"}.

\medskip
\noindent Allowed template ids:
[BT-1,BT-2,BT-3,BT-4,BT-5,BT-6,BT-7,BT-8]

\medskip
\noindent Templates (choose exactly one \texttt{template\_id}):

\medskip
\noindent \textbf{BT-1 Conditional Edge Activation or Deactivation (Single-select)}\par
\noindent \textbf{Core logic:} The text indicates a dependency holds only under [Cond]. If Cond is not satisfied, which end to end capability or output is lost because no path remains?\par
\noindent \textbf{Correct option:} the unique path whose reachability depends on the conditional edge.\par
\noindent \textbf{Distractors:} unconditional paths; alternative bypass paths; relations that do not depend on Cond.

\medskip
\noindent \textbf{BT-2 Narrowest Validity or Most Assumption Dependent Module (Single-select)}\par
\noindent \textbf{Core logic:} Based on explicit assumptions or prerequisites, which component has the narrowest applicability (most assumption dependent)?\par
\noindent \textbf{Correct option:} node most frequently qualified by only when, assuming, under.\par
\noindent \textbf{Distractors:} generic baseline modules; mechanistic steps stated as broadly valid; outputs or metrics.

\medskip
\noindent \textbf{BT-3 Threshold or Saturation or Regime Shift Point (Single-select)}\par
\noindent \textbf{Core logic:} The text implies thresholds, saturation, capacity limits, or regime shifts. Which component is the most likely behavior flip point?\par
\noindent \textbf{Correct option:} node with nonlinearity, capacity, gain, or limit language on a main path.\par
\noindent \textbf{Distractors:} linear relay nodes; pure output nodes; unrelated parallel modules.

\medskip
\noindent \textbf{BT-4 Redundancy or Alternate Path Existence (Multi-select)}\par
\noindent \textbf{Core logic:} If connection [L] is cut, can [Z] still receive input from [S] via alternate routes? Select all options that describe a valid bypass path.\par
\noindent \textbf{Correct options:} real paths S→…→Z that do not traverse L and are consistent with direction or conditions.\par
\noindent \textbf{Distractors:} reversed directions; missing edges; paths requiring an unstated condition; near miss paths that end at the wrong node.

\medskip
\noindent \textbf{BT-5 Most Robust Output Under Upstream Perturbation (Single-select)}\par
\noindent \textbf{Core logic:} Under perturbation, noise, or missingness in [S], which output is most likely to remain stable given buffering, redundancy, fusion, or closed loop correction?\par
\noindent \textbf{Correct option:} output protected by explicit structural safeguards.\par
\noindent \textbf{Distractors:} outputs supported by a single brittle path; outputs relying on a narrow assumption bound module.

\medskip
\noindent \textbf{BT-6 Invariance Across Conditions (Multi-select)}\par
\noindent \textbf{Core logic:} Which dependencies are most likely invariant across changes in [Cond] (mechanistic or structural vs scenario specific)?\par
\noindent \textbf{Correct options:} edges described as structural or mechanistic, not contingent.\par
\noindent \textbf{Distractors:} statistical associations; explicitly scenario conditioned relations; edges stated to vary with context.

\medskip
\noindent \textbf{BT-7 Bypass Leakage Despite Isolation (Single-select)}\par
\noindent \textbf{Core logic:} You isolate [A] to prevent it from influencing [Z], but leakage may occur via a bypass. Which structure best explains how influence still reaches Z?\par
\noindent \textbf{Correct option:} an alternative path from A to Z (or via a shared aggregator) remains intact.\par
\noindent \textbf{Distractors:} structures where all A→Z paths are cut; purely downstream explanations; unrelated parallel branches.

\medskip
\noindent \textbf{BT-8 OOD or Domain Shift Weak Link (Single-select)}\par
\noindent \textbf{Core logic:} Under domain shift (new distribution or source), which module is most likely to fail first and degrade the full chain?\par
\noindent \textbf{Correct option:} module relying on calibration, lookup, mapping tables, or strong distributional assumptions.\par
\noindent \textbf{Distractors:} mechanistic constraints; redundancy protected stages; measurement nodes that do not drive behavior.
\end{tcolorbox}

% ------------------------------------------------------------
% Counterfactual Reasoning (CR)
% ------------------------------------------------------------
\begin{tcolorbox}[promptbox={Counterfactual Reasoning Template}]
\noindent This block is provided only when \texttt{target\_class = "Counterfactual Reasoning"}.

\medskip
\noindent Allowed template ids:
[CR-1,CR-2,CR-3,CR-4,CR-5,CR-6,CR-7,CR-8]

\medskip
\noindent Templates (choose exactly one \texttt{template\_id}):

\medskip
\noindent \textbf{CR-1 Remove an Edge: Output Change (Single-select)}\par
\noindent \textbf{Core logic:} If the relation [A→B] is removed while everything else stays fixed, which output becomes unreachable.\par
\noindent \textbf{Correct option:} B's downstream output node on the main path.\par
\noindent \textbf{Distractors:} upstream of A; unrelated parallel nodes; far downstream nodes when a nearer downstream exists.

\medskip
\noindent \textbf{CR-2 Flip Edge Polarity: Which Outputs Reverse Direction (Multi-select)}\par
\noindent \textbf{Core logic:} Flip [A's effect on B] from promoting→inhibiting (or vice versa). Which outputs most likely have their change direction reversed under the same intervention?\par
\noindent \textbf{Correct options:} outputs primarily dependent on the A→B→…→Z path without strong compensating parallel routes.\par
\noindent \textbf{Distractors:} outputs not reachable from B; outputs dominated by alternative paths; outputs whose direction is ambiguous without extra dynamics.

\medskip
\noindent \textbf{CR-3 Replace a Module with an Almost Equivalent Version (Single-select)}\par
\noindent \textbf{Core logic:} Replace [M] with a version that is functionally similar but lacks [subfunction]. Which observation most cleanly distinguishes before vs after?\par
\noindent \textbf{Correct option:} outcome that depends specifically on that subfunction along a path to Z.\par
\noindent \textbf{Distractors:} outcomes unrelated to the missing subfunction; metrics that do not reflect the affected pathway; parallel branch outcomes.

\medskip
\noindent \textbf{CR-4 Add a Shortcut Edge: What Must or May Change (Multi-select)}\par
\noindent \textbf{Core logic:} Add a new edge [A→Z] that bypasses intermediates. Which structural conclusions are supported? (Multi-select)\par
\noindent \textbf{Correct options (typical):}
\begin{itemize}
 \item Z becomes more sensitive to A
 \item intermediates on the old path become less critical
 \item redundancy may increase if both routes remain
\end{itemize}
\noindent \textbf{Distractors:} claims that require dynamics (e.g., Z is necessarily more stable), which cannot be guaranteed purely from topology.

\medskip
\noindent \textbf{CR-5 Disable a Feedback Loop: Convergence Implications (Single-select)}\par
\noindent \textbf{Core logic:} Disable the feedback loop [X→…→X]. Which consequence best matches structure given the loop's sign (positive vs negative) inferred from text?\par
\noindent \textbf{Correct option:} removal of self correction if negative feedback; removal of amplification or oscillation driver if positive feedback.\par
\noindent \textbf{Distractors:} outcomes unrelated to feedback; purely observational changes; effects that contradict loop sign.

\medskip
\noindent \textbf{CR-6 Flip a Condition: Which Effects Disappear (Multi-select)}\par
\noindent \textbf{Core logic:} A→Z holds only when [Cond] is true. Set Cond to false. Which influence chains should disappear?\par
\noindent \textbf{Correct options:} all paths whose reachability depends on the conditional edge(s).\par
\noindent \textbf{Distractors:} unconditional paths; paths that depend on different conditions; downstream only effects.

\medskip
\noindent \textbf{CR-7 Dual Intervention to Separate Direct vs Indirect Effects (Multi-select, Hard)}\par
\noindent \textbf{Core logic:} Intervene on [A] while holding [M] fixed (or also intervening on M). Which observations best distinguish A's direct effect on Z from its mediated effect via M?\par
\noindent \textbf{Correct options:} statements reflecting block the mediator, test whether A still changes Z, or control M to isolate the direct path.\par
\noindent \textbf{Distractors:} correlation only statements; observations that do not mention blocking or controlling a path; claims requiring parameter values not present.

\medskip
\noindent \textbf{CR-8 Swap Upstream Source: Which Downstream Stays Similar (Single-select)}\par
\noindent \textbf{Core logic:} Swap upstream input from [S1] to [S2], and the text indicates they merge at some node [J]. Which output should change least (or remain invariant) structurally?\par
\noindent \textbf{Correct option:} outputs located downstream of the merge J and dependent only on shared subgraph.\par
\noindent \textbf{Distractors:} outputs upstream of J; outputs depending on S1 specific branch; outputs driven by side branches that do not share.
\end{tcolorbox}

% \section{Quality Check}
% \input{quality}

\subsection{Human Evaluation Checklist}
\label{app:B4}
The construction of the T2S-Bench dataset followed a rigorous three-stage human quality assurance (QA) pipeline to ensure the accuracy, consistency, and usability of each sample. In every stage, we employed a structured checklist-based review process, where each sample was meticulously evaluated across a range of quality dimensions—such as correctness, relevance to the source text, structural coherence, and task difficulty. Only those samples that successfully passed all items in the checklist at every round were retained in the final dataset. This multi-round validation protocol significantly enhances the reliability and interpretability of T2S-Bench, making it a robust benchmark for evaluating text-to-structure capabilities. In this section, we present the detailed QA checklists used across the three rounds.

\textbf{First-Round Human Checklist. }
The goal of the first round of human quality assurance was to filter out structurally noisy or low-quality samples that could compromise the integrity of the dataset. This initial screening focused on identifying and removing examples with incomplete or misleading structures, ensuring that only samples with a clear and coherent structural representation were retained. Specifically, annotators were instructed to evaluate the following aspects:
\begin{enumerate}

    \item Are both nodes and links present?

\item Does every node have a clear corresponding entity/semantic meaning?

\item Does every link have a clear source and target?

\item Does the diagram contain noise?

\item Is the number of nodes appropriate (greater than 5 and fewer than 20)?

\item Is the number of edges appropriate (greater than 5 and fewer than 40)?

\item Are there multiple nodes with duplicate names?

\item Is each node expressed as a phrase or a set of short phrases (rather than long sentences)?

\item Is the sample a single, self-contained structure graph, without nested containment or multiple independent sub-structures?
\end{enumerate}

\textbf{Second-Round Human Checklist.}The second round of quality assurance focused on validating the correctness and solvability of the questions designed for model evaluation. In this phase, annotators were required to carefully read the input text, interpret the associated structure, and independently attempt to solve each task as if they were language models. This ensured that the tasks were not only well-formed but also answerable based on the provided information. The checklist for this round included the following criteria:
\begin{enumerate}

    \item Correctness Check: Is the answer correct?
    \item Text Dependency Check: Can the answer be inferred from the text?
    \item Difficulty Check: Is the question overly easy—i.e., can it be answered by directly locating a span in the text without meaningful reasoning?
    \item Typo Check: Is the question semantically coherent and free of typos?
    \item Format Check: Are symbols and equations in the question properly formatted in Markdown?
\end{enumerate}

\textbf{Third-Round Human Checklist.} The third round of quality assurance was designed to validate the alignment between the input text and its corresponding key structure. In this phase, annotators were first asked to independently derive the core structure—i.e., the key nodes and their relationships—based solely on the input text, without referring to the provided structure. They then compared their inferred structure with the given one to assess consistency and faithfulness. Samples with significant structural deviation, overly complex representations, or inconsistent mappings were filtered out. The checklist used in this stage included the following criteria:
\begin{enumerate}
    \item Structure-Text Consistency: Does the provided structure accurately reflect the key information, logic, and flow of the original text?
    \item Reproducibility: Can a well-informed annotator infer a similar structure from the text alone, without prior exposure to the given structure?
    \item Over-Structuring Avoidance: Is the structure appropriately scoped, avoiding unnecessary complexity or over-fragmentation of ideas?
    \item Key Node Coverage: Are all critical concepts, entities, or steps in the text represented as distinct nodes in the structure?
    \item 
Relation Accuracy: Are the links between nodes semantically correct, reflecting valid dependencies or logical flows?
\end{enumerate}

\section{Model Evaluation Details}
\label{app:C}
\lstdefinestyle{mono}{
  basicstyle=\ttfamily\footnotesize,
  breaklines=true,
  breakatwhitespace=true,
  columns=fullflexible,
  keepspaces=true,
  showstringspaces=false
}

This section summarizes the evaluation procedure used for both multiple choice question answering and structure extraction tasks in the T2S benchmark. We report the prompt template, describe data layout assumptions, and detail the execution paths for API based models and open source models.

\subsection{Experiment setup}
\label{app:C1}

\paragraph{Evaluation scripts.}
We provide two evaluation tracks:
\begin{itemize}
  \item Multiple choice question answering
  \item Structure extraction
\end{itemize}

\paragraph{Dataset layout.}
Each benchmark sample is stored as a directory. The evaluator expects:
\begin{itemize}
  \item \texttt{information.json} inside each sample folder, containing the question and the ground truth answer.
  \item One extracted markdown file, which provides the cleaned text paragraph used as the model input.
\end{itemize}

\paragraph{Prompt assembly.}
For multiple choice evaluation, the final prompt is assembled by concatenating:
\begin{itemize}
  \item a \texttt{TEXT PARAGRAPH} block from the extracted markdown
  \item the \texttt{QUESTION}
  \item the \texttt{OPTIONS} rendered as \texttt{A. ...}, \texttt{B. ...} etc
  \item a strict output format instruction that forces an answer string
\end{itemize}

For structure extraction, we use a two stage protocol:
\begin{itemize}
  \item Stage 1 node labeling: input is text plus graph structure (links without labels) and node ids, output is JSON node labels
  \item Stage 2 link extraction: input is text plus node list, output is JSON links
\end{itemize}

\paragraph{Decoding and determinism.}
All evaluation paths use deterministic decoding with temperature set to zero. The maximum generation length is set conservatively for each task (for example, longer for structure JSON, shorter for multiple choice answers).

\subsection{API evaluation}
\label{app:C2}

\paragraph{API interface.}
API models are accessed through an OpenAI compatible chat completion interface. The evaluator accepts an API key and a base URL so that the same code path can evaluate models hosted behind different compatible gateways.

\paragraph{Rate limit handling and timeouts.}
For question answering, the API wrapper retries upon rate limit signals using exponential backoff. For structure extraction, the API wrapper additionally supports streaming responses and configurable read timeouts to prevent long running calls from hanging the evaluation.

\paragraph{Prompt specification for multiple choice QA.}
We enforce a strict output schema to make answer parsing robust. The following prompt components are used.

\begin{tcolorbox}[
  enhanced,
  breakable,
  colback=black!2,
  colframe=black!25,
  boxrule=0.6pt,
  arc=2mm,
  left=2.5mm,
  right=2.5mm,
  top=2mm,
  bottom=2mm,
  title={QA evaluation prompt contract},
  fonttitle=\bfseries
]
\begin{lstlisting}[style=mono]
SYSTEM PROMPT
You are a helpful assistant. Output only the final answer in the required format.

FORMAT INSTRUCTION
Provide your final answer in the following format:
For single select: [Answer] A
For multi select: [Answer] A,C,D (comma separated option letters, no spaces)
Output only the final answer with no extra explanation
\end{lstlisting}
\end{tcolorbox}

\paragraph{Special casing for certain APIs.}
Some endpoints do not accept system messages reliably. In that case, the evaluator prepends the system instruction to the user content to preserve the same contract.

\paragraph{Structure extraction prompts.}
Structure evaluation uses two separate system prompts and format blocks, one for node labeling and one for link extraction.

\begin{tcolorbox}[
  enhanced,
  breakable,
  colback=black!2,
  colframe=black!25,
  boxrule=0.6pt,
  arc=2mm,
  left=2.5mm,
  right=2.5mm,
  top=2mm,
  bottom=2mm,
  title={Structure evaluation prompt contract},
  fonttitle=\bfseries
]
\begin{lstlisting}[style=mono]
STAGE 1  NODE LABELING

SYSTEM PROMPT
You are a helpful assistant. You will be given a text and a graph structure (links between nodes).
Your task is to read the text and provide a concise label (single word or short phrase) for each node
that best represents its meaning based on the text content. Output only valid JSON with no extra text.
Put your JSON after [Nodes].

FORMAT INSTRUCTION
Output format: Put the JSON after [Nodes]
[Nodes]
{
  "nodes": [
    { "id": "n1", "label": "Your Label for n1" },
    { "id": "n2", "label": "Your Label for n2" }
  ]
}

STAGE 2  LINK EXTRACTION

SYSTEM PROMPT
You are a helpful assistant. Given the text and a list of nodes, identify the relationships
(links or edges) between these nodes based on the text. Each link should connect two nodes and
optionally have a label describing the relationship. Output only valid JSON with no extra text.
Put your JSON after [Links].

FORMAT INSTRUCTION
Output format: Put the JSON after [Links]
[Links]
{
  "links": [
    { "source": "n1", "target": "n2", "label": "Link Label or empty" }
  ]
}
\end{lstlisting}
\end{tcolorbox}

\paragraph{Provider specific behavior for structure evaluation.}
The structure evaluator includes optional streaming and extra request fields for provider specific features such as thinking mode. It also includes a fallback path that extracts reasoning content when the standard content field is empty for certain endpoints.

\subsection{Open source model evaluation}
\label{app:C3}

\paragraph{Local HuggingFace models.}
For local evaluation, open source models are loaded via \texttt{AutoTokenizer} and \texttt{AutoModelForCausalLM}. We run deterministic generation and parse the answer using the same format contract as the API track.

\paragraph{OpenAI compatible providers for open source endpoints.}
In addition to local inference, we evaluate open source models served behind OpenAI compatible endpoints through a unified wrapper. This path supports:
\begin{itemize}
  \item streaming for endpoints that require incremental token delivery
  \item retry and backoff for transient failures
  \item resume support through per model per sample partial caches so interrupted runs can continue without repeating completed samples
\end{itemize}

\paragraph{Consistency across tracks.}
Across API and open source evaluations, we keep the same input text, the same question formatting, the same strict answer schema, and the same EM and F1 computation. This ensures that differences in scores primarily reflect model capability rather than evaluator artifacts.

\section{Sample Examples}
\label{app:D}
\subsection{Structure diagram examples}
\label{app:D1}
We visualize the diversity of structural topologies present in the T2S-Bench dataset. We provide representative diagrams from each of the six core scientific domains: Computer Science, Economics, Environmental Sciences, Life Sciences, Physical Sciences, and Social Sciences. These examples illustrate the wide range of visual structures—including sequential flows, feedback loops, and parallel hierarchies—that the model is required to extract and reason over.

% Preamble:
% \usepackage{graphicx}
% \usepackage{subcaption}
% \usepackage{xcolor}

% Fixed image cell height (tune once)
\newlength{\GridImgH}
\setlength{\GridImgH}{4.6cm}

\begin{figure*}[t]
  \centering
  \captionsetup{font=small}
  \captionsetup[subfigure]{font=footnotesize}

  % --- Row 1 ---
  \begin{subfigure}[t]{0.3\textwidth}
    \centering
    \begin{minipage}[c][\GridImgH][c]{\linewidth}
      \centering
      \includegraphics[width=\linewidth,height=\GridImgH,keepaspectratio]{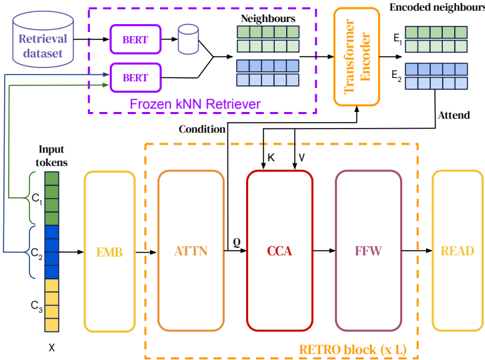}
    \end{minipage}
    \caption{Computer Science Example}
    \label{fig:grid_01}
  \end{subfigure}\hfill
  \begin{subfigure}[t]{0.3\textwidth}
    \centering
    \begin{minipage}[c][\GridImgH][c]{\linewidth}
      \centering
      \includegraphics[width=\linewidth,height=\GridImgH,keepaspectratio]{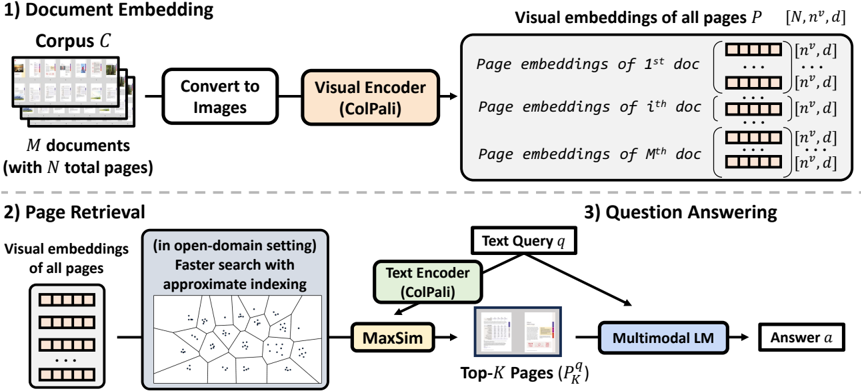}
    \end{minipage}
    \caption{Computer Science Example}
    \label{fig:grid_02}
  \end{subfigure}\hfill
  \begin{subfigure}[t]{0.3\textwidth}
    \centering
    \begin{minipage}[c][\GridImgH][c]{\linewidth}
      \centering
      \includegraphics[width=\linewidth,height=\GridImgH,keepaspectratio]{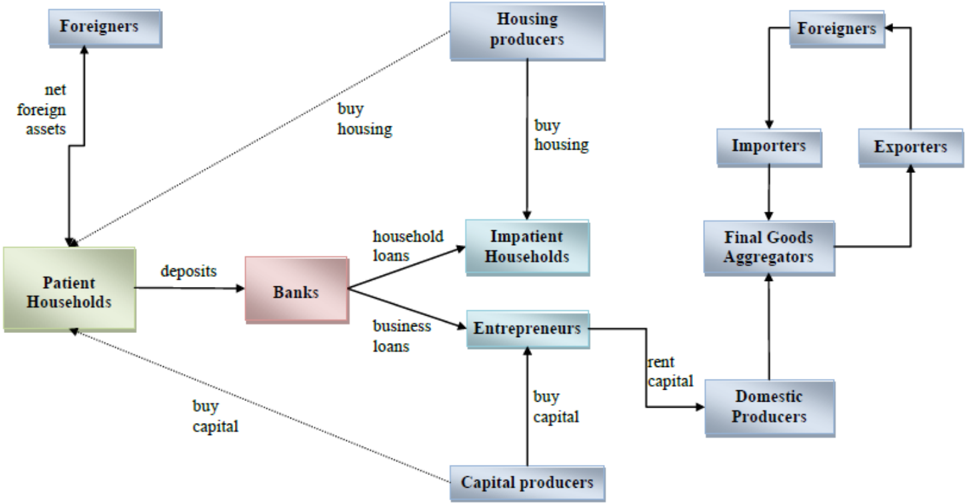}
    \end{minipage}
    \caption{Economic and Management Example}
    \label{fig:grid_03}
  \end{subfigure}

  \vspace{10pt}

  % --- Row 2 ---
  \begin{subfigure}[t]{0.3\textwidth}
    \centering
    \begin{minipage}[c][\GridImgH][c]{\linewidth}
      \centering
      \includegraphics[width=\linewidth,height=\GridImgH,keepaspectratio]{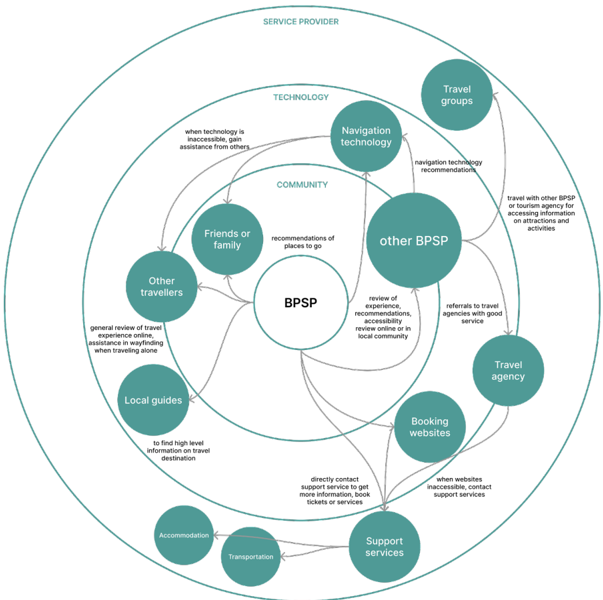}
    \end{minipage}
    \caption{Economic and Management Example}
    \label{fig:grid_04}
  \end{subfigure}\hfill
  \begin{subfigure}[t]{0.3\textwidth}
    \centering
    \begin{minipage}[c][\GridImgH][c]{\linewidth}
      \centering
      \includegraphics[width=\linewidth,height=\GridImgH,keepaspectratio]{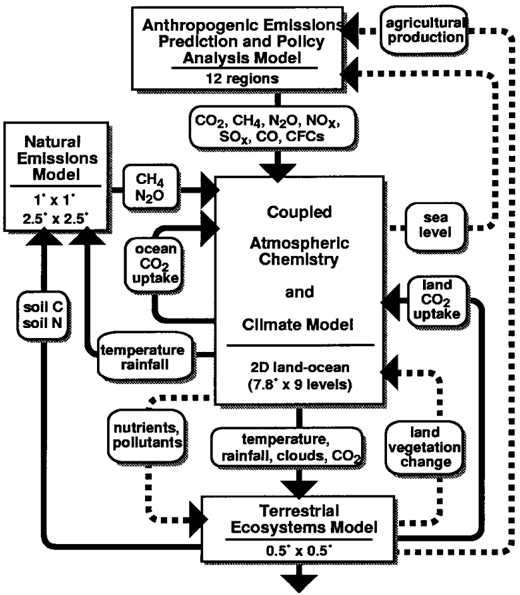}
    \end{minipage}
    \caption{Environmental Sciences Example}
    \label{fig:grid_05}
  \end{subfigure}\hfill
  \begin{subfigure}[t]{0.3\textwidth}
    \centering
    \begin{minipage}[c][\GridImgH][c]{\linewidth}
      \centering
      \includegraphics[width=\linewidth,height=\GridImgH,keepaspectratio]{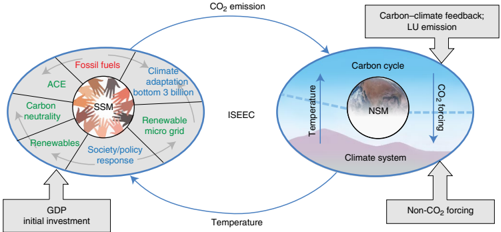}
    \end{minipage}
    \caption{Environmental Sciences Example}
    \label{fig:grid_06}
  \end{subfigure}

  \vspace{10pt}

  % --- Row 3 ---
  \begin{subfigure}[t]{0.3\textwidth}
    \centering
    \begin{minipage}[c][\GridImgH][c]{\linewidth}
      \centering
      \includegraphics[width=\linewidth,height=\GridImgH,keepaspectratio]{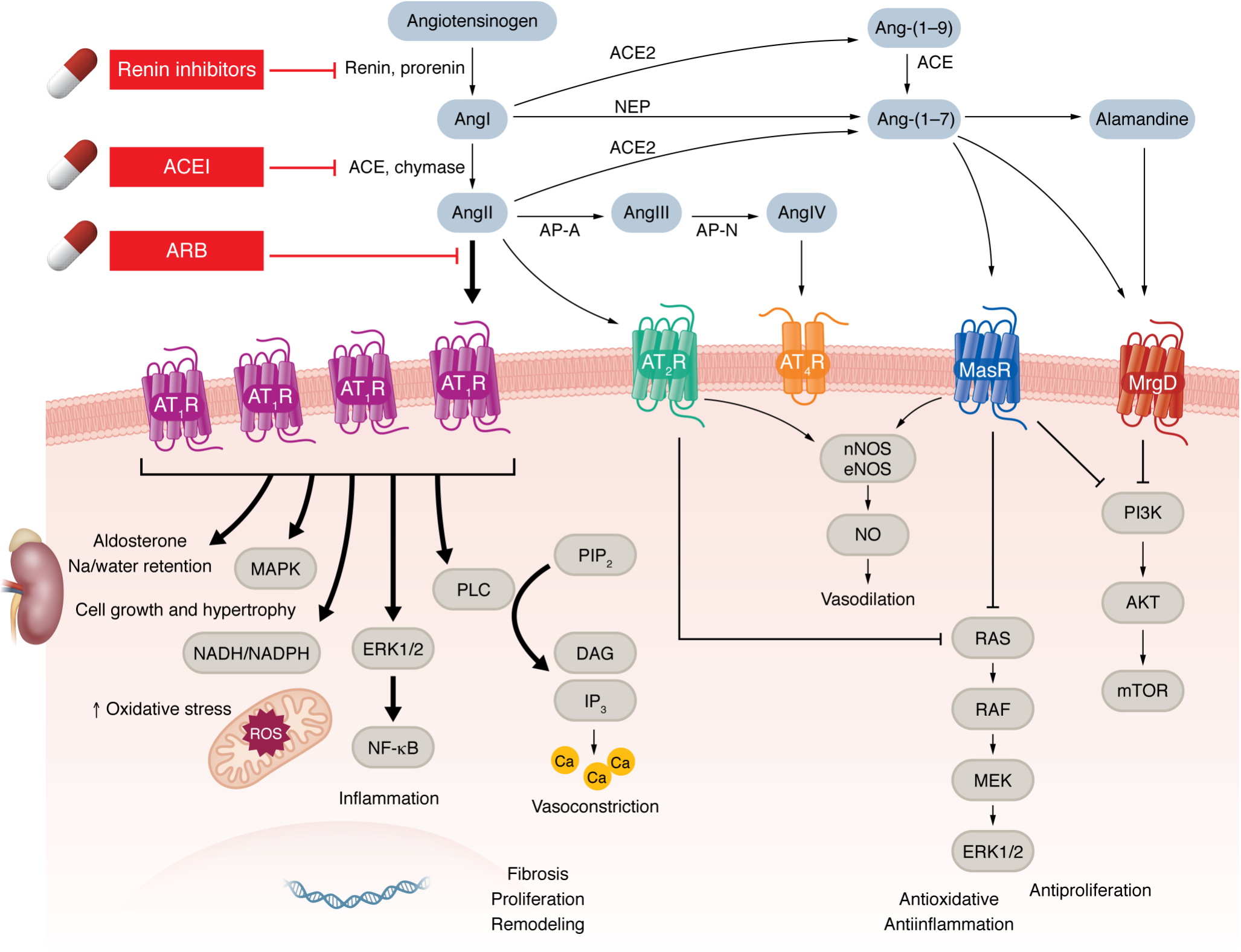}
    \end{minipage}
    \caption{Life Sciences Example}
    \label{fig:grid_07}
  \end{subfigure}\hfill
  \begin{subfigure}[t]{0.3\textwidth}
    \centering
    \begin{minipage}[c][\GridImgH][c]{\linewidth}
      \centering
      \includegraphics[width=\linewidth,height=\GridImgH,keepaspectratio]{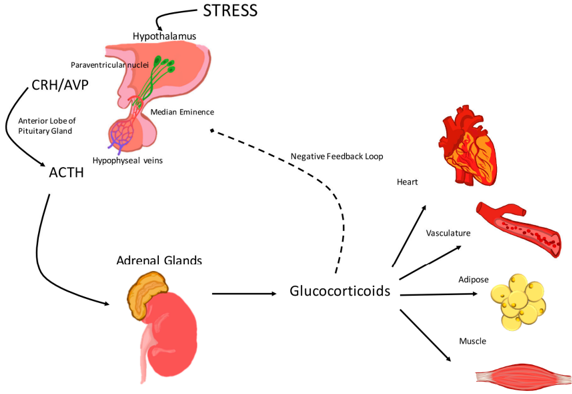}
    \end{minipage}
    \caption{Life Sciences Example}
    \label{fig:grid_08}
  \end{subfigure}\hfill
  \begin{subfigure}[t]{0.3\textwidth}
    \centering
    \begin{minipage}[c][\GridImgH][c]{\linewidth}
      \centering
      \includegraphics[width=\linewidth,height=\GridImgH,keepaspectratio]{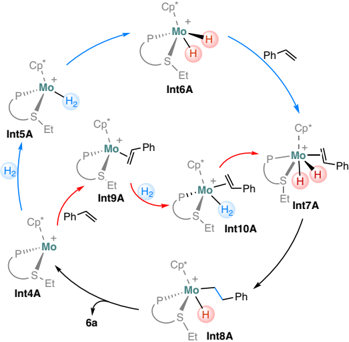}
    \end{minipage}
    \caption{Physical Sciences Example}
    \label{fig:grid_09}
  \end{subfigure}

  \vspace{10pt}

  % --- Row 4 ---
  \begin{subfigure}[t]{0.3\textwidth}
    \centering
    \begin{minipage}[c][\GridImgH][c]{\linewidth}
      \centering
      \includegraphics[width=\linewidth,height=\GridImgH,keepaspectratio]{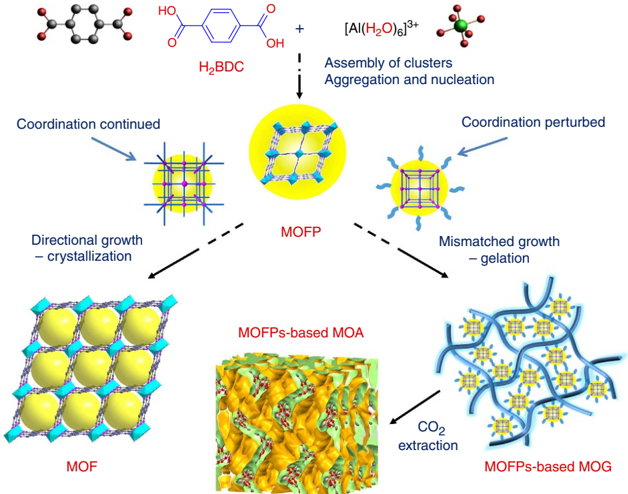}
    \end{minipage}
    \caption{Physical Sciences Example}
    \label{fig:grid_10}
  \end{subfigure}\hfill
  \begin{subfigure}[t]{0.3\textwidth}
    \centering
    \begin{minipage}[c][\GridImgH][c]{\linewidth}
      \centering
      \includegraphics[width=\linewidth,height=\GridImgH,keepaspectratio]{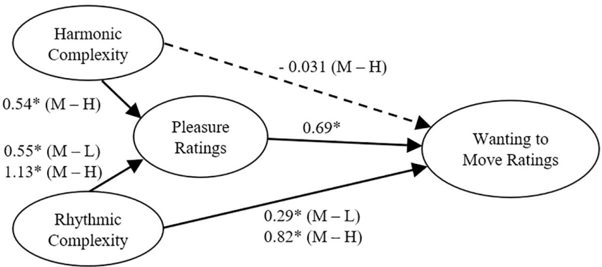}
    \end{minipage}
    \caption{Social Sciences Example}
    \label{fig:grid_11}
  \end{subfigure}\hfill
  \begin{subfigure}[t]{0.3\textwidth}
    \centering
    \begin{minipage}[c][\GridImgH][c]{\linewidth}
      \centering
      \includegraphics[width=\linewidth,height=\GridImgH,keepaspectratio]{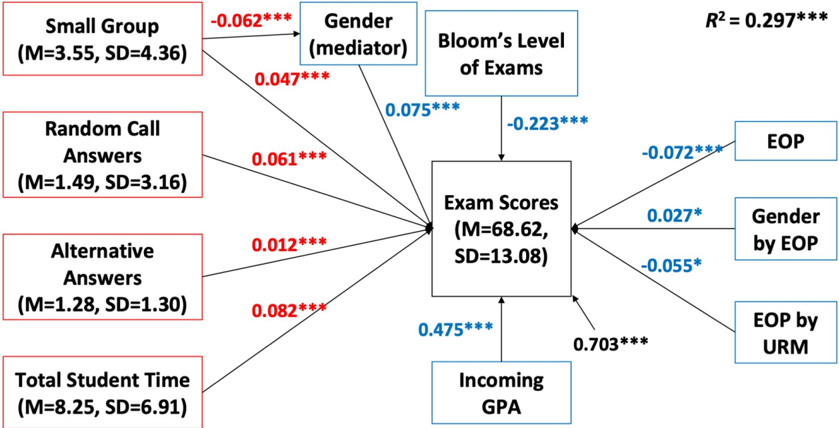}
    \end{minipage}
    \caption{Social Sciences Example}
    \label{fig:grid_12}
  \end{subfigure}

  \caption{Structure diagram examples across main domains}
  \label{fig:grid_3x4}
\end{figure*}

\subsection{Full sample examples}
\label{app:D2}
Here we present a complete instance of the dataset entries to demonstrate the full data format. For each sample, we show the input text paragraph extracted from the scientific paper, the ground-truth topology (node-link graph) derived from that text, and the corresponding multi-hop reasoning question (including the reasoning analysis plan and distractors). 
\lstdefinestyle{mono}{
  basicstyle=\ttfamily\footnotesize,
  breaklines=true,
  breakatwhitespace=true,
  columns=fullflexible,
  keepspaces=true,
  showstringspaces=false,
  literate=
    {’}{{'}}1 {“}{{``}}1 {”}{{''}}1 {–}{{\textendash}}1 {—}{{\textemdash}}1 {…}{{...}}3
}

% In document
\clearpage
\onecolumn

\vspace{0.5\baselineskip}
\begin{tcolorbox}[
  enhanced,
  breakable,
  colback=black!2,
  colframe=black!25,
  boxrule=0.6pt,
  arc=2mm,
  left=2.5mm,
  right=2.5mm,
  top=2mm,
  bottom=2mm,
  title={CS Algorithm Model Architectural Topology \texttt{information.json}},
  fonttitle=\bfseries
]
\centering
\includegraphics[width=0.7\linewidth]{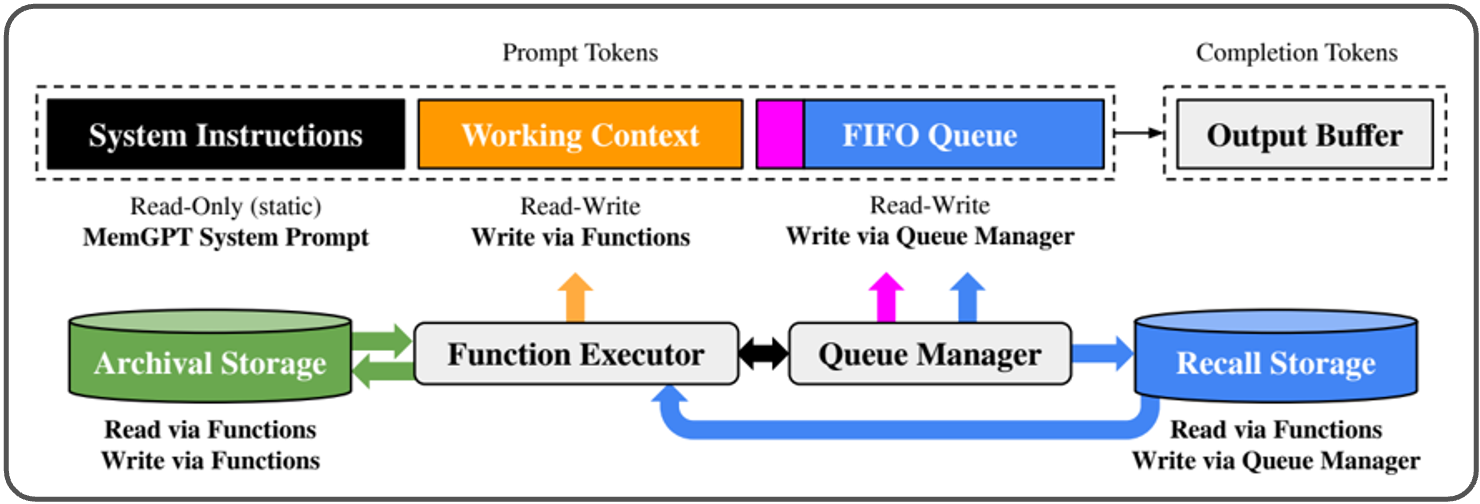}

\medskip

\begin{lstlisting}[style=mono]
% ===== Extracted context (excerpt) =====
### 2. MemGPT (MemoryGPT)
MemGPT's OS-inspired multi-level memory architecture delineates between two primary memory types: **main context** (analogous to main memory/physical memory/RAM) and **external context** (analogous to disk memory/disk storage).
(...)
If this flag is present, MemGPT will add the function output to main context and (as opposed to pausing processor execution). If this flag is not present (a *yield*), MemGPT will not run the LLM processor until the next external event trigger (e.g. a user message or scheduled interrupt).

% ===== information.json =====
{
  "paper_metadata": {
    "title": "MemGPT: Towards LLMs as Operating Systems",
    "paper_link": "https://arxiv.org/pdf/2310.08560.pdf",
    "topology_type": "Model Architectural Topology"
  },
  "figure": {
    "figure_id": "Figure 3",
    "figure_caption_snippet": "Figure 3: In MemGPT, a fixed-context LLM processor is augmented with a hierarchical memory system and functions that let it manage its own memory. (...) MemGPT uses functions to move data between main context and external context (the archival and recall storage databases)."
  },
  "question": {
    "question_class": "Functional Mapping",
    "template_id": "FM-5",
    "type": "single",
    "analysis_plan": {
      "target_query": "Identify the mediator that channels data from Archival Storage into Working Context.",
      "key_nodes": [
        "Archival Storage",
        "Function Executor",
        "Working Context"
      ],
      "key_edges_or_relations": [
        "Archival Storage -> Function Executor (read/write via functions)",
        "Function Executor -> Working Context (writes via functions)"
      ],
      "two_hop_witness": "Archival Storage is accessed through MemGPT function calls handled by the Function Executor, which in turn is the only entity allowed to write those results into the Working Context (Archival Storage -> Function Executor -> Working Context).",
      "why_not_one_hop": "There is no direct edge from Archival Storage to Working Context; one must recognize the intermediate role of the Function Executor."
    },
    "question": "When information stored in Archival Storage needs to be inserted into the main-context Working Context, which component acts as the essential intermediary in that transfer?",
    "options": [
      {
        "id": "A",
        "text": "Queue Manager"
      },
      {
        "id": "B",
        "text": "Output Buffer"
      },
      {
        "id": "C",
        "text": "Function Executor"
      },
      {
        "id": "D",
        "text": "Recall Storage"
      }
    ],
    "answer": [
      "C"
    ],
    "reason": "Data in Archival Storage can only be accessed through MemGPT function calls, and Working Context is writable exclusively via those same functions. The Function Executor is responsible for executing these calls, making it the necessary bridge between Archival Storage and Working Context."
  },
}
\end{lstlisting}
\end{tcolorbox}
\vspace{0.5\baselineskip}

% \clearpage

\section{Additional Results and Analysis}
\label{app:E}
In this section, we present the experimental setup, report comprehensive results across a diverse set of models and tasks, and discuss the novel insights derived from these findings.

\subsection{Experimental Setting}
\label{app:E1}

\paragraph{Dataset.} The T2S‑Bench benchmark is built through a three‑stage pipeline. First, we collect text–structure pairs by extracting diagrammatic structures from scientific papers, covering six major scientific domains—computer science, economics, environmental science, life science, physics and social science—across 33 sub‑domains. Second, we construct a multi‑hop reasoning (MR) multiple‑choice dataset where each question requires reasoning over a graph with multiple nodes. These questions fall into four graph‑reasoning categories and 32 templates. Third, we build an end‑to‑end (E2E) structure extraction task consisting of 88 human‑verified samples. In this task the model must output a node–link graph given only the raw text; to fairly handle the one‑to‑many nature of structuring, we report separate NodeF1 and LinkF1 scores.

\paragraph{Splits and metrics.} We use a training set of 1.2k instruction–answer pairs (T2S‑Bench‑Instruct‑1.2K) to fine‑tune structure‑aware models, a multi‑hop reasoning test set of 500 multiple‑choice questions for evaluation, and an E2E test set of 88 samples for structure extraction. We measure Exact‑Match (EM) and F1 on multiple‑choice tasks, and NodeF1/LinkF1 on E2E tasks. For downstream generalisation, we also evaluate models on external long‑context benchmarks (LongBench, Scrolls). When evaluating reasoning strategies we consider four inference schemes: vanilla prompting, chain‑of‑thought (CoT), structure‑of‑thought (SoT) and models fine‑tuned on T2S‑Bench (T2S‑Train).

\paragraph{Models.} Our study spans a wide spectrum of proprietary and open‑source large language models. In total, we assess more than forty mainstream language models drawn from a broad cross‑section of model families. These include proprietary giants such as Gemini‑2.5‑Pro, Gemini‑2.5‑Flash, Gemini‑2.0 Flash and Flash‑Lite~\cite{team2023gemini,comanici2025gemini}, GPT‑5, GPT‑4o, and GPT-3.5 variants~\cite{achiam2023gpt, singh2025openai, brown2020language}, and multiple Claude versions~\cite{TheC3}, open‑source instruction‑tuned models such as DeepSeek V3/R1/Chat~\cite{guo2025deepseek, liu2024deepseek}, Qwen3 series~\cite{yang2025qwen3}, Kimi‑K2~\cite{team2025kimi}, smaller baselines such as GLM‑4.5/4.6, MiniMax‑M2 and MiniMax‑Text‑01~\cite{zeng2025glm, li2025minimax}, and mid‑sized alternatives such as Mistral and LLaMA~\cite{jiang2023mistral7b, touvron2023llama, grattafiori2024llama} in a range of sizes and instruction variants. Each model receives identical prompts, random seeds and inference scripts to ensure a fair comparison across architectures, scales, and training paradigms.

\definecolor{EMcol}{gray}{0.92}
\definecolor{F1col}{gray}{1.00}

\newcommand{\ModelWithLogo}[2]{%
  \makebox[\linewidth][l]{#1\hfill\raisebox{-0.15\height}{\includegraphics[height=1.15em]{fig/logo/#2}}}%
}
\begin{table*}[t]
\vspace{-8pt}
\caption{Performance on Multi choice QA. \colorbox{EMcol}{Shaded} columns report EM, unshaded columns report F1.}
\centering
\small
\setlength{\tabcolsep}{3.2pt}
\renewcommand{\arraystretch}{1.08}
\resizebox{\textwidth}{!}{%
\begin{tabular}{@{}>{\raggedright\arraybackslash}p{5.4cm}
>{\columncolor{EMcol}}c >{\columncolor{F1col}}c
>{\columncolor{EMcol}}c >{\columncolor{F1col}}c
>{\columncolor{EMcol}}c >{\columncolor{F1col}}c
>{\columncolor{EMcol}}c >{\columncolor{F1col}}c
>{\columncolor{EMcol}}c >{\columncolor{F1col}}c
>{\columncolor{EMcol}}c >{\columncolor{F1col}}c
>{\columncolor{EMcol}}c >{\columncolor{F1col}}c
@{}}
\toprule[1.2pt]
& \multicolumn{2}{c}{\textbf{Overall}} & \multicolumn{2}{c}{\textbf{CS}} & \multicolumn{2}{c}{\textbf{Eco}} & \multicolumn{2}{c}{\textbf{Environ}} & \multicolumn{2}{c}{\textbf{Life}} & \multicolumn{2}{c}{\textbf{Phy}} & \multicolumn{2}{c}{\textbf{Social}} \\
\midrule

Gemini-2.5-Pro & \best{81.40} & \best{91.56} & \best{83.48} & \best{93.12} & \best{78.95} & \best{91.66} & \best{75.76} & \best{91.67} & \second{80.65} & \third{88.57} & \best{88.52} & \third{92.33} & \best{80.90} & \best{91.96} \\
Gemini-2.5-Flash & 72.20 & 83.67 & \third{73.91} & 81.18 & 65.79 & 82.04 & \third{71.21} & 85.53 & 65.59 & 79.09 & \second{81.97} & 88.81 & \third{76.40} & 88.14 \\
Gemini-2.0-flash & 61.20 & 79.63 & 61.74 & 77.07 & 61.84 & 80.95 & 60.61 & 80.81 & 59.14 & 79.32 & 67.21 & 76.12 & 58.43 & 83.66 \\
\ModelWithLogo{Gemini-2.0-flash-lite}{gemini.png} & 53.40 & 72.41 & 53.04 & 72.01 & 50.00 & 69.04 & 56.06 & 74.44 & 53.76 & 73.50 & 60.66 & 68.47 & 49.44 & 75.88 \\

\addlinespace[1.2pt]
\midrule
GPT-5.2 & 71.80 & 84.32 & \third{73.91} & 82.55 & 65.79 & \third{86.44} & 66.67 & 79.85 & 73.12 & 87.10 & 77.05 & 80.77 & 73.03 & 87.65 \\
GPT-5.1 & 67.80 & 83.46 & \third{73.91} & \third{83.83} & 65.79 & 83.65 & 68.18 & 82.22 & 62.37 & 82.32 & 73.77 & 80.42 & 62.92 & 86.99 \\
GPT-4.1-mini & 59.00 & 76.83 & 56.52 & 71.72 & 55.26 & 79.54 & 54.55 & 76.32 & 61.29 & 80.61 & 65.57 & 75.03 & 61.80 & 78.76 \\
GPT-3.5-turbo & 25.40 & 48.63 & 23.48 & 43.80 & 25.00 & 46.23 & 30.30 & 53.18 & 20.43 & 47.13 & 32.79 & 46.28 & 24.72 & 56.74 \\
GPT-4o & 61.80 & 79.24 & 63.48 & 76.67 & 68.42 & 84.38 & 57.58 & 73.84 & 61.29 & 82.59 & 63.93 & 74.37 & 56.18 & 82.03 \\
\ModelWithLogo{GPT-4o-mini}{openai.png} & 20.40 & 54.19 & 22.61 & 50.95 & 19.74 & 52.88 & 24.24 & 59.96 & 19.35 & 54.90 & 21.31 & 51.30 & 15.73 & 56.47 \\

\addlinespace[1.2pt]
\midrule 
Claude-sonnet-4-5-20250929 & \second{76.80} & \second{86.85} & \second{77.39} & \second{84.72} & \third{72.37} & 84.25 & \third{71.21} & \third{87.71} & \best{82.80} & \second{90.97} & \third{80.33} & 85.45 & 75.28 & 87.85 \\
Claude-haiku-4-5-20251001 & 67.40 & 80.87 & 66.09 & 76.42 & 67.11 & 82.72 & 63.64 & 79.13 & 65.59 & 80.80 & \third{80.33} & 85.14 & 65.17 & 83.47 \\
Claude-4-Sonnet-20250514 & \third{75.60} & \third{86.63} & 73.04 & 81.42 & \second{77.63} & \second{86.45} & \second{74.24} & \second{88.95} & \third{78.49} & \best{91.84} & \third{80.33} & 84.70 & 71.91 & 87.68 \\
\ModelWithLogo{Claude-3-haiku-20240307}{claude.png} & 44.80 & 68.77 & 46.09 & 64.70 & 47.37 & 71.42 & 40.91 & 66.70 & 44.09 & 69.74 & 55.74 & 70.33 & 37.08 & 71.21 \\
% Claude-3.7-Sonnet & & & & & & & & & & & & & & \\
\addlinespace[1.2pt]
\midrule 
DeepSeek-V3.1 & 60.40 & 78.59 & 56.52 & 72.56 & 69.74 & 86.10 & 59.09 & 78.20 & 59.14 & 80.70 & 60.66 & 72.77 & 59.55 & 82.05 \\
DeepSeek-V3.2 & 60.00 & 78.31 & 55.65 & 71.84 & 69.74 & 85.84 & 59.09 & 78.20 & 58.06 & 79.97 & 60.66 & 72.77 & 59.55 & 82.39 \\
DeepSeek-R1-0528 & 57.00 & 63.76 & 58.26 & 64.23 & 57.89 & 65.88 & 48.48 & 61.05 & 53.76 & 61.76 & 55.74 & 57.14 & 65.17 & 69.96 \\
DeepSeek-reasoner (R1) & 53.32 & 80.22 & 39.78 & 76.80 & 61.51 & 75.57 & 36.93 & 58.68 & 50.83 & 85.14 & 34.05 & \best{100.00} & \second{76.84} & \second{89.95} \\
DeepSeek-chat & 47.58 & 78.97 & 32.47 & 76.88 & 53.36 & 69.08 & 29.78 & 67.33 & 45.08 & 82.68 & 23.79 & \second{93.33} & 76.37 & \third{88.83} \\
\ModelWithLogo{DeepSeek-V3-0324}{deepseek.png} & 60.60 & 78.75 & 57.39 & 72.83 & 69.74 & 86.10 & 59.09 & 77.95 & 59.14 & 80.70 & 62.30 & 74.41 & 58.43 & 81.67\\

\addlinespace[1.2pt]
\midrule 
Qwen3-235B-A22B-Thinking-2507 & 60.80 & 79.89 & 66.09 & 79.50 & 59.21 & 80.87 & 62.12 & 83.35 & 61.29 & 79.84 & 60.66 & 74.50 & 53.93 & 80.75 \\
Qwen3-235B-A22B-Instruct-2507 & 62.80 & 80.25 & 67.83 & 79.93 & 56.58 & 79.93 & 65.15 & 81.75 & 60.22 & 80.19 & 70.49 & 78.52 & 57.30 & 81.06 \\
Qwen3-Next-80B-A3B-Thinking & 24.40 & 36.21 & 26.09 & 32.54 & 26.32 & 39.93 & 18.18 & 30.09 & 25.81 & 40.43 & 22.95 & 32.61 & 24.72 & 40.36 \\
Qwen3-Next-80B-A3B-Instruct & 46.60 & 57.00 & 44.35 & 49.16 & 42.11 & 54.56 & 53.03 & 61.85 & 44.09 & 59.25 & 54.10 & 61.31 & 46.07 & 60.30 \\
Qwen3-30B-A3B-Thinking-2507 & 43.20 & 64.82 & 41.74 & 58.29 & 38.16 & 62.19 & 53.03 & 73.74 & 38.71 & 63.44 & 54.10 & 69.40 & 39.33 & 67.22 \\
Qwen3-30B-A3B-Instruct-2507 & 47.40 & 72.15 & 49.57 & 69.19 & 39.47 & 67.71 & 56.06 & 78.92 & 44.09 & 74.21 & 55.74 & 70.55 & 42.70 & 73.71 \\
% Qwen3-4B-Thinking-2507 & & & & & & & & & & & & & & \\
% Qwen3-4B-Instruct-2507 & & & & & & & & & & & & & & \\
Qwen3-32B & 69.40 & 83.41 & 69.57 & 81.11 & 71.05 & 84.99 & 60.61 & 77.32 & 68.82 & 85.20 & 70.49 & 81.86 & 74.16 & 88.73 \\
Qwen3-14B & 47.40 & 70.38 & 46.96 & 65.85 & 47.37 & 70.47 & 54.55 & 76.18 & 39.78 & 69.45 & 49.18 & 68.20 & 49.44 & 74.30 \\
\ModelWithLogo{Qwen3-8B}{qwen.png} & 47.60 & 66.47 & 49.57 & 66.07 & 52.63 & 69.68 & 42.42 & 67.53 & 38.71 & 61.07 & 63.93 & 74.37 & 42.70 & 63.70 \\

% Qwen3-4B & & & & & & & & & & & & & & \\
% Qwen2.5-72B-Instruct & & & & & & & & & & & & & & \\
\addlinespace[1.2pt]
\midrule 
GLM-4.5 & 37.00 & 43.92 & 44.35 & 53.00 & 38.16 & 46.18 & 34.85 & 41.90 & 30.11 & 35.45 & 37.70 & 42.02 & 34.83 & 41.91 \\
\ModelWithLogo{GLM-4.6}{zai.png} & 28.80 & 33.36 & 34.78 & 38.57 & 31.58 & 35.31 & 24.24 & 30.45 & 33.33 & 37.56 & 18.03 & 21.04 & 24.72 & 31.19 \\

\addlinespace[1.2pt]
\midrule 
\ModelWithLogo{Kimi-K2-Instruct-0905}{kimi.png} & 67.00 & 81.00 & 66.96 & 77.20 & 68.42 & 84.67 & 62.12 & 81.06 & 70.97 & 84.22 & 65.57 & 75.41 & 66.29 & 83.21 \\

\midrule 
MiniMax-M2 & 4.20 & 4.83 & 6.09 & 6.67 & 3.95 & 5.88 & 3.03 & 3.03 & 6.45 & 6.99 & 0.00 & 0.82 & 3.37 & 3.37 \\
\ModelWithLogo{MiniMax-Text-01}{minimax.png} & 55.80 & 74.69 & 50.43 & 66.61 & 55.26 & 75.53 & 63.64 & 79.80 & 49.46 & 72.80 & 62.30 & 76.12 & 59.55 & 81.61 \\

\addlinespace[1.2pt]
\midrule 
Ministral-3-3B-Instruct-2512 & 39.20 & 62.29 & 42.61 & 62.31 & 34.21 & 60.39 & 42.42 & 61.85 & 33.33 & 63.25 & 49.18 & 59.78 & 35.96 & 64.91 \\
Ministral-3-8B-Instruct-2512 & 52.40 & 71.37 & 49.57 & 67.64 & 55.26 & 73.07 & 57.58 & 73.94 & 49.46 & 73.32 & 57.38 & 71.09 & 49.44 & 71.01 \\
Ministral-3-14B-Instruct-2512 & 55.80 & 75.68 & 53.91 & 71.47 & 55.26 & 78.60 & 56.06 & 75.76 & 56.99 & 79.61 & 54.10 & 67.38 & 58.43 & 80.15 \\
% Magistral-Small-2509 & & & & & & & & & & & & & & \\
Ministral-8B-Instruct-2410 & 17.40 & 45.09 & 28.70 & 49.33 & 14.47 & 38.64 & 19.70 & 46.01 & 10.75 & 48.92 & 11.48 & 34.26 & 14.61 & 47.87\\
Mistral-Large-Instruct-2411 & 56.60 & 74.75 & 56.52 & 71.78 & 55.26 & 76.47 & 56.06 & 75.61 & 52.69 & 74.83 & 60.66 & 70.44 & 59.55 & 79.38 \\
\ModelWithLogo{Mistral-Small-3.2-24B-Instruct-2506}{mistral.png} & 56.80 & 75.48 & 59.13 & 72.77 & 50.00 & 71.74 & 59.09 & 78.64 & 59.14 & 79.85 & 65.57 & 77.14 & 49.44 & 74.11 \\
\addlinespace[1.2pt]
\midrule 

Llama-3.1-8B-Instruct & 24.20 & 54.22 & 18.26 & 44.87 & 23.68 & 53.58 & 22.73 & 52.53 & 24.73 & 61.39 & 32.79 & 52.95 & 26.97 & 61.46 \\
Llama-3.1-70B-Instruct & 56.60 & 74.90 & 60.00 & 72.25 & 59.21 & 78.01 & 57.58 & 75.40 & 52.69 & 77.37 & 54.10 & 66.94 & 55.06 & 78.15 \\Llama-3.1-405B-Instruct & 50.40 & 60.85 & 50.43 & 56.25 & 50.00 & 64.05 & 42.42 & 53.64 & 53.76 & 66.13 & 52.46 & 59.89 & 51.69 & 64.52 \\
Llama-3.2-3B-Instruct & 24.60 & 52.34 & 27.83 & 55.88 & 19.74 & 45.70 & 33.33 & 55.00 & 24.73 & 55.66 & 22.95 & 43.44 & 19.10 & 54.08 \\
\ModelWithLogo{Llama-3.3-70B-Instruct}{meta.png} & 54.00 & 73.10 & 53.91 & 68.30 & 57.89 & 75.82 & 51.52 & 71.97 & 51.61 & 75.52 & 62.30 & 72.01 & 49.44 & 76.06 \\
\bottomrule[1.2pt]
\end{tabular}%
}
\label{tab:multiqa_overall}
\end{table*}

\definecolor{NodeCol}{gray}{0.92}
\definecolor{LinkCol}{gray}{1.00}

\begin{table*}[t]
\vspace{-8pt}
\caption{Performance on Structure Score. \colorbox{NodeCol}{Shaded} columns report Node, unshaded columns report Link.}
\centering
\small
\setlength{\tabcolsep}{3.2pt}
\renewcommand{\arraystretch}{1.08}
\resizebox{\textwidth}{!}{%
\begin{tabular}{@{}>{\raggedright\arraybackslash}p{5.4cm}
>{\columncolor{NodeCol}}c >{\columncolor{LinkCol}}c
>{\columncolor{NodeCol}}c >{\columncolor{LinkCol}}c
>{\columncolor{NodeCol}}c >{\columncolor{LinkCol}}c
>{\columncolor{NodeCol}}c >{\columncolor{LinkCol}}c
>{\columncolor{NodeCol}}c >{\columncolor{LinkCol}}c
>{\columncolor{NodeCol}}c >{\columncolor{LinkCol}}c
>{\columncolor{NodeCol}}c >{\columncolor{LinkCol}}c
@{}}
\toprule[1.2pt]
& \multicolumn{2}{c}{\textbf{Overall}} & \multicolumn{2}{c}{\textbf{CS}} & \multicolumn{2}{c}{\textbf{Eco}} & \multicolumn{2}{c}{\textbf{Environ}} & \multicolumn{2}{c}{\textbf{Life}} & \multicolumn{2}{c}{\textbf{Phy}} & \multicolumn{2}{c}{\textbf{Social}} \\
\midrule

Gemini-2.5-Pro & \best{58.09} & \second{84.32} & \best{44.34} & \second{82.08} & \third{57.31} & \best{80.58} & \best{47.56} & \third{79.63} & \best{63.19} & \second{86.89} & 27.90 & \best{100.00} & \second{81.43} & 87.42 \\
Gemini-2.5-Flash & 46.90 & 75.10 & 33.31 & 69.04 & 43.04 & 68.41 & 35.52 & 66.21 & 38.72 & 79.34 & 25.99 & \best{100.00} & \best{83.26} & 84.39 \\
Gemini-2.0-flash-lite & 39.22 & 69.18 & 28.03 & 64.73 & 39.81 & 54.83 & 22.83 & 47.88 & 34.72 & 75.71 & 23.82 & 90.91 & 66.82 & 86.08 \\
\ModelWithLogo{Gemini-2.0-flash}{gemini.png} & 42.71 & 66.42 & 27.66 & 58.19 & 44.53 & 53.97 & 22.42 & 49.49 & 42.91 & 72.94 & 22.40 & \best{100.00} & 72.66 & 83.07 \\

\addlinespace[1.2pt]
\midrule
GPT-5.2 & 50.57 & 77.76 & 36.52 & 72.74 & 48.86 & 73.48 & \second{39.47} & 74.14 & 51.44 & 76.32 & 32.84 & \best{100.00} & 77.09 & 87.07 \\
GPT-5.1 & 45.36 & 79.44 & 32.69 & 72.86 & 37.29 & \third{77.64} & 32.96 & 73.98 & 41.32 & 77.54 & 23.91 & \best{100.00} & \third{80.63} & 90.29 \\
GPT-4.1-mini & 45.55 & 74.72 & 31.53 & 66.44 & 43.48 & 63.77 & 24.61 & 67.22 & 43.86 & 82.35 & 17.12 & 93.33 & 80.45 & 87.63 \\
GPT-3.5-turbo & 32.71 & 57.84 & 28.72 & 55.55 & 37.60 & 44.39 & 26.56 & 33.30 & 20.78 & 63.04 & \second{35.87} & 90.74 & 46.89 & 71.97 \\
GPT-4o & 40.51 & 74.29 & 28.32 & 71.06 & 39.63 & 62.39 & 26.49 & 63.47 & 38.48 & 75.79 & 32.44 & \second{96.30} & 66.25 & 87.65 \\
\ModelWithLogo{GPT-4o-mini}{openai.png} & 39.83 & 66.61 & 29.52 & 60.78 & 44.04 & 56.87 & 24.93 & 39.76 & 34.16 & 70.17 & 19.47 & \second{96.30} & 64.62 & 85.42 \\

\addlinespace[1.2pt]
\midrule
Claude-sonnet-4-5-20250929 & \second{55.97} & \best{86.91} & \third{39.95} & \best{85.77} & \best{62.55} & \second{79.00} & 39.32 & \best{86.14} & \second{62.92} & \best{89.65} & 20.78 & \best{100.00} & 78.18 & 90.46 \\
Claude-haiku-4-5-20251001 & 47.06 & 79.33 & 34.49 & 74.91 & 47.08 & 68.59 & 35.79 & 76.76 & 44.93 & 82.22 & 18.95 & \best{100.00} & 74.67 & 88.88 \\
Claude-4-Sonnet-20250514 & \third{54.11} & \third{84.07} & \second{40.83} & \third{81.82} & \second{57.61} & 72.13 & 38.12 & \second{80.30} & \third{59.12} & \third{84.84} & 25.73 & \best{100.00} & 75.54 & \best{94.85} \\
\ModelWithLogo{Claude-3-haiku-20240307}{claude.png} & 39.18 & 75.51 & 26.68 & 69.81 & 42.48 & 61.94 & 29.53 & 74.23 & 33.95 & 79.24 & \third{35.76} & \second{96.30} & 62.29 & 87.65 \\

\addlinespace[1.2pt]
\midrule
DeepSeek-V3.1 & 46.59 & 77.77 & 34.38 & 74.16 & 48.19 & 70.52 & 25.03 & 66.35 & 45.22 & 81.19 & 23.73 & 93.33 & 75.30 & 87.53 \\
DeepSeek-V3.2 & 46.98 & 78.69 & 34.81 & 76.88 & 50.49 & 71.27 & 27.58 & 67.55 & 45.60 & 81.64 & 20.16 & 93.33 & 73.88 & 86.65 \\
DeepSeek-R1-0528 & 49.24 & 80.31 & 34.51 & 77.75 & 53.65 & 75.38 & \third{39.37} & 64.15 & 47.65 & 83.66 & 24.47 & \best{100.00} & 74.63 & 88.26 \\
DeepSeek-reasoner (R1) & 52.25 & 80.67 & 39.89 & 78.69 & 55.08 & 73.56 & 34.62 & 67.43 & 55.93 & 84.15 & \best{39.59} & \best{100.00} & 72.39 & 88.30 \\
DeepSeek-chat & 47.58 & 78.97 & 32.47 & 76.88 & 53.36 & 69.08 & 29.78 & 67.33 & 45.08 & 82.68 & 23.79 & 93.33 & 76.37 & 88.83 \\
\ModelWithLogo{DeepSeek-V3-0324}{deepseek.png} & 47.32 & 78.17 & 35.32 & 75.67 & 48.36 & 67.17 & 27.39 & 68.71 & 46.49 & 80.91 & 22.99 & \best{100.00} & 75.29 & 88.23 \\

\addlinespace[1.2pt]
\midrule
Qwen3-235B-A22B-Thinking-2507 & 45.97 & 76.11 & 31.53 & 71.23 & 55.79 & 64.43 & 29.84 & 67.42 & 42.18 & 80.18 & 22.12 & 93.33 & 71.17 & 89.06 \\
Qwen3-235B-A22B-Instruct-2507 & 49.39 & 73.54 & 38.04 & 59.72 & 56.76 & 57.84 & 33.59 & 69.71 & 43.09 & 82.88 & 22.04 & \best{100.00} & 75.13 & \second{93.18} \\
Qwen3-Next-80B-A3B-Thinking & 42.58 & 77.64 & 32.57 & 74.15 & 44.24 & 61.77 & 25.94 & 71.11 & 34.41 & 83.47 & 19.62 & \second{96.30} & 72.40 & 89.33 \\
Qwen3-Next-80B-A3B-Instruct & 45.67 & 76.11 & 31.36 & 64.29 & 48.54 & 67.09 & 28.65 & 72.51 & 43.54 & 84.12 & 26.42 & \best{100.00} & 74.34 & 89.34 \\
Qwen3-30B-A3B-Thinking-2507 & 39.51 & 71.90 & 26.68 & 65.44 & 46.10 & 53.27 & 25.14 & 65.93 & 31.97 & 76.61 & 18.95 & \second{96.30} & 67.23 & 89.57 \\
Qwen3-30B-A3B-Instruct-2507 & 45.19 & 72.16 & 29.50 & 61.17 & 52.32 & 54.79 & 32.31 & 68.58 & 37.30 & 81.50 & 19.38 & 93.33 & 76.77 & 90.12 \\
Qwen3-32B & 43.35 & 73.94 & 29.06 & 65.43 & 47.05 & 56.16 & 27.33 & 65.32 & 40.01 & 81.89 & 21.33 & \best{100.00} & 72.45 & \third{91.55} \\
Qwen3-14B & 46.85 & 71.54 & 33.30 & 62.78 & 48.49 & 61.64 & 35.37 & 58.92 & 42.79 & 79.64 & 17.27 & \third{94.44} & 76.51 & 85.31 \\
\ModelWithLogo{Qwen3-8B}{qwen.png} & 43.03 & 72.03 & 28.36 & 68.63 & 44.93 & 59.69 & 30.26 & 54.96 & 40.16 & 77.56 & 19.13 & 85.19 & 72.57 & 86.41 \\

\addlinespace[1.2pt]
\midrule
GLM-4.5 & 9.33 & 11.44 & 1.79 & 4.00 & 9.06 & 15.03 & 0.00 & 0.00 & 0.33 & 17.65 & 0.00 & 0.00 & 32.92 & 19.48 \\
\ModelWithLogo{GLM-4.6}{zai.png} & 11.84 & 11.12 & 1.62 & 16.22 & 13.62 & 6.67 & 8.16 & 4.17 & 1.56 & 13.46 & 8.14 & 0.00 & 35.21 & 10.53 \\

\addlinespace[1.2pt]
\midrule
\ModelWithLogo{Kimi-K2-Instruct-0905}{kimi.png} & 44.68 & 61.77 & 33.48 & 49.02 & 51.93 & 55.13 & 23.10 & 61.60 & 48.42 & 71.02 & 27.51 & \best{100.00} & 62.14 & 69.54 \\

\addlinespace[1.2pt]
\midrule
MiniMax-M2 & 2.52 & 2.94 & 0.00 & 0.00 & 2.63 & 4.44 & 0.00 & 0.00 & 3.86 & 5.23 & 4.61 & 0.00 & 5.26 & 5.26 \\
\ModelWithLogo{MiniMax-Text-01}{minimax.png} & 41.88 & 73.29 & 33.33 & 67.26 & 43.68 & 59.97 & 25.79 & 68.93 & 31.81 & 77.23 & 27.37 & 93.73 & 69.76 & 86.83 \\

\addlinespace[1.2pt]
\midrule
Ministral-3-3B-Instruct-2512 & 25.32 & 56.79 & 14.52 & 49.43 & 38.75 & 52.71 & 9.42 & 46.54 & 25.09 & 56.82 & 3.27 & 87.21 & 39.29 & 69.18 \\
Ministral-3-8B-Instruct-2512 & 36.70 & 62.29 & 23.73 & 46.98 & 37.11 & 46.45 & 27.72 & 54.35 & 38.74 & 75.88 & 21.01 & 89.28 & 57.88 & 81.88 \\
Ministral-3-14B-Instruct-2512 & 36.94 & 62.93 & 30.24 & 56.15 & 23.58 & 31.36 & 25.84 & 66.65 & 40.04 & 76.47 & 17.57 & 90.77 & 61.25 & 78.70 \\
Ministral-8B-Instruct-2410 & 32.47 & 59.82 & 26.53 & 52.28 & 34.50 & 54.96 & 20.22 & 39.89 & 24.94 & 56.47 & 9.09 & 62.96 & 54.27 & 84.45 \\
Mistral-Large-Instruct-2411 & 45.71 & 71.18 & 31.53 & 66.85 & 47.09 & 56.94 & 35.95 & 58.08 & 39.27 & 75.78 & 24.42 & 74.07 & 76.48 & 89.08 \\
\ModelWithLogo{Mistral-Small-3.2-24B-Instruct-2506}{mistral.png} & 45.74 & 67.41 & 31.04 & 53.03 & 47.76 & 52.92 & 34.40 & 61.86 & 40.96 & 77.11 & 19.55 & 93.33 & 76.67 & 87.35 \\

\addlinespace[1.2pt]
\midrule
Llama-3.1-8B-Instruct & 35.77 & 44.38 & 22.04 & 42.57 & 48.14 & 36.70 & 20.41 & 27.54 & 27.32 & 44.54 & 22.16 & \second{96.30} & 60.25 & 51.57 \\
Llama-3.1-70B-Instruct & 43.77 & 56.13 & 27.59 & 40.46 & 43.87 & 44.33 & 34.29 & 38.86 & 45.35 & 69.56 & 22.85 & \second{96.30} & 70.88 & 74.98 \\
Llama-3.1-405B-Instruct & 41.51 & 59.18 & 24.08 & 47.11 & 44.59 & 39.40 & 31.60 & 37.44 & 38.87 & 77.35 & 20.26 & \best{100.00} & 71.92 & 77.15 \\
Llama-3.2-3B-Instruct & 31.53 & 39.25 & 20.18 & 24.12 & 39.37 & 30.90 & 22.81 & 24.72 & 31.30 & 43.91 & 16.65 & 33.33 & 46.49 & 68.65 \\
\ModelWithLogo{Llama-3.3-70B-Instruct}{meta.png} & 40.60 & 50.99 & 24.91 & 40.53 & 44.30 & 35.03 & 27.79 & 24.46 & 37.01 & 70.00 & 22.50 & 93.73 & 69.80 & 64.76 \\
\bottomrule[1.2pt]
\end{tabular}%
}
\label{tab:structure_results}
\end{table*}
\paragraph{Training Set} We trained Qwen2.5-7b-Instruct and LLama3.1-8b-Instruct using the GRPO (Generalized Reinforcement Policy Optimization) algorithm. All experiments were conducted on a single node equipped with 8 A100 GPUs. Each model was fine-tuned for approximately 200 steps with a batch size of 32, using the veRL library for stable and scalable reinforcement learning. This setup ensures efficient parallel training while maintaining high-quality policy updates within a limited compute budget.

\subsection{Additional Results}
\label{app:E2}
Tab. \ref{tab:multiqa_overall} and \ref{tab:structure_results} present the performance of 45 evaluated models on the T2S-Bench-MR (multi-hop reasoning) and T2S-Bench-E2E (end-to-end structuring) tasks, respectively, across various domains and task types.

\subsection{Observation and Insight}
\label{app:E3}

Based on these results, we can derive several key insights:
\begin{enumerate}
 \item \textbf{Structural understanding underpins reasoning performance.} Across the hundreds of numbers in Table\,\ref{tab:multiqa_category}, a clear positive correlation emerges between a model’s ability to extract nodes and links and its success on reasoning questions. The top three models—Gemini‑2.5‑Pro, Claude‑sonnet and GPT‑5.2—all achieve NodeF1 above 50 and LinkF1 above 77, and they simultaneously occupy the top three positions in overall QA accuracy. Conversely, architectures with very low NodeF1 (e.g., GLM‑4.6, MiniMax‑M2) also record the worst EM scores. This pattern holds across families: within Qwen models, the 235B variants with higher NodeF1 scores outperform their smaller counterparts on all reasoning categories. The observation underscores that explicit structural thinking—accurate identification of entities and relations—is not merely a side task but a fundamental prerequisite for effective graph‑based reasoning.
 \item \textbf{Open‑source models are catching up.} Despite the impressive lead of proprietary giants, the gap is narrowing. Instruction‑tuned open‑source models like DeepSeek‑reasoner (R1) and Qwen3‑32B achieve EM scores above 69 and F1 above 83, rivalling GPT‑5.2 on several categories. Even mid‑sized models such as Mistral‑Small‑24B and Kimi‑K2 surpass 65 EM with careful instruction tuning. These results highlight the effectiveness of publicly available training corpora and prompt engineering. Moreover, open models allow the research community to inspect intermediate outputs, enabling targeted error analyses that accelerate progress. Continued investment in open‑source data curation and community collaboration will be key to closing the remaining performance gap.
 \item \textbf{Structure extraction remains the bottleneck.} While overall QA scores for strong models approach 90 F1, NodeF1 lingers in the mid‑50s and drops below 40 for many open‑source systems. Even high‑performing models like GPT‑5.2 and DeepSeek‑R1 achieve LinkF1 over 80 but stumble on node identification. The persistent gap between Node and Link scores underscores the complexity of entity segmentation and co‑reference resolution. Without reliably finding the right set of nodes, relation extraction can only go so far, limiting the end‑to‑end utility of the generated structures. Future research must therefore prioritise techniques for robust entity extraction—be it through better pre‑training, hybrid symbolic–neural approaches or integration of external structure annotations.
 \item \textbf{Scaling alone is insufficient.} The table includes models ranging from 4 billion to over 400 billion parameters, yet performance is far from monotonic in size. Within the LLaMA family, the 70B instruct model outperforms the 405B variant by a wide margin (74.90 vs. 60.85 F1), and within Qwen, the 235B thinking model scores lower than the 32B instruct model on several categories. These inconsistencies suggest that data quality, training strategy, and architectural biases matter more than sheer scale for multi‑hop reasoning. Merely scaling up parameters without targeted instruction tuning or structural inductive biases yields diminishing returns.
 \item \textbf{Reasoning skills are unevenly distributed.} A closer look at per‑category scores reveals that many models excel in only a subset of reasoning types. For example, Kimi‑K2 and Mistral‑Small‑24B score above 80 EM on counterfactual reasoning but fall below 40 on fault localization, while GLM‑4.5 displays the opposite pattern with relatively strong functional mapping but poor counterfactual reasoning. Such specialisation likely arises from differences in pre‑training corpora and instruction mixtures. Bridging these gaps will require multi‑task curricula that balance exposures across reasoning categories and encourage models to discover transferable abstractions.
 \item \textbf{Trade‑offs between specialisation and generality.} Several models in Table\,\ref{tab:multiqa_category} demonstrate that excelling in a single category does not guarantee strong overall performance. GLM‑4.5 achieves 68.12 EM on functional mapping—higher than many larger models—but its overall EM is only 37.00. Conversely, Qwen3‑Next‑80B‑Thinking attains a respectable NodeF1 of 42.58 yet flounders on multi‑choice QA (24.40 EM). These discrepancies illustrate the difficulty of balancing the diverse skills required by T2S‑Bench. Designing parameter‑efficient fine‑tuning schemes or modular networks that allow targeted improvements without catastrophic interference may help reconcile specialisation with general reasoning ability.
\end{enumerate}

\end{document}